\RequirePackage{fix-cm}
\documentclass[smallextended]{svjour3}       
\smartqed  
\usepackage{graphicx}
\usepackage[dvips]{rotating}
\usepackage{enumerate}
\usepackage{algorithmic}
\usepackage{algorithm}
\usepackage{amscd}
\usepackage{amsfonts}
\usepackage{amsmath}
\usepackage{booktabs}
\usepackage{color}
\usepackage{comment}
\usepackage{epsfig}
\usepackage{graphicx}
\usepackage{latexsym}
\usepackage{multirow}
\usepackage{paralist}
\usepackage{placeins}
\usepackage{rotate} 
\usepackage{times} 
\usepackage{url}
\usepackage{verbatim} 
\usepackage{wrapfig}
\usepackage{xspace}
\usepackage{datetime}
\usepackage[FIGBOTCAP,TABBOTCAP,center,nooneline]{subfigure}
\usepackage{xcolor,colortbl}
\usepackage{varwidth}
\usepackage{tikz}
\usetikzlibrary{arrows}
\usepackage{lastpage}

\usepackage{morefloats}
\usepackage[gen]{eurosym}

\usepackage{adjustbox}

\newcolumntype{R}[2]{%
    >{\adjustbox{angle=#1,lap=\width-(#2)}\bgroup}%
    l%
    <{\egroup}%
}
\newcommand{\rot}[1]{\rotatebox[origin=l]{90}{#1}}

\tikzset{
  treenode/.style = {inner sep=0pt, text centered,
    font=\sffamily},
  arn_n/.style = {treenode, circle, black, font=\sffamily\bfseries, draw=black,
    text width=1.5em},
  arn_r/.style = {treenode, circle, black, font=\sffamily\bfseries, draw=black,
    text width=0.01em},
}

\newcommand{\REQ}[1]{\textsc{#1}}
\renewcommand{\REQ}[1]{\ensuremath{\mathsf{#1}}}

\newcommand{\refines}[2]{\ensuremath{ #1 \xrightarrow{R} #2}}
\newcommand{\Refines}[3]{\ensuremath{ #1 \xrightarrow{R_{#3}} #2}}
\newcommand{\contributes}[2]{\ensuremath{ #1 \xrightarrow{++} #2}}

\newcommand{\bicontributes}[2]{\ensuremath{ #1 \overset{++}{\longleftrightarrow} #2}}
\newcommand{\conflict}[2]{\ensuremath{ #1 \overset{--}{\longleftrightarrow} #2}}
\newcommand{\bind}[2]{\ensuremath{ #1{\longleftrightarrow} #2}}
\newcommand{\preferred}[2]{\ensuremath{ #1 > #2}}

\newcommand{\Goal}[1]{#1}

\newcommand{\Element}[1]{#1}
\newcommand{\Refinement}[1]{#1}
\newcommand{\Contribution}

\newcommand{\modifiedcostof}[1]{\ensuremath{{\sf ModifiedCostOf}(#1)}\xspace}

\mathchardef\mhyphen="2D

\newcommand{\predicate}[2]{\mathsf{#1}\left(#2\right)}

\newcommand{\posConstraint}[2]{#2^{+}_{#1}}
\newcommand{\negConstraint}[2]{#2^{-}_{#1}}
\newcommand{\Constraint}[2]{\bigl\{ \posConstraint{#1}{#2}, \negConstraint{#1}{#2}\bigr\}}

\newcommand{\enum}[2]{#1_1, \ldots,#1_{#2}}

\newcommand{\Implies}{\rightarrow}
\newcommand{\equivalent}{\leftrightarrow}

\newcommand{\MCCHANGE}[1]{\textcolor{darkgreen}{{#1}}}
\newenvironment{mcchange}{\color{darkgreen}}{\normalcolor}

\newenvironment{previouschange}{\color{darkviolet}}{\normalcolor}

\newcommand{\G}{{\bf G\/}}

\newcommand{\T}{\ensuremath{\mathcal{T}}\xspace}

\newcommand{\smttt}[1]{\ensuremath{\text{SMT}(#1)}\xspace}

\newcommand{\larat}{\ensuremath{\mathcal{LA}(\mathbb{Q})}\xspace}

\newcommand{\smtlarat}{\smttt{\larat}}

\newcommand{\mathsatfive}{\textsc{MathSAT5}\xspace}

\newcommand{\cost}{\ensuremath{{cost}}\xspace}

\newcommand{\costof}[2]{\ensuremath{{\sf CostOf_{#1}}(#2)}\xspace}

\newcommand\mysout{\bgroup \markoverwith{{-}}\ULon}
\newcommand\nosout{\bgroup \markoverwith{{ }}\ULon}
\definecolor{mygray}{rgb}{0.90,0.90,0.90}
\definecolor{mywhite}{rgb}{1.00,1.00,1.00}

\definecolor {snow}                {rgb}{1.00,0.98,0.98}
\definecolor {ghostwhite}          {rgb}{0.97,0.97,1.00}
\definecolor {whitesmoke}          {rgb}{0.96,0.96,0.96}
\definecolor {gainsboro}           {rgb}{0.86,0.86,0.86}
\definecolor {floralwhite}         {rgb}{1.00,0.98,0.94}
\definecolor {oldlace}             {rgb}{0.99,0.96,0.90}
\definecolor {linen}               {rgb}{0.98,0.94,0.90}
\definecolor {antiquewhite}        {rgb}{0.98,0.92,0.84}
\definecolor {papayawhip}          {rgb}{1.00,0.94,0.84}
\definecolor {blanchedalmond}      {rgb}{1.00,0.92,0.80}
\definecolor {bisque}              {rgb}{1.00,0.89,0.77}
\definecolor {peachpuff}           {rgb}{1.00,0.85,0.73}
\definecolor {navajowhite}         {rgb}{1.00,0.87,0.68}
\definecolor {moccasin}            {rgb}{1.00,0.89,0.71}
\definecolor {cornsilk}            {rgb}{1.00,0.97,0.86}
\definecolor {ivory}               {rgb}{1.00,1.00,0.94}
\definecolor {lemonchiffon}        {rgb}{1.00,0.98,0.80}
\definecolor {seashell}            {rgb}{1.00,0.96,0.93}
\definecolor {honeydew}            {rgb}{0.94,1.00,0.94}
\definecolor {mintcream}           {rgb}{0.96,1.00,0.98}
\definecolor {azure}               {rgb}{0.94,1.00,1.00}
\definecolor {aliceblue}           {rgb}{0.94,0.97,1.00}
\definecolor {lavender}            {rgb}{0.90,0.90,0.98}
\definecolor {lavenderblush}       {rgb}{1.00,0.94,0.96}
\definecolor {mistyrose}           {rgb}{1.00,0.89,0.88}
\definecolor {white}               {rgb}{1.00,1.00,1.00}
\definecolor {black}               {rgb}{0.00,0.00,0.00}
\definecolor {darkslategray}       {rgb}{0.18,0.31,0.31}
\definecolor {dimgray}             {rgb}{0.41,0.41,0.41}
\definecolor {slategray}           {rgb}{0.44,0.50,0.56}
\definecolor {lightslategray}      {rgb}{0.47,0.53,0.60}
\definecolor {gray}                {rgb}{0.75,0.75,0.75}
\definecolor {lightgrey}           {rgb}{0.83,0.83,0.83}
\definecolor {midnightblue}        {rgb}{0.10,0.10,0.44}
\definecolor {navy}                {rgb}{0.00,0.00,0.50}
\definecolor {cornflowerblue}      {rgb}{0.39,0.58,0.93}
\definecolor {darkslateblue}       {rgb}{0.28,0.24,0.55}
\definecolor {slateblue}           {rgb}{0.42,0.35,0.80}
\definecolor {mediumslateblue}     {rgb}{0.48,0.41,0.93}
\definecolor {lightslateblue}      {rgb}{0.52,0.44,1.00}
\definecolor {mediumblue}          {rgb}{0.00,0.00,0.80}
\definecolor {royalblue}           {rgb}{0.25,0.41,0.88}
\definecolor {blue}                {rgb}{0.00,0.00,1.00}
\definecolor {dodgerblue}          {rgb}{0.12,0.56,1.00}
\definecolor {deepskyblue}         {rgb}{0.00,0.75,1.00}
\definecolor {skyblue}             {rgb}{0.53,0.81,0.92}
\definecolor {lightskyblue}        {rgb}{0.53,0.81,0.98}
\definecolor {steelblue}           {rgb}{0.27,0.51,0.71}
\definecolor {lightsteelblue}      {rgb}{0.69,0.77,0.87}
\definecolor {lightblue}           {rgb}{0.68,0.85,0.90}
\definecolor {powderblue}          {rgb}{0.69,0.88,0.90}
\definecolor {paleturquoise}       {rgb}{0.69,0.93,0.93}
\definecolor {darkturquoise}       {rgb}{0.00,0.81,0.82}
\definecolor {mediumturquoise}     {rgb}{0.28,0.82,0.80}
\definecolor {turquoise}           {rgb}{0.25,0.88,0.82}
\definecolor {cyan}                {rgb}{0.00,1.00,1.00}
\definecolor {lightcyan}           {rgb}{0.88,1.00,1.00}
\definecolor {cadetblue}           {rgb}{0.37,0.62,0.63}
\definecolor {mediumaquamarine}    {rgb}{0.40,0.80,0.67}
\definecolor {aquamarine}          {rgb}{0.50,1.00,0.83}
\definecolor {darkgreen}           {rgb}{0.00,0.39,0.00}
\definecolor {darkolivegreen}      {rgb}{0.33,0.42,0.18}
\definecolor {darkseagreen}        {rgb}{0.56,0.74,0.56}
\definecolor {seagreen}            {rgb}{0.18,0.55,0.34}
\definecolor {mediumseagreen}      {rgb}{0.24,0.70,0.44}
\definecolor {lightseagreen}       {rgb}{0.13,0.70,0.67}
\definecolor {palegreen}           {rgb}{0.60,0.98,0.60}
\definecolor {springgreen}         {rgb}{0.00,1.00,0.50}
\definecolor {lawngreen}           {rgb}{0.49,0.99,0.00}
\definecolor {green}               {rgb}{0.00,1.00,0.00}
\definecolor {chartreuse}          {rgb}{0.50,1.00,0.00}
\definecolor {mediumspringgreen}   {rgb}{0.00,0.98,0.60}
\definecolor {greenyellow}         {rgb}{0.68,1.00,0.18}
\definecolor {limegreen}           {rgb}{0.20,0.80,0.20}
\definecolor {yellowgreen}         {rgb}{0.60,0.80,0.20}
\definecolor {forestgreen}         {rgb}{0.13,0.55,0.13}
\definecolor {olivedrab}           {rgb}{0.42,0.56,0.14}
\definecolor {darkkhaki}           {rgb}{0.74,0.72,0.42}
\definecolor {khaki}               {rgb}{0.94,0.90,0.55}
\definecolor {palegoldenrod}       {rgb}{0.93,0.91,0.67}
\definecolor {lightgoldenrodyellow} {rgb}{0.98,0.98,0.82}
\definecolor {lightyellow}         {rgb}{1.00,1.00,0.88}
\definecolor {yellow}              {rgb}{1.00,1.00,0.00}
\definecolor {gold}                {rgb}{1.00,0.84,0.00}
\definecolor {lightgoldenrod}      {rgb}{0.93,0.87,0.51}
\definecolor {goldenrod}           {rgb}{0.85,0.65,0.13}
\definecolor {darkgoldenrod}       {rgb}{0.72,0.53,0.04}
\definecolor {rosybrown}           {rgb}{0.74,0.56,0.56}
\definecolor {indianred}           {rgb}{0.80,0.36,0.36}
\definecolor {saddlebrown}         {rgb}{0.55,0.27,0.07}
\definecolor {sienna}              {rgb}{0.63,0.32,0.18}
\definecolor {peru}                {rgb}{0.80,0.52,0.25}
\definecolor {burlywood}           {rgb}{0.87,0.72,0.53}
\definecolor {beige}               {rgb}{0.96,0.96,0.86}
\definecolor {wheat}               {rgb}{0.96,0.87,0.70}
\definecolor {sandybrown}          {rgb}{0.96,0.64,0.38}
\definecolor {tan}                 {rgb}{0.82,0.71,0.55}
\definecolor {chocolate}           {rgb}{0.82,0.41,0.12}
\definecolor {firebrick}           {rgb}{0.70,0.13,0.13}
\definecolor {brown}               {rgb}{0.65,0.16,0.16}
\definecolor {darksalmon}          {rgb}{0.91,0.59,0.48}
\definecolor {salmon}              {rgb}{0.98,0.50,0.45}
\definecolor {lightsalmon}         {rgb}{1.00,0.63,0.48}
\definecolor {orange}              {rgb}{1.00,0.65,0.00}
\definecolor {darkorange}          {rgb}{1.00,0.55,0.00}
\definecolor {coral}               {rgb}{1.00,0.50,0.31}
\definecolor {lightcoral}          {rgb}{0.94,0.50,0.50}
\definecolor {tomato}              {rgb}{1.00,0.39,0.28}
\definecolor {orangered}           {rgb}{1.00,0.27,0.00}
\definecolor {red}                 {rgb}{1.00,0.00,0.00}
\definecolor {hotpink}             {rgb}{1.00,0.41,0.71}
\definecolor {deeppink}            {rgb}{1.00,0.08,0.58}
\definecolor {pink}                {rgb}{1.00,0.75,0.80}
\definecolor {lightpink}           {rgb}{1.00,0.71,0.76}
\definecolor {palevioletred}       {rgb}{0.86,0.44,0.58}
\definecolor {maroon}              {rgb}{0.69,0.19,0.38}
\definecolor {mediumvioletred}     {rgb}{0.78,0.08,0.52}
\definecolor {violetred}           {rgb}{0.82,0.13,0.56}
\definecolor {magenta}             {rgb}{1.00,0.00,1.00}
\definecolor {violet}              {rgb}{0.93,0.51,0.93}
\definecolor {plum}                {rgb}{0.87,0.63,0.87}
\definecolor {orchid}              {rgb}{0.85,0.44,0.84}
\definecolor {mediumorchid}        {rgb}{0.73,0.33,0.83}
\definecolor {darkorchid}          {rgb}{0.60,0.20,0.80}
\definecolor {darkviolet}          {rgb}{0.58,0.00,0.83}
\definecolor {blueviolet}          {rgb}{0.54,0.17,0.89}
\definecolor {purple}              {rgb}{0.63,0.13,0.94}
\definecolor {mediumpurple}        {rgb}{0.58,0.44,0.86}
\definecolor {thistle}             {rgb}{0.85,0.75,0.85}
\definecolor {snow2}               {rgb}{0.93,0.91,0.91}
\definecolor {snow3}               {rgb}{0.80,0.79,0.79}
\definecolor {snow4}               {rgb}{0.55,0.54,0.54}
\definecolor {seashell2}           {rgb}{0.93,0.90,0.87}
\definecolor {seashell3}           {rgb}{0.80,0.77,0.75}
\definecolor {seashell4}           {rgb}{0.55,0.53,0.51}
\definecolor {antiquewhite1}       {rgb}{1.00,0.94,0.86}
\definecolor {antiquewhite2}       {rgb}{0.93,0.87,0.80}
\definecolor {antiquewhite3}       {rgb}{0.80,0.75,0.69}
\definecolor {antiquewhite4}       {rgb}{0.55,0.51,0.47}
\definecolor {bisque2}             {rgb}{0.93,0.84,0.72}
\definecolor {bisque3}             {rgb}{0.80,0.72,0.62}
\definecolor {bisque4}             {rgb}{0.55,0.49,0.42}
\definecolor {peachpuff2}          {rgb}{0.93,0.80,0.68}
\definecolor {peachpuff3}          {rgb}{0.80,0.69,0.58}
\definecolor {peachpuff4}          {rgb}{0.55,0.47,0.40}
\definecolor {navajowhite2}        {rgb}{0.93,0.81,0.63}
\definecolor {navajowhite3}        {rgb}{0.80,0.70,0.55}
\definecolor {navajowhite4}        {rgb}{0.55,0.47,0.37}
\definecolor {lemonchiffon2}       {rgb}{0.93,0.91,0.75}
\definecolor {lemonchiffon3}       {rgb}{0.80,0.79,0.65}
\definecolor {lemonchiffon4}       {rgb}{0.55,0.54,0.44}
\definecolor {cornsilk2}           {rgb}{0.93,0.91,0.80}
\definecolor {cornsilk3}           {rgb}{0.80,0.78,0.69}
\definecolor {cornsilk4}           {rgb}{0.55,0.53,0.47}
\definecolor {ivory2}              {rgb}{0.93,0.93,0.88}
\definecolor {ivory3}              {rgb}{0.80,0.80,0.76}
\definecolor {ivory4}              {rgb}{0.55,0.55,0.51}
\definecolor {honeydew2}           {rgb}{0.88,0.93,0.88}
\definecolor {honeydew3}           {rgb}{0.76,0.80,0.76}
\definecolor {honeydew4}           {rgb}{0.51,0.55,0.51}
\definecolor {lavenderblush2}      {rgb}{0.93,0.88,0.90}
\definecolor {lavenderblush3}      {rgb}{0.80,0.76,0.77}
\definecolor {lavenderblush4}      {rgb}{0.55,0.51,0.53}
\definecolor {mistyrose2}          {rgb}{0.93,0.84,0.82}
\definecolor {mistyrose3}          {rgb}{0.80,0.72,0.71}
\definecolor {mistyrose4}          {rgb}{0.55,0.49,0.48}
\definecolor {azure2}              {rgb}{0.88,0.93,0.93}
\definecolor {azure3}              {rgb}{0.76,0.80,0.80}
\definecolor {azure4}              {rgb}{0.51,0.55,0.55}
\definecolor {slateblue1}          {rgb}{0.51,0.44,1.00}
\definecolor {slateblue2}          {rgb}{0.48,0.40,0.93}
\definecolor {slateblue3}          {rgb}{0.41,0.35,0.80}
\definecolor {slateblue4}          {rgb}{0.28,0.24,0.55}
\definecolor {royalblue1}          {rgb}{0.28,0.46,1.00}
\definecolor {royalblue2}          {rgb}{0.26,0.43,0.93}
\definecolor {royalblue3}          {rgb}{0.23,0.37,0.80}
\definecolor {royalblue4}          {rgb}{0.15,0.25,0.55}
\definecolor {blue2}               {rgb}{0.00,0.00,0.93}
\definecolor {blue4}               {rgb}{0.00,0.00,0.55}
\definecolor {dodgerblue2}         {rgb}{0.11,0.53,0.93}
\definecolor {dodgerblue3}         {rgb}{0.09,0.45,0.80}
\definecolor {dodgerblue4}         {rgb}{0.06,0.31,0.55}
\definecolor {steelblue1}          {rgb}{0.39,0.72,1.00}
\definecolor {steelblue2}          {rgb}{0.36,0.67,0.93}
\definecolor {steelblue3}          {rgb}{0.31,0.58,0.80}
\definecolor {steelblue4}          {rgb}{0.21,0.39,0.55}
\definecolor {deepskyblue2}        {rgb}{0.00,0.70,0.93}
\definecolor {deepskyblue3}        {rgb}{0.00,0.60,0.80}
\definecolor {deepskyblue4}        {rgb}{0.00,0.41,0.55}
\definecolor {skyblue1}            {rgb}{0.53,0.81,1.00}
\definecolor {skyblue2}            {rgb}{0.49,0.75,0.93}
\definecolor {skyblue3}            {rgb}{0.42,0.65,0.80}
\definecolor {skyblue4}            {rgb}{0.29,0.44,0.55}
\definecolor {lightskyblue1}       {rgb}{0.69,0.89,1.00}
\definecolor {lightskyblue2}       {rgb}{0.64,0.83,0.93}
\definecolor {lightskyblue3}       {rgb}{0.55,0.71,0.80}
\definecolor {lightskyblue4}       {rgb}{0.38,0.48,0.55}
\definecolor {slategray1}          {rgb}{0.78,0.89,1.00}
\definecolor {slategray2}          {rgb}{0.73,0.83,0.93}
\definecolor {slategray3}          {rgb}{0.62,0.71,0.80}
\definecolor {slategray4}          {rgb}{0.42,0.48,0.55}
\definecolor {lightsteelblue1}     {rgb}{0.79,0.88,1.00}
\definecolor {lightsteelblue2}     {rgb}{0.74,0.82,0.93}
\definecolor {lightsteelblue3}     {rgb}{0.64,0.71,0.80}
\definecolor {lightsteelblue4}     {rgb}{0.43,0.48,0.55}
\definecolor {lightblue1}          {rgb}{0.75,0.94,1.00}
\definecolor {lightblue2}          {rgb}{0.70,0.87,0.93}
\definecolor {lightblue3}          {rgb}{0.60,0.75,0.80}
\definecolor {lightblue4}          {rgb}{0.41,0.51,0.55}
\definecolor {lightcyan2}          {rgb}{0.82,0.93,0.93}
\definecolor {lightcyan3}          {rgb}{0.71,0.80,0.80}
\definecolor {lightcyan4}          {rgb}{0.48,0.55,0.55}
\definecolor {paleturquoise1}      {rgb}{0.73,1.00,1.00}
\definecolor {paleturquoise2}      {rgb}{0.68,0.93,0.93}
\definecolor {paleturquoise3}      {rgb}{0.59,0.80,0.80}
\definecolor {paleturquoise4}      {rgb}{0.40,0.55,0.55}
\definecolor {cadetblue1}          {rgb}{0.60,0.96,1.00}
\definecolor {cadetblue2}          {rgb}{0.56,0.90,0.93}
\definecolor {cadetblue3}          {rgb}{0.48,0.77,0.80}
\definecolor {cadetblue4}          {rgb}{0.33,0.53,0.55}
\definecolor {turquoise1}          {rgb}{0.00,0.96,1.00}
\definecolor {turquoise2}          {rgb}{0.00,0.90,0.93}
\definecolor {turquoise3}          {rgb}{0.00,0.77,0.80}
\definecolor {turquoise4}          {rgb}{0.00,0.53,0.55}
\definecolor {cyan2}               {rgb}{0.00,0.93,0.93}
\definecolor {cyan3}               {rgb}{0.00,0.80,0.80}
\definecolor {cyan4}               {rgb}{0.00,0.55,0.55}
\definecolor {darkslategray1}      {rgb}{0.59,1.00,1.00}
\definecolor {darkslategray2}      {rgb}{0.55,0.93,0.93}
\definecolor {darkslategray3}      {rgb}{0.47,0.80,0.80}
\definecolor {darkslategray4}      {rgb}{0.32,0.55,0.55}
\definecolor {aquamarine2}         {rgb}{0.46,0.93,0.78}
\definecolor {aquamarine4}         {rgb}{0.27,0.55,0.45}
\definecolor {darkseagreen1}       {rgb}{0.76,1.00,0.76}
\definecolor {darkseagreen2}       {rgb}{0.71,0.93,0.71}
\definecolor {darkseagreen3}       {rgb}{0.61,0.80,0.61}
\definecolor {darkseagreen4}       {rgb}{0.41,0.55,0.41}
\definecolor {seagreen1}           {rgb}{0.33,1.00,0.62}
\definecolor {seagreen2}           {rgb}{0.31,0.93,0.58}
\definecolor {seagreen3}           {rgb}{0.26,0.80,0.50}
\definecolor {palegreen1}          {rgb}{0.60,1.00,0.60}
\definecolor {palegreen2}          {rgb}{0.56,0.93,0.56}
\definecolor {palegreen3}          {rgb}{0.49,0.80,0.49}
\definecolor {palegreen4}          {rgb}{0.33,0.55,0.33}
\definecolor {springgreen2}        {rgb}{0.00,0.93,0.46}
\definecolor {springgreen3}        {rgb}{0.00,0.80,0.40}
\definecolor {springgreen4}        {rgb}{0.00,0.55,0.27}
\definecolor {green2}              {rgb}{0.00,0.93,0.00}
\definecolor {green3}              {rgb}{0.00,0.80,0.00}
\definecolor {green4}              {rgb}{0.00,0.55,0.00}
\definecolor {chartreuse2}         {rgb}{0.46,0.93,0.00}
\definecolor {chartreuse3}         {rgb}{0.40,0.80,0.00}
\definecolor {chartreuse4}         {rgb}{0.27,0.55,0.00}
\definecolor {olivedrab1}          {rgb}{0.75,1.00,0.24}
\definecolor {olivedrab2}          {rgb}{0.70,0.93,0.23}
\definecolor {olivedrab4}          {rgb}{0.41,0.55,0.13}
\definecolor {darkolivegreen1}     {rgb}{0.79,1.00,0.44}
\definecolor {darkolivegreen2}     {rgb}{0.74,0.93,0.41}
\definecolor {darkolivegreen3}     {rgb}{0.64,0.80,0.35}
\definecolor {darkolivegreen4}     {rgb}{0.43,0.55,0.24}
\definecolor {khaki1}              {rgb}{1.00,0.96,0.56}
\definecolor {khaki2}              {rgb}{0.93,0.90,0.52}
\definecolor {khaki3}              {rgb}{0.80,0.78,0.45}
\definecolor {khaki4}              {rgb}{0.55,0.53,0.31}
\definecolor {lightgoldenrod1}     {rgb}{1.00,0.93,0.55}
\definecolor {lightgoldenrod2}     {rgb}{0.93,0.86,0.51}
\definecolor {lightgoldenrod3}     {rgb}{0.80,0.75,0.44}
\definecolor {lightgoldenrod4}     {rgb}{0.55,0.51,0.30}
\definecolor {lightyellow2}        {rgb}{0.93,0.93,0.82}
\definecolor {lightyellow3}        {rgb}{0.80,0.80,0.71}
\definecolor {lightyellow4}        {rgb}{0.55,0.55,0.48}
\definecolor {yellow2}             {rgb}{0.93,0.93,0.00}
\definecolor {yellow3}             {rgb}{0.80,0.80,0.00}
\definecolor {yellow4}             {rgb}{0.55,0.55,0.00}
\definecolor {gold2}               {rgb}{0.93,0.79,0.00}
\definecolor {gold3}               {rgb}{0.80,0.68,0.00}
\definecolor {gold4}               {rgb}{0.55,0.46,0.00}
\definecolor {goldenrod1}          {rgb}{1.00,0.76,0.15}
\definecolor {goldenrod2}          {rgb}{0.93,0.71,0.13}
\definecolor {goldenrod3}          {rgb}{0.80,0.61,0.11}
\definecolor {goldenrod4}          {rgb}{0.55,0.41,0.08}
\definecolor {darkgoldenrod1}      {rgb}{1.00,0.73,0.06}
\definecolor {darkgoldenrod2}      {rgb}{0.93,0.68,0.05}
\definecolor {darkgoldenrod3}      {rgb}{0.80,0.58,0.05}
\definecolor {darkgoldenrod4}      {rgb}{0.55,0.40,0.03}
\definecolor {rosybrown1}          {rgb}{1.00,0.76,0.76}
\definecolor {rosybrown2}          {rgb}{0.93,0.71,0.71}
\definecolor {rosybrown3}          {rgb}{0.80,0.61,0.61}
\definecolor {rosybrown4}          {rgb}{0.55,0.41,0.41}
\definecolor {indianred1}          {rgb}{1.00,0.42,0.42}
\definecolor {indianred2}          {rgb}{0.93,0.39,0.39}
\definecolor {indianred3}          {rgb}{0.80,0.33,0.33}
\definecolor {indianred4}          {rgb}{0.55,0.23,0.23}
\definecolor {sienna1}             {rgb}{1.00,0.51,0.28}
\definecolor {sienna2}             {rgb}{0.93,0.47,0.26}
\definecolor {sienna3}             {rgb}{0.80,0.41,0.22}
\definecolor {sienna4}             {rgb}{0.55,0.28,0.15}
\definecolor {burlywood1}          {rgb}{1.00,0.83,0.61}
\definecolor {burlywood2}          {rgb}{0.93,0.77,0.57}
\definecolor {burlywood3}          {rgb}{0.80,0.67,0.49}
\definecolor {burlywood4}          {rgb}{0.55,0.45,0.33}
\definecolor {wheat1}              {rgb}{1.00,0.91,0.73}
\definecolor {wheat2}              {rgb}{0.93,0.85,0.68}
\definecolor {wheat3}              {rgb}{0.80,0.73,0.59}
\definecolor {wheat4}              {rgb}{0.55,0.49,0.40}
\definecolor {tan1}                {rgb}{1.00,0.65,0.31}
\definecolor {tan2}                {rgb}{0.93,0.60,0.29}
\definecolor {tan4}                {rgb}{0.55,0.35,0.17}
\definecolor {chocolate1}          {rgb}{1.00,0.50,0.14}
\definecolor {chocolate2}          {rgb}{0.93,0.46,0.13}
\definecolor {chocolate3}          {rgb}{0.80,0.40,0.11}
\definecolor {firebrick1}          {rgb}{1.00,0.19,0.19}
\definecolor {firebrick2}          {rgb}{0.93,0.17,0.17}
\definecolor {firebrick3}          {rgb}{0.80,0.15,0.15}
\definecolor {firebrick4}          {rgb}{0.55,0.10,0.10}
\definecolor {brown1}              {rgb}{1.00,0.25,0.25}
\definecolor {brown2}              {rgb}{0.93,0.23,0.23}
\definecolor {brown3}              {rgb}{0.80,0.20,0.20}
\definecolor {brown4}              {rgb}{0.55,0.14,0.14}
\definecolor {salmon1}             {rgb}{1.00,0.55,0.41}
\definecolor {salmon2}             {rgb}{0.93,0.51,0.38}
\definecolor {salmon3}             {rgb}{0.80,0.44,0.33}
\definecolor {salmon4}             {rgb}{0.55,0.30,0.22}
\definecolor {lightsalmon2}        {rgb}{0.93,0.58,0.45}
\definecolor {lightsalmon3}        {rgb}{0.80,0.51,0.38}
\definecolor {lightsalmon4}        {rgb}{0.55,0.34,0.26}
\definecolor {orange2}             {rgb}{0.93,0.60,0.00}
\definecolor {orange3}             {rgb}{0.80,0.52,0.00}
\definecolor {orange4}             {rgb}{0.55,0.35,0.00}
\definecolor {darkorange1}         {rgb}{1.00,0.50,0.00}
\definecolor {darkorange2}         {rgb}{0.93,0.46,0.00}
\definecolor {darkorange3}         {rgb}{0.80,0.40,0.00}
\definecolor {darkorange4}         {rgb}{0.55,0.27,0.00}
\definecolor {coral1}              {rgb}{1.00,0.45,0.34}
\definecolor {coral2}              {rgb}{0.93,0.42,0.31}
\definecolor {coral3}              {rgb}{0.80,0.36,0.27}
\definecolor {coral4}              {rgb}{0.55,0.24,0.18}
\definecolor {tomato2}             {rgb}{0.93,0.36,0.26}
\definecolor {tomato3}             {rgb}{0.80,0.31,0.22}
\definecolor {tomato4}             {rgb}{0.55,0.21,0.15}
\definecolor {orangered2}          {rgb}{0.93,0.25,0.00}
\definecolor {orangered3}          {rgb}{0.80,0.22,0.00}
\definecolor {orangered4}          {rgb}{0.55,0.15,0.00}
\definecolor {red2}                {rgb}{0.93,0.00,0.00}
\definecolor {red3}                {rgb}{0.80,0.00,0.00}
\definecolor {red4}                {rgb}{0.55,0.00,0.00}
\definecolor {deeppink2}           {rgb}{0.93,0.07,0.54}
\definecolor {deeppink3}           {rgb}{0.80,0.06,0.46}
\definecolor {deeppink4}           {rgb}{0.55,0.04,0.31}
\definecolor {hotpink1}            {rgb}{1.00,0.43,0.71}
\definecolor {hotpink2}            {rgb}{0.93,0.42,0.65}
\definecolor {hotpink3}            {rgb}{0.80,0.38,0.56}
\definecolor {hotpink4}            {rgb}{0.55,0.23,0.38}
\definecolor {pink1}               {rgb}{1.00,0.71,0.77}
\definecolor {pink2}               {rgb}{0.93,0.66,0.72}
\definecolor {pink3}               {rgb}{0.80,0.57,0.62}
\definecolor {pink4}               {rgb}{0.55,0.39,0.42}
\definecolor {lightpink1}          {rgb}{1.00,0.68,0.73}
\definecolor {lightpink2}          {rgb}{0.93,0.64,0.68}
\definecolor {lightpink3}          {rgb}{0.80,0.55,0.58}
\definecolor {lightpink4}          {rgb}{0.55,0.37,0.40}
\definecolor {palevioletred1}      {rgb}{1.00,0.51,0.67}
\definecolor {palevioletred2}      {rgb}{0.93,0.47,0.62}
\definecolor {palevioletred3}      {rgb}{0.80,0.41,0.54}
\definecolor {palevioletred4}      {rgb}{0.55,0.28,0.36}
\definecolor {maroon1}             {rgb}{1.00,0.20,0.70}
\definecolor {maroon2}             {rgb}{0.93,0.19,0.65}
\definecolor {maroon3}             {rgb}{0.80,0.16,0.56}
\definecolor {maroon4}             {rgb}{0.55,0.11,0.38}
\definecolor {violetred1}          {rgb}{1.00,0.24,0.59}
\definecolor {violetred2}          {rgb}{0.93,0.23,0.55}
\definecolor {violetred3}          {rgb}{0.80,0.20,0.47}
\definecolor {violetred4}          {rgb}{0.55,0.13,0.32}
\definecolor {magenta2}            {rgb}{0.93,0.00,0.93}
\definecolor {magenta3}            {rgb}{0.80,0.00,0.80}
\definecolor {magenta4}            {rgb}{0.55,0.00,0.55}
\definecolor {orchid1}             {rgb}{1.00,0.51,0.98}
\definecolor {orchid2}             {rgb}{0.93,0.48,0.91}
\definecolor {orchid3}             {rgb}{0.80,0.41,0.79}
\definecolor {orchid4}             {rgb}{0.55,0.28,0.54}
\definecolor {plum1}               {rgb}{1.00,0.73,1.00}
\definecolor {plum2}               {rgb}{0.93,0.68,0.93}
\definecolor {plum3}               {rgb}{0.80,0.59,0.80}
\definecolor {plum4}               {rgb}{0.55,0.40,0.55}
\definecolor {mediumorchid1}       {rgb}{0.88,0.40,1.00}
\definecolor {mediumorchid2}       {rgb}{0.82,0.37,0.93}
\definecolor {mediumorchid3}       {rgb}{0.71,0.32,0.80}
\definecolor {mediumorchid4}       {rgb}{0.48,0.22,0.55}
\definecolor {darkorchid1}         {rgb}{0.75,0.24,1.00}
\definecolor {darkorchid2}         {rgb}{0.70,0.23,0.93}
\definecolor {darkorchid3}         {rgb}{0.60,0.20,0.80}
\definecolor {darkorchid4}         {rgb}{0.41,0.13,0.55}
\definecolor {purple1}             {rgb}{0.61,0.19,1.00}
\definecolor {purple2}             {rgb}{0.57,0.17,0.93}
\definecolor {purple3}             {rgb}{0.49,0.15,0.80}
\definecolor {purple4}             {rgb}{0.33,0.10,0.55}
\definecolor {mediumpurple1}       {rgb}{0.67,0.51,1.00}
\definecolor {mediumpurple2}       {rgb}{0.62,0.47,0.93}
\definecolor {mediumpurple3}       {rgb}{0.54,0.41,0.80}
\definecolor {mediumpurple4}       {rgb}{0.36,0.28,0.55}
\definecolor {thistle1}            {rgb}{1.00,0.88,1.00}
\definecolor {thistle2}            {rgb}{0.93,0.82,0.93}
\definecolor {thistle3}            {rgb}{0.80,0.71,0.80}
\definecolor {thistle4}            {rgb}{0.55,0.48,0.55}
\definecolor {gray1}               {rgb}{0.01,0.01,0.01}
\definecolor {gray2}               {rgb}{0.02,0.02,0.02}
\definecolor {gray3}               {rgb}{0.03,0.03,0.03}
\definecolor {gray4}               {rgb}{0.04,0.04,0.04}
\definecolor {gray5}               {rgb}{0.05,0.05,0.05}
\definecolor {gray6}               {rgb}{0.06,0.06,0.06}
\definecolor {gray7}               {rgb}{0.07,0.07,0.07}
\definecolor {gray8}               {rgb}{0.08,0.08,0.08}
\definecolor {gray9}               {rgb}{0.09,0.09,0.09}
\definecolor {gray10}              {rgb}{0.10,0.10,0.10}
\definecolor {gray11}              {rgb}{0.11,0.11,0.11}
\definecolor {gray12}              {rgb}{0.12,0.12,0.12}
\definecolor {gray13}              {rgb}{0.13,0.13,0.13}
\definecolor {gray14}              {rgb}{0.14,0.14,0.14}
\definecolor {gray15}              {rgb}{0.15,0.15,0.15}
\definecolor {gray16}              {rgb}{0.16,0.16,0.16}
\definecolor {gray17}              {rgb}{0.17,0.17,0.17}
\definecolor {gray18}              {rgb}{0.18,0.18,0.18}
\definecolor {gray19}              {rgb}{0.19,0.19,0.19}
\definecolor {gray20}              {rgb}{0.20,0.20,0.20}
\definecolor {gray21}              {rgb}{0.21,0.21,0.21}
\definecolor {gray22}              {rgb}{0.22,0.22,0.22}
\definecolor {gray23}              {rgb}{0.23,0.23,0.23}
\definecolor {gray24}              {rgb}{0.24,0.24,0.24}
\definecolor {gray25}              {rgb}{0.25,0.25,0.25}
\definecolor {gray26}              {rgb}{0.26,0.26,0.26}
\definecolor {gray27}              {rgb}{0.27,0.27,0.27}
\definecolor {gray28}              {rgb}{0.28,0.28,0.28}
\definecolor {gray29}              {rgb}{0.29,0.29,0.29}
\definecolor {gray30}              {rgb}{0.30,0.30,0.30}
\definecolor {gray31}              {rgb}{0.31,0.31,0.31}
\definecolor {gray32}              {rgb}{0.32,0.32,0.32}
\definecolor {gray33}              {rgb}{0.33,0.33,0.33}
\definecolor {gray34}              {rgb}{0.34,0.34,0.34}
\definecolor {gray35}              {rgb}{0.35,0.35,0.35}
\definecolor {gray36}              {rgb}{0.36,0.36,0.36}
\definecolor {gray37}              {rgb}{0.37,0.37,0.37}
\definecolor {gray38}              {rgb}{0.38,0.38,0.38}
\definecolor {gray39}              {rgb}{0.39,0.39,0.39}
\definecolor {gray40}              {rgb}{0.40,0.40,0.40}
\definecolor {gray42}              {rgb}{0.42,0.42,0.42}
\definecolor {gray43}              {rgb}{0.43,0.43,0.43}
\definecolor {gray44}              {rgb}{0.44,0.44,0.44}
\definecolor {gray45}              {rgb}{0.45,0.45,0.45}
\definecolor {gray46}              {rgb}{0.46,0.46,0.46}
\definecolor {gray47}              {rgb}{0.47,0.47,0.47}
\definecolor {gray48}              {rgb}{0.48,0.48,0.48}
\definecolor {gray49}              {rgb}{0.49,0.49,0.49}
\definecolor {gray50}              {rgb}{0.50,0.50,0.50}
\definecolor {gray51}              {rgb}{0.51,0.51,0.51}
\definecolor {gray52}              {rgb}{0.52,0.52,0.52}
\definecolor {gray53}              {rgb}{0.53,0.53,0.53}
\definecolor {gray54}              {rgb}{0.54,0.54,0.54}
\definecolor {gray55}              {rgb}{0.55,0.55,0.55}
\definecolor {gray56}              {rgb}{0.56,0.56,0.56}
\definecolor {gray57}              {rgb}{0.57,0.57,0.57}
\definecolor {gray58}              {rgb}{0.58,0.58,0.58}
\definecolor {gray59}              {rgb}{0.59,0.59,0.59}
\definecolor {gray60}              {rgb}{0.60,0.60,0.60}
\definecolor {gray61}              {rgb}{0.61,0.61,0.61}
\definecolor {gray62}              {rgb}{0.62,0.62,0.62}
\definecolor {gray63}              {rgb}{0.63,0.63,0.63}
\definecolor {gray64}              {rgb}{0.64,0.64,0.64}
\definecolor {gray65}              {rgb}{0.65,0.65,0.65}
\definecolor {gray66}              {rgb}{0.66,0.66,0.66}
\definecolor {gray67}              {rgb}{0.67,0.67,0.67}
\definecolor {gray68}              {rgb}{0.68,0.68,0.68}
\definecolor {gray69}              {rgb}{0.69,0.69,0.69}
\definecolor {gray70}              {rgb}{0.70,0.70,0.70}
\definecolor {gray71}              {rgb}{0.71,0.71,0.71}
\definecolor {gray72}              {rgb}{0.72,0.72,0.72}
\definecolor {gray73}              {rgb}{0.73,0.73,0.73}
\definecolor {gray74}              {rgb}{0.74,0.74,0.74}
\definecolor {gray75}              {rgb}{0.75,0.75,0.75}
\definecolor {gray76}              {rgb}{0.76,0.76,0.76}
\definecolor {gray77}              {rgb}{0.77,0.77,0.77}
\definecolor {gray78}              {rgb}{0.78,0.78,0.78}
\definecolor {gray79}              {rgb}{0.79,0.79,0.79}
\definecolor {gray80}              {rgb}{0.80,0.80,0.80}
\definecolor {gray81}              {rgb}{0.81,0.81,0.81}
\definecolor {gray82}              {rgb}{0.82,0.82,0.82}
\definecolor {gray83}              {rgb}{0.83,0.83,0.83}
\definecolor {gray84}              {rgb}{0.84,0.84,0.84}
\definecolor {gray85}              {rgb}{0.85,0.85,0.85}
\definecolor {gray86}              {rgb}{0.86,0.86,0.86}
\definecolor {gray87}              {rgb}{0.87,0.87,0.87}
\definecolor {gray88}              {rgb}{0.88,0.88,0.88}
\definecolor {gray89}              {rgb}{0.89,0.89,0.89}
\definecolor {gray90}              {rgb}{0.90,0.90,0.90}
\definecolor {gray91}              {rgb}{0.91,0.91,0.91}
\definecolor {gray92}              {rgb}{0.92,0.92,0.92}
\definecolor {gray93}              {rgb}{0.93,0.93,0.93}
\definecolor {gray94}              {rgb}{0.94,0.94,0.94}
\definecolor {gray95}              {rgb}{0.95,0.95,0.95}
\definecolor {gray97}              {rgb}{0.97,0.97,0.97}
\definecolor {gray98}              {rgb}{0.98,0.98,0.98}
\definecolor {gray99}              {rgb}{0.99,0.99,0.99}
\definecolor {darkgrey}            {rgb}{0.66,0.66,0.66}

\newcommand{\new}[1]{{\color{blue}{#1}}\/}

\newcommand{\TODO}[1]{{}}

\newcommand{\ignore}[1]{}
\newcommand{\todo}[1] {}
\newcommand{\RSCHANGE}[1]{{\color{blue}{{#1}}} }
\newcommand{\RSTODO}[1]{{\bf {\color{blue}{{\fbox{RS TODO:} #1}}}} }

\newenvironment{rschange}{\color{blue}}{\normalcolor}
\newenvironment{futureversion}{\color{red}}{\normalcolor}

 \newcommand{\ignoreinshort}[1]{}
 \newcommand{\ignoreinlong}[1]{{\color{blue}{#1}}}

\def\makenewenumerate#1#2{%
\newcounter{cnt#1}
\newenvironment{#1}%
{\begin{list}{\makebox[0pt][r]{#2}}%
{\setlength{\itemsep}{0pt}%
 \setlength{\parsep}{.2em}%
 \setlength{\leftmargin}{1.5em}%
 \setlength{\labelwidth}{.4em}%
 \usecounter{cnt#1}}}
{\end{list}}}

\makenewenumerate{myenumerate}{\arabic{cntmyenumerate}.}

\makenewenumerate{renumerate}{\rm(\roman{cntrenumerate})}
\makenewenumerate{renumerateprime}{\rm(\roman{cntrenumerateprime}$'$)}
\makenewenumerate{renumeratesecond}{\rm(\roman{cntrenumeratesecond}$''$)}

\makenewenumerate{aenumerate}{({\it\alph{cntaenumerate}})}
\makenewenumerate{aenumerateprime}{({\it\alph{cntaenumerateprime}$'$})}
\makenewenumerate{aenumeratesecond}{({\it\alph{cntaenumeratesecond}$''$})}

\def\newplaintheorem#1#2{%
\newtheorem{#1plain}{#2}[section]%
\newenvironment{#1}{\begin{#1plain}\rm }{\end{#1plain}}}

\newplaintheorem{notation}{Notation}
\newplaintheorem{cexample}{Counter-Example}

\newcommand{\sref}[1]{\S{}\ref{#1}}
\newcommand{\noi}{\noindent}

\newcommand{\tuple}[1]{\ensuremath{\langle{#1}\rangle}\xspace}
\newcommand{\set}[1]{\ensuremath{\{{#1}\}}\xspace}
\newcommand{\imp}{\ensuremath{\rightarrow}\xspace}
\newcommand{\limp}{\ensuremath{\leftarrow}\xspace}
\renewcommand{\iff}{\ensuremath{\leftrightarrow}\xspace}
\newcommand{\defas}{\ensuremath{\stackrel{\text{\tiny def}}{=}}\xspace}

\newcommand{\pos}{\phantom{\neg}}

\newcommand\cala{\ensuremath{\mathcal{A}}\xspace}
\newcommand\calb{\ensuremath{\mathcal{B}}\xspace}

\newcommand\cald{\ensuremath{\mathcal{D}}\xspace}
\newcommand\cale{\ensuremath{\mathcal{E}}\xspace}

\newcommand\calg{\ensuremath{\mathcal{G}}\xspace}

\newcommand\calm{\ensuremath{\mathcal{M}}\xspace}
\newcommand\caln{\ensuremath{\mathcal{N}}\xspace}

\newcommand\calp{\ensuremath{\mathcal{P}}\xspace}

\newcommand\calr{\ensuremath{\mathcal{R}}\xspace}

\newcommand\calt{\ensuremath{\mathcal{T}}\xspace}

\newcommand\calw{\ensuremath{\mathcal{W}}\xspace}

\renewcommand{\RSCHANGE}[1]{{\color{blue}{{#1}}}}

\renewcommand{\costof}[1]{\ensuremath{{\sf CostOf}(#1)}\xspace}
\newcommand{\addedcostof}[1]{\ensuremath{{\sf AddedCostOf}(#1)}\xspace}

\newcommand{\minaddedcostof}[2]{\ensuremath{{\sf MinAddedCostOf}(#1,#2)}\xspace}

\newcommand{\refinements}[1]{\ensuremath{{\sf Refinements}(#1)}\xspace}
\newcommand{\sources}[1]{\ensuremath{{\sf Sources}(#1)}\xspace}
\newcommand{\target}[1]{\ensuremath{{\sf Target}(#1)}\xspace}

\newcommand{\roots}[1]{\ensuremath{{\sf Roots}(#1)}\xspace}
\newcommand{\leaves}[1]{\ensuremath{{\sf Leaves}(#1)}\xspace}
\newcommand{\internals}[1]{\ensuremath{{\sf Internals}(#1)}\xspace}

\newcommand{\assertions}{\tuple{\calg',\call',\eta}}
\renewcommand{\assertions}{\ensuremath{\eta}\xspace}

\newcommand{\enc}{\ensuremath{\Psi_{\calm}}}

\newcommand{\encaddcostass}{\ensuremath{\Psi_{\tuple{\calm_2|\mu_1,\cost}}}}

\newcommand{\rangeone}{\{E_i\in\cale_2\setminus \cale_1\ |\ \calw_2(E_i)> 0, \mu_2(E_i)=\top\}}
\newcommand{\rangetwo}{\{E_i\in\cale_2\cap \cale_1\ |\ \calw_2(E_i)> 0,\mu_2(E_i)=\top,\mu_1(E_i)=\bot\}}
 
\newcommand{\elementsone}{\ensuremath{\cale_{new}}}
\newcommand{\elementstwo}{\ensuremath{\cale_{common}}}

\renewcommand{\larat}{\ensuremath{\mathcal{LRA}}\xspace}
\renewcommand{\smtlarat}{\ensuremath{\text{SMT}_{\larat}}\xspace}
\renewcommand{\smtlarat}{\smttt{\larat}}

\newcommand{\omttt}[1]{\ensuremath{\text{OMT}(#1)}\xspace}
\newcommand{\omtlarat}{\omttt{\larat}}

\renewcommand{\new}[1]{{\em #1}}

\newcommand{\PGJMTODO}[1]{{\bf {\color{darkviolet}{{\fbox{PG/JM TODO:} #1}}}}}

\renewcommand{\preferred}[2]{\ensuremath{#1\succeq #2}}

\renewcommand{\G}{\ensuremath{\Goal{G}}}
\newcommand{\Gprime}{\ensuremath{\Goal{G}'}}
\newcommand{\Gsecond}{\ensuremath{\Goal{G}"}}

\newcommand{\Goneone}{\ensuremath{\Goal{G}_{11}}}
\newcommand{\Gonetwo}{\ensuremath{\Goal{G}_{12}}}
\newcommand{\Gtwotwo}{\ensuremath{\Goal{G}_{22}}}
\newcommand{\Gtwoone}{\ensuremath{\Goal{G}_{21}}}

\newcommand{\el}{\ensuremath{E}}
\newcommand{\elone}{\ensuremath{E_1}}
\newcommand{\eltwo}{\ensuremath{E_2}}

\newcommand{\ellast}{\ensuremath{E_n}}
\newcommand{\refi}{\ensuremath{R}}
\newcommand{\refione}{\ensuremath{R_1}}
\newcommand{\refitwo}{\ensuremath{R_2}}
\newcommand{\refii}{\ensuremath{R_i}}
\newcommand{\refik}{\ensuremath{R_m}}

\newcommand{\prefers}[2]{\ensuremath{ #1 \overset{\text{\tiny prefer}}{\longrightarrow} #2}}

\newcommand{\numunsatprefs}{\REQ{numUnsatPrefs}}
\newcommand{\numdeniedrequirements}{\REQ{numUnsatRequirements}}
\newcommand{\numsatisfiedtasks}{\REQ{numSatTasks}}

\newcommand{\weight}{\REQ{Weight}}

\renewcommand{\RSCHANGE}[1]{{\color{blue}{{#1}}}}
\renewcommand{\MCCHANGE}[1]{{\color{darkgreen}{{#1}}}}

\renewenvironment{mcchange}{\color{darkgreen}}{\normalcolor}
\renewenvironment{rschange}{\color{blue}}{\normalcolor}
\newenvironment{jmchange}{\color{red}}{\normalcolor}

\newcommand{\longversion}{true}
\ifthenelse{\equal{\longversion}{true}}
{%
  \renewcommand{\ignoreinshort}[1]{{\color{darkviolet}{#1}}}
  \renewcommand{\ignoreinshort}[1]{{#1}}
  \renewcommand{\ignoreinlong}[1]{}
\newcommand{\fakesubsubsection}[1]{\paragraph{#1}}
\renewcommand{\fakesubsubsection}[1]{\medskip\noindent{\bf #1}}
}%
{%
 \renewcommand{\ignoreinshort}[1]{}
 \renewcommand{\ignoreinlong}[1]{{#1}}
\newcommand{\fakesubsubsection}[1]{\paragraph{#1}}
}

\renewcommand{\MCCHANGE}[1]{{\color{black}{{#1}}}}

\begin{document}

\title{Multi-Objective
Reasoning \\
with Constrained Goal Models
\thanks{This research was partially supported by the ERC advanced grant
267856, `Lucretius: Foundations for Software Evolution'. 
\ignore{Roberto Sebastiani is also supported in part by
{Semiconductor Research Corporation} (SRC)
  under
GRC Research Project 2012-TJ-2266 WOLF.}
}
}



\author{Chi Mai Nguyen \and
Roberto Sebastiani \\ \and
Paolo Giorgini \and
John Mylopoulos 
}


\institute{
 Chi Mai Nguyen \at
 \email{chimai.nguyen@unitn.it} 
 \and Roberto Sebastiani \at
 \email{roberto.sebastiani@unitn.it}
 \and Paolo Giorgini \at
 \email{paolo.giorgini@disi.unitn.it}
 \and John Mylopoulos \at
 \email{jm@cs.toronto.edu}
 }

\date{Received: date / Accepted: date}

\maketitle

\begin{abstract}
 Goal models have been widely used in Computer Science to 
represent software requirements, business objectives, and 
design qualities. Existing goal modelling techniques, however, 
have shown limitations of expressiveness and/or tractability 
in coping with complex real-world problems. In this work, we 
exploit advances in automated reasoning technologies, 
notably Satisfiability and Optimization Modulo Theories (SMT/OMT), 
and we propose and formalize: 
(i) an extended modelling language for goals, namely 
the \new{Constrained Goal Model (CGM)}, which makes 
explicit the notion of \new{goal refinement} and of
\new{domain assumption}, 
allows for expressing \new{preferences}
 between goals and refinements, and
allows 
for associating \new{numerical attributes} to goals and refinements for 
defining 
\new{constraints} and \new{optimization goals} over multiple \new{objective functions},
refinements and their numerical attributes;
(ii) a novel set of automated reasoning functionalities over CGMs, 
allowing for automatically generating suitable refinements of input 
CGMs, under user-specified assumptions and constraints, that also 
maximize preferences and
optimize given objective functions. 
We have implemented 
these modelling and reasoning functionalities in a tool, 
{named CGM-Tool},
using the OMT solver OptiMathSAT 
as automated reasoning backend.
Moreover, we have conducted an experimental evaluation on large CGMs to support the
claim that our proposal scales well for goal models with thousands of elements.

\end{abstract}

\newcommand{\secinf}[2]{[{\bf #1}{#2p}]}
\section{Introduction}
\label{sec:intro}
The concept of goal has long been used as useful abstraction 
in many areas of computer science,  for example 
artificial intelligence planning%
{~\cite{newell63}}, agent-based systems%
{~\cite{BDIAgents95}}, and  knowledge management%
{~\cite{Jarvis01}}. More recently, software engineering has also 
been using goals to model requirements for software systems, 
business objectives for enterprises, and design qualities
{~\cite{anton96,anton98,dardenne93,Lamsweerde01,Giorgini04formalreasoning}.}

Goal-oriented requirements engineering approaches have gained 
popularity for a number of significant 
benefits in conceptualizing and analyzing requirements
~\cite{Lamsweerde01}.  Goal models provide a broader system engineering 
perspective compared to traditional requirements engineering methods, 
a precise criterion for completeness of the requirements analysis process, 
and rationale for requirements specification, 
as well as automated support for early requirements analysis.
Moreover, goal models are useful in explaining requirements to stakeholders, 
and goal refinements offer an accessible level of abstraction for 
validating choices among alternative designs.

Current goal modelling and reasoning techniques, however, have
limitations {with respect to expressiveness and/or scalability.}
{Among leading approaches for goal modelling, KAOS 
offers a very expressive modelling language but reasoning isn't scalable 
(in fact, it is undecidable). i*, on the other hand, is missing constructs 
such as preferences, priorities and optimization goals. Although more 
recent proposals, such as Techne ~\cite{JuretaBEM10,Liaskos10} 
offer expressive extensions to goal models, they still lack some 
features of our proposal, notably optimization goals, 
and also lack scalable reasoning facilities.

As a result of these deficiencies, no goal modelling framework can 
express goals such as ``Select which new requirements to implement 
for the next release, such as to optimize customer value while 
maintaining costs below some threshold" \new{and} be able to 
reason about it and generate a specification/solution for it. 
As another example, consider a situation where a goal model 
changes and a new specification/solution needs to be generated 
for the new goal model. In this case, the new specification may be 
required to fulfill the evolution goal ``Minimize implementation effort" 
or ``Maximize user familiarity by changing as little as possible the new 
functionality of the system relative to the old one". 
{(For the latter case, see also \cite{Nguyen16}.)} In both cases we 
are dealing with requirements that are beyond the state-of-the-art for 
goal modelling and reasoning. As we will discuss in \sref{sec:goalmodels_example}, our 
proposal can accommodate such requirements both with respect to 
modelling and scalable reasoning.}



{We are interested in advancing the state-of-the-art in goal models
and reasoning by proposing a more expressive modelling languages
that encompasses many of the 
modelling constructs proposed in the literature, and at the same 
time offers sound, complete and tractable reasoning facilities. 
We are aiming for  a goal modelling language in the spirit of  
Sebastiani et al. \cite{sebastiani_caise04}, rather than a social dependencies modelling language,
 such as i*. To accomplish this,}
we exploit advances in
automated reasoning technologies,
notably {\em Satisfiability Modulo Theories (SMT)}~\cite{barrettsst09} and 
{\em Optimization Modulo Theories
  (OMT)}~\cite{sebastiani15_optimathsat}, to propose and formalize
an extended notion of goal model, 
namely \new{Constrained Goal Model (CGM)}.  
%
%

{
CGMs treat (AND/OR) refinements as first class citizens allowing
associated constraints, such as {Boolean formulas or SMT/OMT
  formulas. For instance, when modelling a meeting scheduling system,
we may want to express the fact that, to fulfill the nice-to-have requirement of 
keeping the scheduling fast enough (e.g., strictly less than 5 hours)
we cannot afford both the
time-consuming tasks of performing the schedule manually (3 hours)
and of calling the participant one-by-one by phone  (2 hours).
CGMs provide user-friendly constructs by which
 we can encode constraints like this,  either by adding 
Boolean formulas on the propositions 
which label such requirement and tasks%
,
or by associating to those propositions 
numerical variables 
and by adding SMT formulas encoding 
 mixed Boolean-arithmetical constraints on those variables and propositions.
(See \sref{sec:goalmodels_example}.)
To the best of our knowledge, this was not possible with previous goal
modelling techniques, including that in \cite{sebastiani_caise04}.
}

\ignore{
Consider the case in which keeping the schedule fast is one possible
requirement (FastSchedule), so that you cannot afford both the
time-consuming tasks of performing the schedule manually
(ScheduleManually) and of calling the participant one-by-one by phone
(CallParticipants). This can be expressed by the simple Boolean
constraint (2)  in section 3.2: 
FastSchedule implies  (not (ScheduleManually and CallParticipants))
which cannot be expressed by positive/negative contribution and
conflict edges of [Sebastiani04].  
}
At the
same time, the CGM tool we developed can cope with goal models an
order of magnitude beyond what has been reported in the literature in
most cases. In some cases involving optimization goals, e.g.,
``minimize development costs for the next release of software product
S'', the CGM tool performs more modestly, but can still handle models
of size in the hundreds of elements. 


The main contributions of this work include:}

\begin{enumerate}[I.]
\item {An integration} within one modelling framework {of} constructs that have been proposed in the literature in a piecemeal fashion, specifically,  
	\begin{enumerate}[(i)]
	\item Allow for explicit labelling of goal refinements with Boolean propositions that can be interactively/automatically reasoned upon;
	\item Provide an explicit representation of domain assumptions to represent preconditions to goals;
	\item Allow for Boolean constraints over goals, domain assumptions and refinements;
	\item Provide a representation of preferences over goals and their refinements, by distinguishing between mandatory and nice-to-have 	
	       requirements and by assigning preference weights (i.e., penalties/rewards) to goals and domain assumptions. Alternatively, preferences can 
	       be expressed explicitly by setting binary preference relations between pairs of goals or pairs of refinements;
	\item Assign numerical attributes (e.g., resources like cost, worktime, and room) to goals and/or refinements and define constraints and multiple 
	      objective functions over goals, refinements and their numerical attributes.
	\item Define optimization goals over numerical attributes, such as cost or customer value;
	\end{enumerate}
	
\item Fully support automated reasoning over {CGMs} that is both sound and complete, i.e., returns only solutions that are consistent with CGM semantics, and all such solutions;

\item Establish that reasoning with CGM models is scalable with models including thousands of elements.
\end{enumerate}
\ignore{
\begin{enumerate}[(i)]
\item
Allow for explicit labelling of \new{goal refinements} with Boolean propositions that can be
  interactively/automatically reasoned upon; 

\item
Provide an explicit representation of \new{domain assumptions} to represent
  preconditions to goals;

\item
Allow for  \new{Boolean constraints} over goals, domain assumptions
and refinements;

\item
Provide a representation of \new{preferences} over goals and their
refinements,
 by distinguishing between mandatory and nice-to-have
requirements and by assigning preference \new{weights} 
(i.e., penalties/rewards) 
to goals and domain assumptions. 
{
Alternatively, preferences can be expressed  explicitly  by
setting \new{binary
preference relations} between pairs of goals or pairs of refinements;}

\item
Assign \new{numerical attributes} (e.g., resources like 
cost, worktime, and room) to goals and/or refinements and define
\new{constraints} and multiple \new{objective functions} over goals,
refinements and their numerical attributes.


\end{enumerate}
  }


\noindent
Taking advantage of CGMs' formal semantics and
the expressiveness and efficiency of current SMT
 and OMT solvers, we also provide a set of automated
reasoning functionalities on CGMs. {Specifically}, on a given CGM, our
approach allows for: 

\begin{itemize}
\item[(a)] the automatic check of the CGM's realizability 
{(i.e., check if the goal model has any solution)}; 
\item[(b)] the interactive/automatic search for realizations; 
\item[(c)] the automatic search for the ``best" realization in terms
of penalties/rewards {and/or of user-defined preferences;}
\item[(d)] the automatic search for the  realization(s) which optimize 
given objective functions.
 
\end{itemize}
Our approach is
implemented as a tool (CGM-Tool), a standalone java
application based on the Eclipse RCP engine. The tool offers
functionalities to create CGM models as graphical diagrams and to
explore alternatives scenarios running automated reasoning
techniques. CGM-Tool uses the SMT/OMT solver OptiMathSAT%
~{\cite{sebastiani15_optimathsat,st_tacas15,st_cav15},
  which is built on top of the SMT solver \mathsatfive \cite{mathsat5_tacas13},}
as automated reasoning backend.~\footnote{{The OMT solver OptiMathSAT
  can be used also as an SMT solver if no objective function is set: in
  such case it works as a wrapper of \mathsatfive.}} 

The structure of the paper is as follows: 
\sref{sec:background} provides {a succinct account of} 
necessary background  on goal
modelling and on SMT/OMT;
\sref{sec:goalmodels_example} introduces the notion of CGM
through an example;
\sref{sec:goalmodels} introduces the syntax and semantics of CGMs;
\sref{sec:functionalities} presents the set of automated reasoning functionalities for CGMs;
\sref{sec:implementation} gives a quick overview of our tool
based on the presented approach;
{\sref{sec:expeval} provides an experimental evaluation of the
performances of our tool on large CGMs, showing that the
approach scales well with respect to CGM size;}
\sref{sec:related} gives overview of related work,
while in \sref{sec:concl} we draw conclusions and present future
research challenges.

\section{Background}
\label{sec:background}

Our research baseline consists of our previous work on qualitative
goal models and of Satisfiability and Optimization Modulo Theories (SMT
and OMT respectively). 
{Our aim in this section is to introduce the necessary 
background notions on the these topics, so that the reader can follow the
narrative in subsequent sections.}
{As prerequisite knowledge, we assume only that the reader is
familiar with the syntax and semantics of standard Boolean logic and
of linear arithmetic over the rationals.}

\subsection{Goal Models.}
\label{sec:background_goalmodels}
Qualitative goal models are introduced in~\cite{Mylopoulos92}, where the concept of goal 
is used to represent respectively a functional and non-functional requirement in terms of a
proposition. A goal can be refined by means of AND/OR refinement relationships and
qualitative evidence (strong and weak) for/against the fulfillment of a goal is provided by
contribution links labelled $+, -$ etc. In~\cite{Giorgini04formalreasoning}, goal models are formalized by replacing
each proposition $g$, standing for a goal, by four propositions ($FS_g$, $PS_g$, $PD_g$, $FD_g$) 
representing full (and partial) evidence for the satisfaction/denial of $g$. A traditional
implication such as $(p \wedge q) \Implies r$ is then translated into a series of implications connecting
these new symbols, including $(FS_p \wedge FS_q) \Implies FS_r$, $(PS_p \wedge PS_q) \Implies
PS_r$, as well as $FD_p \Implies FD_r$, $FD_q \Implies FD_r$, etc. The conflict between goals $a$ and $b$ is captured by axioms of the form $FS_a\Implies FD_b$, and it is consistent to have both $FS_a$ and $FD_a$
evaluated to true at the same time. As a result, even though the semantics of a goal model is a classical propositional theory, inconsistency does not result in everything being true. In fact, a predicate $g$ can be assigned a subset of truth values $\{FS, PS,
FD, PD\}$.   

~\cite{sebastiani_caise04} extended the approach further by including axioms
for avoiding conflicts of the form $FS_a \wedge FD_a$. The approach
recognized the need to formalize goal models so as to automatically
evaluate the satisfiability of goals. These goal models, however, do not incorporate the notion of conflict as inconsistency, they do not include concepts other than goals, cannot distinguish ``nice-to-have'' from mandatory requirements and have no notion of a robust
solution, i.e. solution without "conflict", where a goal can not be
(fully or partially) denied and (respectively, fully or partially) satisfied at the same
time.

\subsection{Satisfiability and Optimization Modulo Theories.}
\new{Satisfiability Modulo Theories (SMT)} is the problem of deciding
the satisfiability of a quantifier-free first-order
formula $\Phi$ with respect to some decidable theory \T (see
\cite{sebastiani07,barrettsst09}).
In this paper, we focus on the theory of \new{linear arithmetic over
the rationals, \larat}: \smtlarat is the problem of checking the
satisfiability of a formula $\Phi$ consisting in atomic propositions
$A_1,A_2,...$ and linear-arithmetic constraints over rational variables
like ``$(2.1 x_1 -3.4
x_2 + 3.2 x_3 \le 4.2)$'', combined by means of Boolean operators
$\neg,\wedge,\vee,\imp,\iff$.
{(Notice that a Boolean formula is also a
\smtlarat formula, but not vice versa.)}
An \new{\larat-interpretation} $\mu$ is a function
  which assigns truth values to Boolean atoms and rational values to
  numerical variables; $\mu$ \new{satisfies} $\Phi$ in \larat{},
written
  ``$\mu\models\Phi$'' --aka, $\mu$
  is a \new{solution} for $\Phi$ in \larat--  iff $\mu$ makes the formula $\Phi$ evaluate to
  true; $\Phi$ is \larat-satisfiable iff it has at least one
  \larat-interpretation $\mu$ s.t. $\mu\models\Phi$.

An \new{Optimization Modulo
Theories over \larat (\omtlarat)} problem \tuple{\Phi,\tuple{obj_1,...,obj_k}} is the problem of finding
solution(s) to an \smtlarat formula $\Phi$ which optimize the
 rational-valued objective functions $obj_1,...,obj_k$, either
 singularly or lexicographically 
{\cite{nieuwenhuis_sat06,st-ijcar12,sebastiani15_optimathsat,st_tacas15}}%
).~%
{{A solution \new{optimizes lexicographically}
\tuple{obj_1,...,obj_k} if it optimizes $obj_1$ and, if more than 
one such $obj_1$-optimum solutions exists, it also optimizes $obj_2$,..., and so on.}}

Very efficient \smtlarat and \omtlarat solvers are available, 
which combine the power of modern SAT
solvers with dedicated linear-programming decision and minimization
procedures (see
{\cite{sebastiani07,barrettsst09,mathsat5_tacas13,nieuwenhuis_sat06,st-ijcar12,sebastiani15_optimathsat,st_tacas15,st_cav15}}%
).
For instance, in the empirical evaluation reported in
  \cite{sebastiani15_optimathsat}
 the \omtlarat solver OptiMathSAT \cite{sebastiani15_optimathsat,st_cav15} was able to
handle optimization problems with up to thousands Boolean/rational
variables  in less than 10 minutes each.

\label{sec:background_sat}

\section{Constrained Goal Models}
\label{sec:goalmodels_example}

\begin{figure*}
\centering 
\includegraphics[height=0.95\textheight, width=\textwidth]{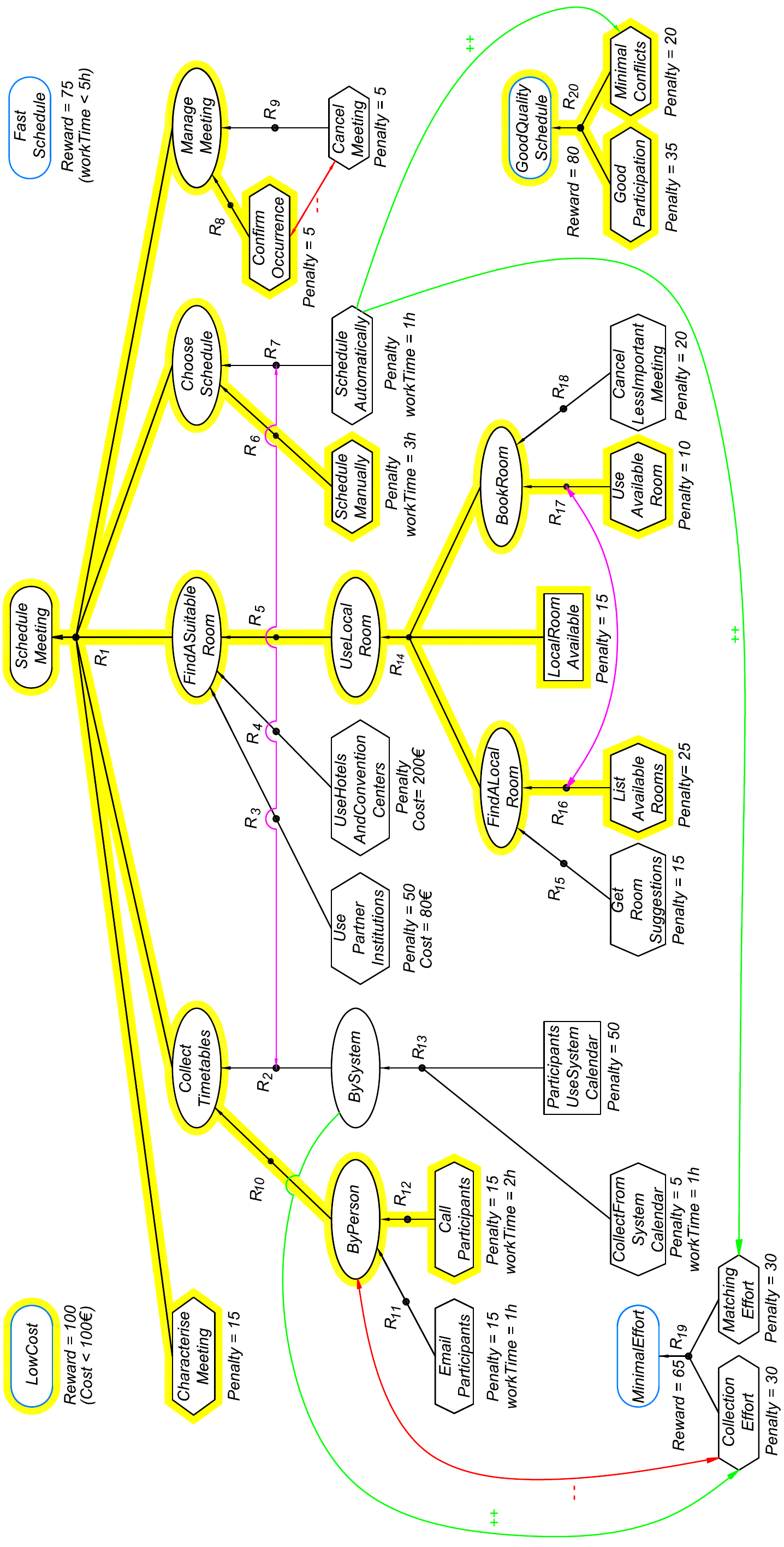}
\caption{An example of a CGM with one of its realizations. 
{Here and elsewhere,}
{round-corner rectangles are requirements; ovals are intermediate goals; hexagons are tasks; rectangles are domain assumptions.
Labeled bullets at the merging point of a group of edges are
refinements; contribution edges are labeled with ++; conflict edges
are labeled with --; and refinement bindings are edges between
refinements only. Values of numerical attributes associated with the
elements and their {positive prerequisite formulas} are written below
the respective elements.} 
The realization is highlighted in yellow, and the denied
  elements \MCCHANGE{are visible but are not highlighted.}
\label{figGMEx}
}
\end{figure*}

\begin{figure*}
\centering 
\includegraphics[height=0.95\textheight, width=\textwidth]{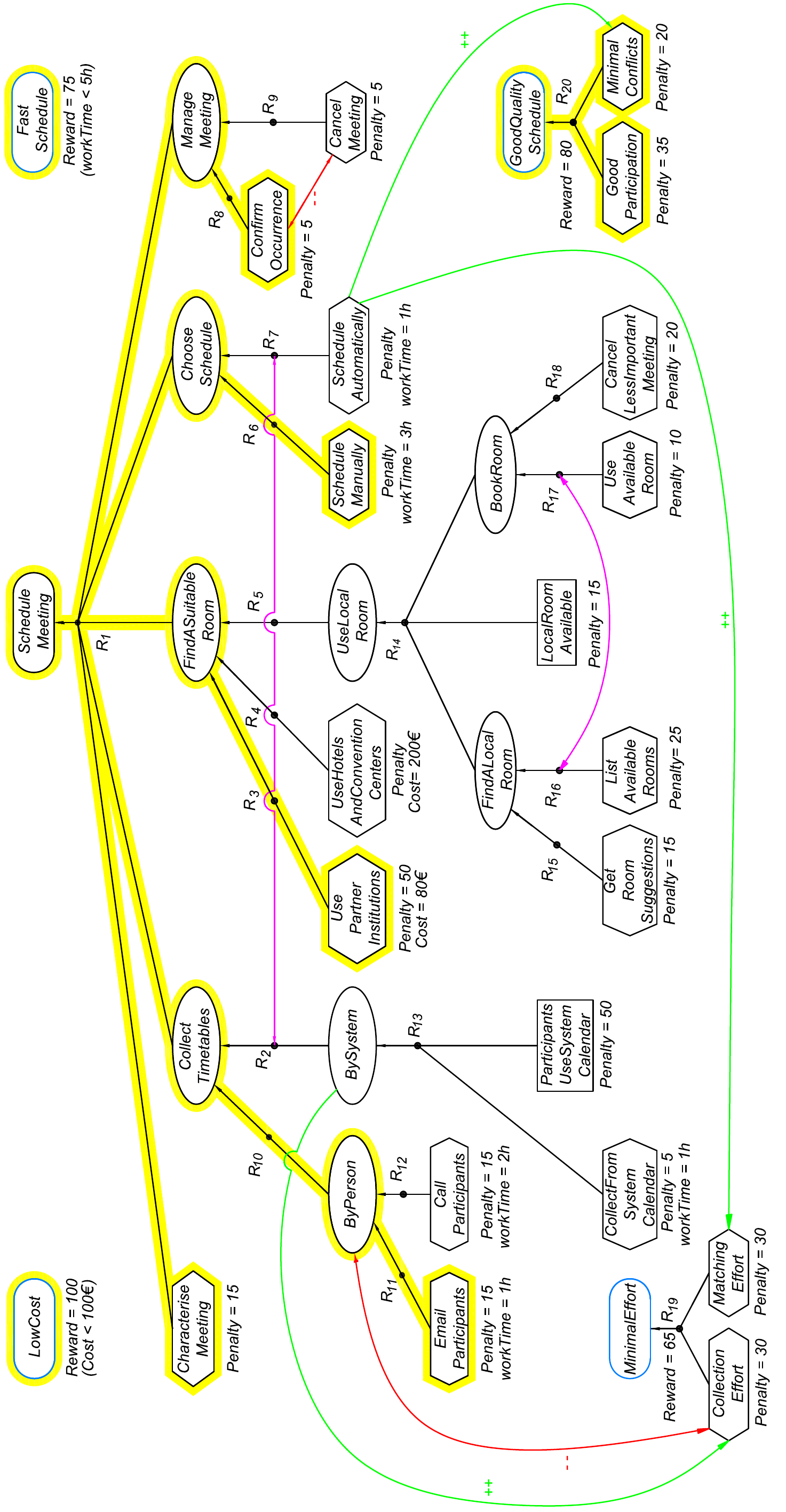}
\caption{{A CGM and its realization with minimized \REQ{Weight}.}
The realization is highlighted in yellow, and the denied
  elements  \MCCHANGE{are visible but are not highlighted.}
\label{figCGMRelization}
}
\end{figure*}

\begin{figure*}
\centering 
\includegraphics[height=0.95\textheight, width=\textwidth]{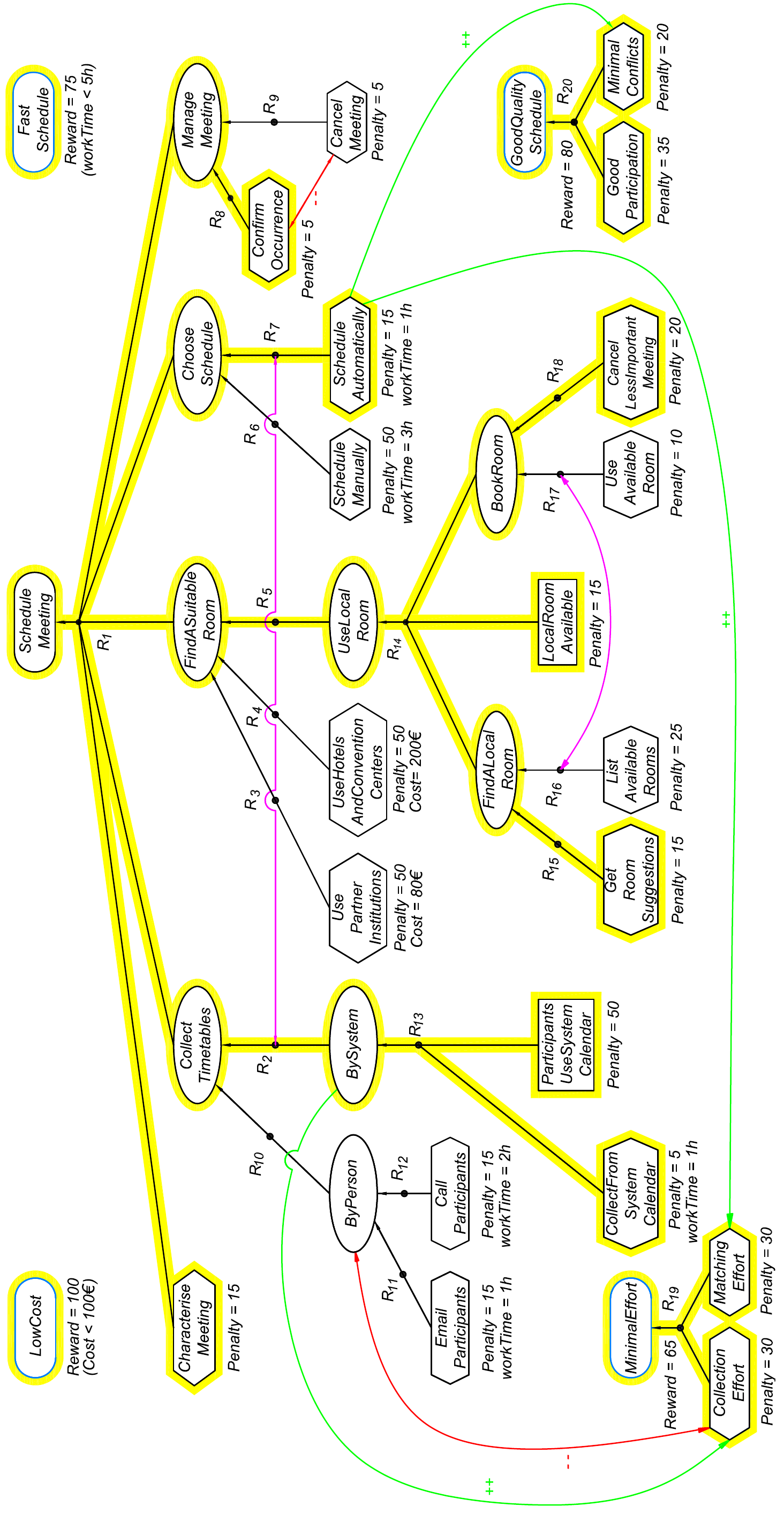}
\caption{{A CGM and its realization with minimized lexicographically
  \tuple{\REQ{Weight}, \REQ{workTime}, \REQ{cost}}, or minimized
lexicographically \tuple{\REQ{Weight},\numunsatprefs{}}.
}
The realization is highlighted in yellow, and the denied
  elements  \MCCHANGE{are visible but are not highlighted.}
\label{figCGMLex}
}
\end{figure*}

%
  The narrative of the next 3 sections {is} in line with the
  following schema. 
 
  In this section (\sref{sec:goalmodels_example}), we introduce the notions
  of constrained goal model {(CGM), and of} realization of a CGM;
{we also} present the automated-reasoning functionalities of our
  CGM-Tool through a meeting scheduling example
  (Figure~\ref{figGMEx}), \MCCHANGE{without getting into the formal details yet.}

In \sref{sec:goalmodels} we present the abstract syntax and semantics of 
CGMs, defining formally the building blocks of a CGM and of its
realizations, {to} which the reader {has already
been introduced informally in} \sref{sec:goalmodels_example}. 

In \sref{sec:functionalities} 
we describe how to {support} automated
  reasoning functionalities on CGMs by encoding them into SMT and OMT.
We first show how to encode a CGM \calm into a \smtlarat formula
  $\Psi_\calm$, so that the search for an optimum realization of \calm 
reduces to an \omtlarat problem over the formula $\Psi_\calm$, which is then
  fed to an OMT solver. Then we present the reasoning functionalities
  over CGMs we have implemented on top of our OMT solver.

\ignore{
We introduce the main ideas of CGMs and the main functionalities of
our CGM-Tool through a meeting scheduling example
(Figure~\ref{figGMEx}).
}

%
\subsection{{The CGM Backbone: Goals, Refinements, and Domain Assumptions.}}
We model the requirements for a meeting scheduling system, including
the functional requirement \REQ{Schedule\-Meeting}, 
as well as non-functional/quality requirements \REQ{Low\-Cost},  ~\REQ{Fast\-Schedule}, 
\REQ{Minimal\-Effort} and \REQ{Good\-Quality\-Schedule}. 
They are represented as root goals.

Notationally, round-corner rectangles (e.g., \REQ{Schedule\-Meeting})
are root goals, representing stakeholder \new{requirements}; 
ovals (e.g. \REQ{Collect\-Timetables})
are \new{intermediate goals};
\new{hexagons} (e.g. \REQ{Characterise\-Meeting})
are \new{tasks}, i.e.  non-root leaf goals;
rectangles (e.g., \REQ{Participants\-Use\-System\-Calendar}) are
\new{domain assumptions}. 
We call \new{elements} both goals and domain assumptions.
Labeled bullets at the merging point of the edges connecting a group
of source elements to a target element are \new{refinements} (e.g.,
~\Refines{(\REQ{Good\-Participation},\REQ{Minimal\-Conflict})}{
  \REQ{Good Quality Schedule}}{20}), while the $R_i$s denote their
labels.~%

\begin{remark}
Unlike {previous goal modelling proposals}, refinements are
  explicitly labeled, so that stakeholders can refer to them
  in relations, constraints and preferences. 
(This fact will be eventually discussed  with more details.)
The label of a refinement can be omitted when {there is no need to
  refer to it explicitly.}
\end{remark}

%
Intuitively, requirements represent desired states of affairs we want
the system-to-be to achieve (either mandatorily or {preferrably}); they are
progressively refined into intermediate goals, until the process
produces actionable goals (tasks) that need no further decomposition
and can be executed; domain assumptions are propositions about the
domain that need to hold for a goal refinement to work.  {{Refinements}
are used to represent alternatives of how to achieve a non-leaf
element, i.e., a refinement of an element represents one of
  the alternative of sub-elements that are necessary to
  achieve it}.

The principal aim of the CGM in Figure~\ref{figGMEx}
is to achieve the requirement \REQ{Schedule\-Meeting}, 
which is  \new{mandatory}. 
{(A requirement is set to
 be mandatory by means of user assertions, see below.)}
\REQ{Schedule\-Meeting} has only one candidate refinement $R_1$, 
consisting in five sub-goals: \REQ{Characterise\-Meeting}, \REQ{Collect\-Timetables}, 
\REQ{Find\-A\-Suitable\-Room}, \REQ{Choose\-Schedule}, and
~\REQ{Manage\-Meeting}. Since $R_1$ is the only refinement of the
requirement, all these sub-goals must be satisfied in order to
satisfy it. 
There may be more than one way to refine an element;
e.g., \REQ{Collect\-Timetables} is further refined either by
$R_{10}$ into the single goal \REQ{By\-Person} or by 
$R_{2}$ into the single goal \REQ{By\-System}.
{Similarly, \REQ{Find\-A\-Suitable\-Room} and 
\REQ{Choose\-Schedule} have three and two possible refinements respectively.}
The subgoals are further refined until they
reach the level of  domain assumptions and tasks.

{The {requirements that} are not set to be mandatory are 
 {``\new{nice-to-have}''} ones, 
like 
\REQ{Low\-Cost}, \REQ{Minimal\-Effort}, \REQ{Fast\-Schedule}, and
\REQ{Good\-Quality\-Schedule} (in blue in  Figure~\ref{figGMEx}).
They are requirements that we would like to fulfill with our solution,
 provided they do not conflict with other requirements.}

\subsection{{Boolean Constraints: Relation Edges, Boolean Formulas and User Assertions.}}
Importantly, in a CGM, elements and refinements are enriched by user-defined
\new{{Boolean} constraints}, which can be expressed either graphically 
as  \new{relation edges},
 or textually as \new{Boolean or \smtlarat{} formulas}{, or as 
\new{user assertions}}.

\fakesubsubsection{Relation Edges.}
We have three kinds of relation edges.
\new{Contribution edges} ``\contributes{E_i}{E_j}'' between elements  (in green in Figure~\ref{figGMEx}),
like ``\contributes{\REQ{Schedule\-Automatically}}{\REQ{Minimal\-Conflicts}}'', 
mean that if the source element $E_i$ is satisfied, then 
also the target element $E_j$ must be satisfied (but not vice versa). 
\new{Conflict edges} ``\conflict{E_i}{E_j}'' between elements (in red), like 
``\conflict{\REQ{Confirm\-Occurrence}}{\REQ{Cancel\-Meeting}}'',
mean that $E_i$ and $E_j$ cannot be both satisfied. 
\new{Refinement bindings} ``\bind{R_i}{R_j}'' between two refinements
(in purple), like ``\bind{R_2}{R_7}'', are used to state that, if the  target
elements $E_i$ and $E_j$ of the two refinements $R_i$ and $R_j$,
respectively, are both
satisfied, then $E_i$ is refined by $R_i$ if and only if $E_j$ is
refined by $R_j$. Intuitively, this means that the two refinements are
bound, as if they were two different instances of the same global choice.

For instance, in Figure~\ref{figGMEx}, the refinements $R_2$ and $R_7$
are bound because such binding reflects a 
global choice between a manual approach and an automated one.

\fakesubsubsection{Boolean Formulas.}
It is possible to enrich CGMs with Boolean formulas, representing
arbitrary 
constraints on elements and refinements.
Such constraints can be either \new{global} or \new{local to elements or to
refinements}, that is, each goal $\Goal{G}$ can be tagged with   a pair 
of \new{prerequisite} formulas $\Constraint{G}{\phi}$ 
{--called \new{positive} and \new{negative} prerequisite
  formulas respectively--}
so that 
 $\posConstraint{G}{\phi}$ [resp. $\negConstraint{G}{\phi}$] must be
 satisfied  when  $\Goal{G}$ is
satisfied [resp. denied].
(The same holds for each requirement $R$.)

For example, to require that, as a prerequisite for \REQ{Fast\-Schedule},
\REQ{Schedule\-Manually} and \REQ{Call\-Participants} cannot be both satisfied, 
one can add a constraint to the positive 
prerequisite formula of \REQ{Fast\-Schedule}:
\begin{eqnarray}
\label{eq:lowtimelocal}
\posConstraint{~\REQ{Fast\-Schedule}}{\phi} \defas
&& ... 
\wedge \neg (\REQ{Schedule\-Manually} \wedge \REQ{Call\-Participants}),
\end{eqnarray}
or, equivalently, add globally to the CGM the following
Boolean formula: 
\begin{eqnarray}
\label{eq:lowtimeglobal}
\REQ{Fast\-Schedule} & \Implies & \neg (\REQ{Schedule\-Manually} \wedge \REQ{Call\-Participants}).
\end{eqnarray}
\noindent
Notice that there is no way we can express \eqref{eq:lowtimelocal}
or \eqref{eq:lowtimeglobal} with the relation edges above.

\fakesubsubsection{User Assertions.}
With CGM-Tool, one can interactively mark [or unmark] 
requirements as satisfied (true), thus making them mandatory 
(if unmarked, they are nice-to-have ones). In our example
\REQ{Schedule\-Meeting} is asserted as true to make it mandatory, 
which is equivalent to
add globally to the CGM the unary Boolean constraint:
\begin{eqnarray}
  (\REQ{Schedule\-Meeting}).
\end{eqnarray}
Similarly, one can interactively mark/unmark (effortful) tasks as
denied (false).  More generally, one can mark as satisfied or denied every
goal or domain assumption.  We call these marks \new{user assertions},
because they correspond to asserting that an element must be true,
 {i.e., it is part of the solutions we are interested in}, or false,
 {i.e., we are interested in solutions that do not include it}.

Notice that the process of marking/unmarking elements is conceived to
be more {\em interactive} than that of
adding/dropping relation edges or constraints.

{\subsection{Arithmetical Constraints: Numerical Attributes
    and \smtlarat Formulas}}

\fakesubsubsection{Numerical Attributes.}
In addition to Boolean constraints, 
it is also possible to use numerical variables to 
express different numerical attributes of elements 
(such as cost, worktime, space, fuel, etc.) and to add arithmetical
constraints in the form of \smtlarat formulas over such numerical variables.

For example,  suppose we estimate that fulfilling
~\REQ{Use\-Partner\-Institutions} costs 80\euro{}, whereas fulfilling
~\REQ{Use\-Hotels\-And\-Convention\-Centers}
costs 200\euro{}.
With CGM-Tool one can express these facts straightforwardly by adding
a global numerical variable \REQ{cost} to the model;

then, for every element $E$ in the CGM,~%
 %
CGM-Tool automatically generates a numerical
variable $\REQ{cost_E}$ 
representing the attribute \REQ{cost} of the element $E$,
it adds the following
default
 global constraint and prerequisite formulas:
\begin{eqnarray}
\label{eq:globalsum}
&& (\REQ{cost} = \sum_{E}\REQ{cost_{E}}),%
\\
\label{eq:globalposdefault}
\mbox{for every element ${E}$,\ \ \ \ }
&&\posConstraint{~\REQ{E}}{\phi} \defas
... \wedge (\REQ{cost_{E}} = 0)\\    
\label{eq:globalnegdefault}
&&\negConstraint{~\REQ{E}}{\phi} \defas
... \wedge (\REQ{cost_{E}} = 0),
  \end{eqnarray}   
{that} set the default value 0 for each \REQ{cost_{E}}.
%
{(Notice that \eqref{eq:globalsum} is a {\em default} global
  constraint: the user is free to define his/her own objective functions.)}
%
Eventually, for the elements $E$ of interest, 
one can set a new value for \REQ{cost_{E}} in case $E$
is satisfied: e.g.,
~$\REQ{cost_{Use\-Partner\-Institutions}} := 80\euro{}$ and
~$\REQ{cost_{Use\-Hotels\-And\-Convention\-Centers}} := 200\euro{}$.
When so, CGM-Tool automatically updates the values in the positive prerequisite
formulas \eqref{eq:globalposdefault}, e.g.:
\begin{eqnarray}
\label{eq:costlocalsmt}
\posConstraint{~\REQ{Use\-Partner\-Institutions}}{\phi} \defas
&&... \wedge (\REQ{cost_{Use\-Partner\-Institutions}} = 80)\\
\nonumber
\posConstraint{~\REQ{Use\-Hotels\-And\-Convention\-Centers}}{\phi} \defas
&& ... \wedge (\REQ{cost_{Use\-Hotels\-And\-Convention\-Centers}} = 200),
\end{eqnarray} 
whereas the corresponding constraint \eqref{eq:globalnegdefault} is not
changed. Similarly, one can set a new
value for \REQ{cost_{E}} in case $E$
is denied by updating the values in the negative prerequisite
formulas \eqref{eq:globalnegdefault}.

\begin{remark}
 Notationally, 
we use variables and formulas indexed by the 
element they belong to 
(like, e.g., $\REQ{cost_{Use\-Partner\-Institutions}}$ and
$\posConstraint{~\REQ{Use\-Partner\-Institutions}}{\phi}$)
rather than attribute variables and formulas of the elements
in an object-oriented notation
(like, e.g., $\REQ{Use\-Partner\-Institutions.cost}$ and $\REQ{Use\-Partner\-Institutions.\phi^+}$)
because they are more suitable to be used within the \smtlarat{} encodings 
(\sref{sec:goalmodels} and \sref{sec:functionalities}).
\ignore{
{E.g., the formula 
$$\posConstraint{~\REQ{Use\-Hotels\-And\-Convention\-Centers}}{\phi}
\imp
(\REQ{cost_{Use\-Hotels\-And\-Convention\-Centers.cost}}=200)$$
is more readable than
$$\REQ{Use\-Hotels\-And\-Convention\-Centers.\Phi^+}
\imp
(\REQ{Use\-Hotels\-And\-Convention\-Centers.cost}=200)$$}
}
\end{remark}

\fakesubsubsection{\smtlarat{} Formulas.}
Suppose that, in order to achieve the nice-to-have requirement \REQ{Low\-Cost}, 
we need to have a total cost smaller than 100\euro{}. This can be expressed
by adding to \REQ{Low\-Cost} the prerequisite formula:
\begin{eqnarray}
\label{eq:lowcostposconstraint}
\posConstraint{~\REQ{Low\-Cost}}{\phi} = \ldots \wedge (\REQ{cost} < 100) .
\end{eqnarray}
Hence, e.g., due to
\eqref{eq:globalsum}-\eqref{eq:lowcostposconstraint}, 
\REQ{Low\-Cost} and \REQ{Use\-Hotels\-And\-Convention\-Centers} 
cannot be both satisfied, matching the intuition that the latter is
too expensive to comply to the nice-to-have \REQ{Low\-Cost} requirement.

Similarly to \REQ{cost}, 
one can introduce, e.g.,  another global numerical attribute \REQ{workTime} 
to reason on working time, and 
estimate, e.g.,  that the total working time for \REQ{Schedule\-Manually},
\REQ{Schedule\-Automatically}, \REQ{Email\-Participants}, \REQ{Call\-Participants}, 
\REQ{Collect\-From\-System\-Calendar} are 3, 1, 1, 2, and 1
hour(s), respectively, and state that the nice-to-have
requirement 
 \REQ{Fast\-Schedule} must require a global time smaller than 5 hours.
As a result of this process, the system will produce the following
constraints. 
\begin{eqnarray}
\label{eq:timelocalconstraint}
&&(\REQ{workTime} =  \sum_{E}\REQ{workTime_{E}})\\
\posConstraint{~\REQ{Fast\-Schedule}}{\phi} &\defas & ... \wedge (\REQ{workTime} <  5)\\
\posConstraint{~\REQ{Schedule\-Manually}}{\phi} 
&\defas & ... \wedge (\REQ{workTime_{Schedule\-Manually}} = 3)\\
\nonumber
\posConstraint{~\REQ{Schedule\-Automatically}}{\phi} 
&\defas & ... \wedge (\REQ{workTime_{Schedule\-Automatically}} = 1)\\
\nonumber
\posConstraint{~\REQ{Email\-Participants}}{\phi} 
&\defas &... \wedge (\REQ{workTime_{Email\-Participants}} = 1) \\
\nonumber
\posConstraint{~\REQ{Call\-Participants}}{\phi} 
&\defas &... \wedge (\REQ{workTime_{Call\-Participants}} = 2) \\
\nonumber
\posConstraint{~\REQ{Collect\-From\-System\-Calendar}}{\phi} 
&\defas &... \wedge (\REQ{workTime_{Collect\-From\-System\-Calendar}} = 1),
\end{eqnarray}
plus the corresponding negative prerequisite formula, which force
the corresponding numerical attributes to be zero.

As with the previous case, e.g., 
the arithmetic constraints      make the combination of 
\REQ{Schedule\-Manually} and \REQ{Call\-Participants} 
incompatible with the nice-to-have requirement 
\REQ{Fast\-Schedule}.

Notice that one can build combinations of numerical attributes.
E.g., 
 if labor cost is $35\euro/hour$, then one can redefine
\REQ{cost} as 
$(\REQ{cost} = \sum_{E}\REQ{cost_{E}}+35\cdot\REQ{workTime})$,
or introduce  a new global variable \REQ{totalCost} as 
$(\REQ{totalCost} = \REQ{cost}+35\cdot\REQ{workTime})$.

\begin{remark}
Although the nice-to-have requirements \REQ{Low\-Cost} and
\REQ{Fast\-Schedule} look isolated in Figure~\ref{figGMEx}, they are
implicitly linked to the rest of the CGM by means of arithmetic 
constraints on the numerical variables \REQ{cost} and
\REQ{workTime} respectively, which implicitly imply Boolean constraints
like: 
\begin{eqnarray}
  \label{eq:implicitconstraints1}
  \REQ{Low\-Cost} &\imp& \neg \REQ{UseHotelsAndConventionCenters} 
\\
  \label{eq:implicitconstraints2}
  \REQ{Fast\-Schedule} &\imp& 
\neg (
\REQ{Schedule\-Manually} \wedge
\REQ{Call\-Participants}
)
\\
  \label{eq:implicitconstraints3}
  \REQ{Fast\-Schedule} &\imp& 
\neg \left (
  \begin{array}{l}
\REQ{Schedule\-Manually}\ \wedge \\
\REQ{Email\-Participants}\ \wedge \\
\REQ{Collect\-From\-System\-Calendar}
  \end{array}
\right )
\\
\nonumber
...
\end{eqnarray}
Nevertheless, there is no need for stakeholders to consider these
implicit constraints, since they are automatically handled 
by the internal \omtlarat reasoning
capabilities of CGM-Tool. 
\end{remark}

\subsection{Realizations of a CGM.}

%
We suppose now that \REQ{Schedule\-Meeting} is marked satisfied by
means of an user assertion
(i.e. it is mandatory) and that no other element is marked.
Then the CGM in Figure~\ref{figGMEx} has more than 20 possible
\new{realizations}. 
The sub-graph which is 
highlighted in yellow describes one of them.

Intuitively, a realization of a CGM under given user assertions 
 represents one of the alternative ways of refining the mandatory
requirements (plus possibly some of the nice-to-have ones) in compliance
with the user assertions and user-defined constraints.  It is a
sub-graph of the CGM including a set of satisfied elements and
refinements: it includes all mandatory requirements, and
[resp. does not include] all elements satisfied [resp. denied] in the
user assertions; for each non-leaf element included, at least one of
its refinement is included; for each refinement included, all its
target elements are included; finally, a realization complies with all
relation edges and with all Boolean and \smtlarat{} constraints.
{(Notationally, in Figures~\ref{figGMEx},
  \ref{figCGMRelization} and \ref{figCGMLex}
a realization is highlighted in yellow, and the denied
  elements are visible but they are not highlighted.)}

Apart from the mandatory requirement, 
the realization in Figure~\ref{figGMEx} allows to achieve 
also the nice-to-have requirements \REQ{Low\-Cost},
\REQ{Good\-Quality\-Schedule}, but not \REQ{Fast\-Schedule} and
\REQ{Minimal\-Effort}; in order to do this,
it  requires accomplishing the tasks 
\REQ{Characterise\-Meeting}, 
\REQ{Call\-Participants},
\REQ{List\-Available\-Rooms},
\REQ{Use\-Available\-Room},
\REQ{Schedule\-Manually}, 
\REQ{Confirm\-Occurrence}, 
\REQ{Good\-Participation}, \REQ{Minimal\-Conflicts},
and it requires  the domain assumption 
\REQ{Local\-Room\-Available}. 

\medskip
\subsection{{Setting Preferences in a CGM.}}
\label{sec:preferences}

In general, a CGM under given user assertions
has many possible realizations. To distinguish among them, 
stakeholders may want to express \new{preferences} on the requirements
 to achieve, on the tasks to accomplish, and on 
elements and refinements to choose. The CGM-Tool 
provides various methods to express preferences:
\begin{itemize}
\item attribute {\em penalties and rewards} for tasks and requirements;
\item introduce  {\em numerical objectives} to optimize;
\item introduce {\em binary preference relations} between elements
  and between refinements.
\end{itemize}
These methods, which are described in what follows, can also be combined. 

\fakesubsubsection{Preferences via Penalties/Rewards.}
{First, stakeholders
can define two numerical attributes called \REQ{Penalty} and \REQ{Reward}, 
then stakeholders can assign \new{penalty} values to tasks
and \new{reward} values to (non-mandatory)
requirements (the numbers ``$\REQ{Penalty} = \ldots$'' and ``$\REQ{Reward} = \ldots$" in Figure~\ref{figGMEx}).}
This implies that requirements [resp. tasks] with higher rewards
[resp. smaller penalties] are preferable.
{Next, stakeholders can define another numerical attribute \REQ{Weight}, that represents 
 the total difference between the penalties and rewards. 
({This} can be {defined} as a global constraint: $(\REQ{Weight} = \REQ{Penalty} - \REQ{Rewards})$.)}
 When a model represents preferences, an OMT solver will look for a realization that 
 minimizes its global weight.
For instance, one minimum-weight realization of the example CGM, as shown in Figure~\ref{figCGMRelization},
achieves all the nice-to-have requirements except \REQ{Minimal\-Effort},
with a total weight of $-65$, {which is the minimum which can be
achieved with this CGM.}
Such realization requires accomplishing the tasks
\REQ{Characterise\-Meeting},
\REQ{Email\-Participants},
\REQ{Use\-Partner\-Institution},
\REQ{Schedule\-Manually}, 
\REQ{Confirm\-Occurrence},
\REQ{Good\-Participation}, and
\REQ{Minimal\-Conflicts},
and requires no domain assumption.
(This was found automatically by our CGM-Tool 
in $0.008$ seconds on an Apple MacBook Air laptop.)

\fakesubsubsection{Preferences via Multiple Objectives.} 
Stakeholders may define rational-valued \new{objectives}
$obj_1,...,obj_k$ to optimize (i.e., maximize or minimize) as
functions of Boolean and numerical variables
---e.g., 
$\REQ{cost}$, \REQ{workTime}, 
$\REQ{totalCost}$ can be suitable objectives--- and ask the tool to automatically 
generate realization(s) which optimize one objective, or some
combination of more objectives (like \REQ{totalCost}), or which
optimizes lexicographically an ordered list of objectives
\tuple{obj_1,obj_2,...}.
{(We recall that
 a solution optimizes
    lexicographically an ordered list of objectives
    \tuple{obj_1,obj_2,...} if it makes $obj_1$ optimum
  and, if more than one such solution exists, it makes also $obj_2$
  optimum, ..., etc.) 
Notice that lexicographic optimization allows for defining objective
functions in a very fine-grained way and for preventing ties:
if the stakeholder wants to prevent tie solutions on objective
$obj_1$, he/she can define one further preference criterion $obj_2$ in
case of tie on $obj_1$, and so on.} 

Importantly, our CGM-Tool provides some pre-defined objectives of
frequent usage.  
\REQ{Weight} (see last paragraph) is one of them. 
Other examples of pre-defined objectives stakeholders may want to
 minimize, either singularly or in combination
with other objectives, are:
\begin{itemize}
\item[\numdeniedrequirements{}:] the number of nice-to-have
requirements which are not included in the realization;
\item[\numsatisfiedtasks{}:]  the number of 
tasks which are included in the realization;
\item[\numunsatprefs:]  the number of user-defined binary preference
  relations which are not fulfilled by the realization (see later).
\end{itemize}

For example, the previously-mentioned optimum-weight realization of
Figure~\ref{figCGMRelization} 
is such that
~$\REQ{Weight} = -65$,
~$\REQ{workTime} = 4$ and 
~$\REQ{cost} = 80$.
Our CGM has many different
minimum-weight realizations s.t. 
~$\REQ{Weight} = -65$, 
with different values of \REQ{cost} and \REQ{workTime}. 
Among them, 
 it is possible to search, e.g., for the realizations with minimum \REQ{workTime},
and among these  for those with minimum \REQ{cost}, by setting
lexicographic minimization with order
\tuple{\REQ{Weight},\REQ{workTime},\REQ{cost}}. 
This results into one realization with $\REQ{Weight} = -65$,
$\REQ{workTime} =  2$ and $\REQ{cost}=0$ achieving all the nice-to-have
requirements, as shown in Figure~\ref{figCGMLex}, which  requires
 accomplishing the tasks: 
\REQ{Characterise\-Meeting},
\REQ{Collect\-From\-System\-Calendar},
\REQ{Get\-Room\-Suggestions},
\REQ{Cancel\-Less\-Important\-Meeting},
\REQ{Schedule\-Automatically},
\REQ{Confirm\-Occurrence},
\REQ{Good\-Participation},
\REQ{Minimal\-Conflicts},
\REQ{Collection\-Effort}, 
\REQ{Matching\-Effort},
and which requires the domain assumptions:
\REQ{Participants\-Use\-System\-Calendar},
\REQ{Local\-Room\-Available}.
(This was found automatically by our CGM-Tool 
in $0.016$ seconds on an Apple MacBook Air laptop.)
%

\fakesubsubsection{Preferences via Binary Preference Relations.}
In general, stakeholders might not always be at ease in assigning
 numerical values to state their preferences, or in dealing with
 \smtlarat terms, constraints and objectives. 
Thus, as a more {coarse-grained and} user-friendly solution, it 
is also possible for stakeholders to express their preferences 
in a more direct way by stating explicitly a list of
{\em binary preference relations}, denoted as ``\preferred{P_1}{P_2}'',
between pairs of elements of the same kind (e.g. pair of requirements,
of tasks, of domain assumptions) or pairs of
refinements. ``\preferred{P_1}{P_2}'' means that one prefers to have
$P_1$ satisfied than $P_2$ satisfied, that is, that he/she would rather avoid
having $P_1$ denied and $P_2$ satisfied. In the latter case, we say that a
preference is unsatisfied. 
Notice that \preferred{P_1}{P_2} allows for having both $P_1$ and
$P_2$ satisfied or both denied.

\begin{remark}
\label{remark:binaryprefs}
  {These are {\em binary} preferences, so that they say
  nothing on the fact that each $P_i$ is singularly desirable or not,
  which in case must be stated separately (e.g., by penalties/rewards.)
  }
  Thus, the fact that a binary preference \preferred{P_1}{P_2} allows for
  having both $P_1$ and $P_2$ denied should not be a surprise:
  if both $\set{P_1=false,P_2=true}$ and
  $\set{P_1=false,P_2=false}$ violated \preferred{P_1}{P_2}, then
  $P_2$ would play no role in the preference, so that it would reduce
  to the {\em unary} preference ``I'd rather have $P_1$ than not
  have it.'' {A dual argument holds for the fact that
    \preferred{P_1}{P_2} allows for having both $P_1$ and $P_2$
    satisfied.

    Also, this choice is a very general one, since it implements the case
    in which $\tuple{P_1,P_2}$ are both desirable/rewarding (``I
    prefer winning the Turing Award than winning at the lottery.'')
    like the preference between two requirements, as well as the
    opposite case
    in which they are both undesirable/expensive (``I prefer being shot
    than being hanged.'') like the preference between two tasks, plus
    obviously the trivial case in which $P_1$ is desirable and $P_2$
    is undesirable. If this choice is considered too general, then
    the stakeholder can add mutual-exclusion constraints, or combine
    it lexicographically with penalty/rewards, 
    or directly use penalty/rewards instead.} 
\end{remark}

With CGM-Tool, binary preference relations can be expressed either
graphically, via a ``prefer'' arc ``\prefers{P_1}{P_2}'', or via and
ad-hoc menu window.  Once a list of binary preference relations is
set, the system can be asked to consider the number of unsatisfied
preference relations as a pre-defined
 objective (namely \numunsatprefs{}), and it
searches for a realization which minimizes it.
It is also possible to combine such objective lexicographically with
the other objectives.

  One typical usage we {envision} for binary preferences is between
  pairs of refinements of the same element --or equivalently, in case of single-source refinements, between their relative
  source elements. This allows for expressing stakeholders' preferences
  between possible ways one intermediate element can be
  refined.

For example, suppose we want to minimize the total weight of our
example goal model. As  previously mentioned, there is more than one
realization with minimum weight $-65$. 
Unlike the previous example, as a secondary choice we
disregard \REQ{workTime} and \REQ{cost}; rather, we express also the
following  binary preferences:
\begin{eqnarray}
\label{eq:preferences}
&&\preferred{\REQ{By\-System}}{\REQ{By\-Person}},\\
\nonumber
&&\preferred{\REQ{Use\-Local\-Room}}{\REQ{Use\-Partner\-Institutions}},\\
\nonumber
&&\preferred{\REQ{Use\-Local\-Room}}{\REQ{Use\-Hotels\-And\-Convention\-Centers}}.
\end{eqnarray}
\noi 

(Notice that the goal preferences in \eqref{eq:preferences} are
pairwise equivalent to the following refinement preferences:
\begin{eqnarray}
\label{eq:preferencesRef}
\preferred{R_{2}}{R_{10}},\ 
\preferred{R_{5}}{R_{3}},\ \mbox{and}\
\preferred{R_{5}}{R_{4}}
\end{eqnarray}
because the refinements in \eqref{eq:preferencesRef}  are all
single-source ones, whose sources are pairwise the goals in
\eqref{eq:preferences}.)     

Then we set \numunsatprefs{} as secondary objective to
minimize after \REQ{Weight},
that is, we set the lexicographic order \tuple{\REQ{Weight},\numunsatprefs{}}.
Then our tool returned the same realization of
Figure~\ref{figCGMLex} ---that is, the same as with minimizing 
\REQ{workTime} and \REQ{cost} as secondary and tertiary choice---
 instead of that in Figure \ref{figCGMRelization}. (This solution was found in $0.018$ seconds on an Apple MacBook Air laptop.)

\section{Abstract Syntax and Semantics}
\label{sec:goalmodels}
%

\label{sec:goalmodels_structure}
In this section we describe formally the abstract syntax and
  semantics of CGMs. 

\subsection{Abstract Syntax}

{We introduce first some general definitions.}
\label{def:gg}
\noindent
We call a \new{goal graph} \cald a directed acyclic graph (DAG)
  alternating {element nodes}   and {refinement
    nodes} (collapsed into bullets), s.t.:
$(a)$ each element has from zero to many outgoing edges to
  distinct refinements
  and from zero to many incoming edges from distinct refinements;
$(b)$ each refinement node has exactly one outgoing edge to an
  element (\new{target}) and one or more incoming edges from distinct
  elements (\new{sources}). 

We call a \new{root element node } any element node that has no
  outgoing refinement edges, a \new{leaf element node} any (non-root) element
  node that has no incoming refinement edges, and an \new{internal
    element node} any other element node.~%
(Hereafter we will usually drop 
  the word ``node", simply saying ``refinement" for ``refinement node",
  ``element" for ``element node", etc.)

Notice that, by construction, only elements can be
  roots and leaves of a goal graph. 
The sets of root, leaf and internal elements of a goal graph
\cald are denoted as \new{\roots{\cald}}, \new{\leaves{\cald}},
\new{\internals{\cald}} respectively. 
Given a refinement  $R$
 with outgoing edge to the element
 $E$ and incoming edges from the element s $\enum{E}{n}$,
we call $\enum{E}{n}$ the \new{source elements} of $R$ and $E$ the
\new{target element} of $R$, which are denoted by 
\new{\sources{R}} and \new{\target{R}} respectively. We say that 
$R$ is \new{a refinement of $E$} and that 
\new{$R$ refines $E$ into $\enum{E}{n}$}, denoted ``\refines{(\enum{E}{n})}{E}''.
The set of refinements of an element $E$
are denoted with \new{\refinements{E}}.

Elements are \new{goals} or \new{domain assumptions}, subject
to the following rules:
\begin{itemize}
\item a domain assumption cannot be a root element;
\item if the target of a refinement $R$ is a domain assumption, then it sources are only  domain assumptions;
\item if the target of a refinement $R$ is a goal,
then at least one of its sources is a goal.
\end{itemize}
We call root goals and leaf goals \new{requirements} and
\new{tasks} respectively.
%

\begin{table*}
\centering
\caption{\label{tabSumGM}Summary of Goal Model Structure}
\resizebox{\textwidth}{!}{
\begin{tabular}{ccccc}
\hline\noalign{\smallskip}
Constructor & Textual Representation & Graphical Representation & Propositional
Encoding \\ 
\noalign{\smallskip}\hline\noalign{\smallskip}
Goal refinement & $\refines{\bigl(\enum{\Element{E}}{n}\bigr)}{\Element{E}}$ & 
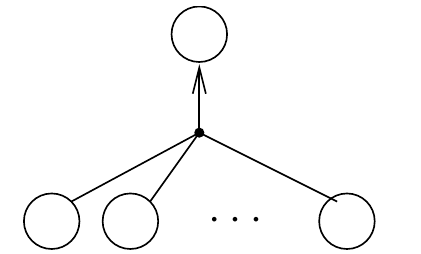
& $((\bigwedge_{j=1}^{n} E_j) \equivalent R) \wedge$ \\
& & & $(R \Implies E)$\\
\noalign{\smallskip}\hline\noalign{\smallskip}
Closed world & --- & 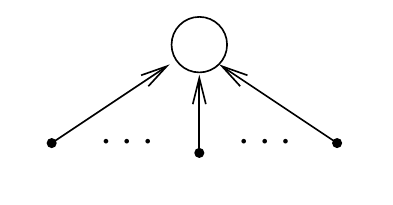 & $  E \Implies \bigl(\bigvee_{R_i \in \rm{Ref}(G)} R_i\bigl)$\\
\noalign{\smallskip}\hline\noalign{\smallskip}
Contribution & $\contributes{E_1}{E_2}$ & 
\begin{tikzpicture}[->,>=stealth',shorten >=1pt,auto,node distance=2.5cm,
  thick,main node/.style={circle,fill=white,draw,font=\sffamily\bfseries}]

  \node[main node] (1) {$E_2$};
  \node[main node] (2) [left of=1] {$E_1$};

  \path[every node/.style={font=\sffamily\small}]
    (2) edge node {$++$} (1) 
;
\end{tikzpicture} 
& $ (E_1 \Implies E_2)$\\
\noalign{\smallskip}\hline\noalign{\smallskip}
Conflict & $ \conflict{\Element{E_1}}{\Element{E_2}}$ &
\begin{tikzpicture}[<->,>=stealth',shorten >=1pt,auto,node distance=2.5cm,
  thick,main node/.style={circle,fill=white,draw,font=\sffamily\bfseries}]

  \node[main node] (1) {$E_2$};
  \node[main node] (2) [left of=1] {$E_1$};

  \path[every node/.style={font=\sffamily\small}]
    (2) edge node {$--$} (1) 
;
\end{tikzpicture} 
& $ \neg \bigl(E_1 \wedge E_2\bigr)$\\
\noalign{\smallskip}\hline\noalign{\smallskip}
Preferences & $\preferred{E_1}{E_2}$ & 
\begin{tikzpicture}[->,>=stealth',shorten >=1pt,auto,node distance=2.5cm,
  thick,main node/.style={circle,fill=white,draw,font=\sffamily\bfseries}]

  \node[main node] (1) {$E_2$};
  \node[main node] (2) [left of=1] {$E_1$};

  \path[every node/.style={font=\sffamily\small}]
    (2) edge node {$prefer$} (1) 
;
\end{tikzpicture} 
& $ (E_1 \vee (\neg E_2))$\\
\noalign{\smallskip}\hline
\end{tabular}
}
\end{table*}

Notationally, 
we use the symbols 
$R$, $R_j$ for  labeling refinements, 
$E$, $E_i$ for generic elements (without specifying 
if goals or domain assumptions), 
$G$, $G_i$ for goals, 
$A$, $A_i$ for domain assumptions.
Graphically (see Figure~\ref{figGMEx})
 we collapse refinements nodes into one bullet, so
that we see a refinement as an aggregation of edges from a set of
other goals.
%
%
(See Table~\ref{tabSumGM}.)
Hence, in a goal graph we consider element nodes as the only nodes,
and refinements as (aggregations of) edges from a group of source
elements to a target element.
%

\begin{definition}[Constrained Goal Model] 
\label{def:cgm}
A \new{Constrained Goal Model (CGM)} is a tuple $\calm\defas\tuple{\calb,\caln,\cald,\Psi}$, s.t.
\begin{itemize}
\item $\calb\defas\calg\cup\calr\cup\cala$ is a set of atomic
  propositions, where
$\calg\defas\{G_1,...,G_N\}$,
$\calr\defas\{R_1,...,R_K\}$,
$\cala\defas\{A_1,...,A_M\}$ are respectively sets of 
{goal},
{refinement} and
{domain-assumption labels}.
We denote with \cale the set of element labels:
$\cale\defas\calg\cup\cala$;
\item \caln is a set of numerical variables in the rationals;

\item \cald is a goal graph, s.t. all its goal nodes are univocally labeled
  by a goal label in \calg, all its refinements are univocally
  labelled by a refinement label in \calr,
 and all its domain assumption are univocally labeled
 by a assumption label in \cala;
\item $\Psi$ is a \smtlarat formula on \calb and \caln.

\end{itemize}
\end{definition}

A CGM is thus a ``backbone'' goal graph
  \cald{} --i.e., an and-or directed acyclic graph (DAG) of
\emph{elements}, as 
nodes, and \emph{refinements}, as (grouped) edges, which are labeled
by atomic propositions in \calb-- which is augmented with 
an  \smtlarat{} formula $\Psi$ on the element and refinement labels in
\calb and on the numerical variables in \caln.  
The \smtlarat formula $\Psi$ is a conjunction of smaller formulas
 encoding relation edges, global
and local Boolean/\smtlarat
constraints, user assertions, and the definition of numerical
objectives, all of which will be described later in
this section.

Intuitively, a CGM describes a (possibly complex) combination of
alternative ways of realizing a set of requirements in terms 
of a set of tasks, under certain domain assumptions and constraints.
A couple of remarks are in order.
\begin{remark}
\label{remark:andor}
The fact that the goal graph \cald is an \new{and-or} graph can be deduced
from the propositional encoding of Goal refinement and Closed {World} in
Table~\ref{tabSumGM}:
by combining the propositional encodings of goal refinement
and Closed {World} in Table~\ref{tabSumGM}, we can infer the formulas:~%
\footnote{\label{footnote:andor}{We recall that in Boolean logic the formula
    $\bigwedge_i (R_i\imp E)$, which comes from the goal refinement
    encoding in Table~\ref{tabSumGM}, is equivalent to $E \limp
    (\bigvee_i R_i)$. The latter, combined with the encoding of Closed
    World $E \imp (\bigvee_i R_i)$, gives the left formula in
    \eqref{eq:andor}. The right formula in \eqref{eq:andor} is the
    other part of the goal refinement encoding in
    Table~\ref{tabSumGM}.}}

\begin{eqnarray}
\label{eq:andor}
E \iff (\bigvee_i R_i)\ \  &\mbox{and}&\ \  R \iff (\bigwedge_j E_j).
\end{eqnarray}
Thus, each non-leaf element $E$ is or-decomposed into 
the set of its incoming refinements $\set{R_i}_i$, and each refinement $R$
is and-decomposed into the set of its source elements $\set{E_j}_j$.~
\end{remark}  

%
%
\begin{figure}[t]
  \centering
\resizebox{\textwidth}{!}{
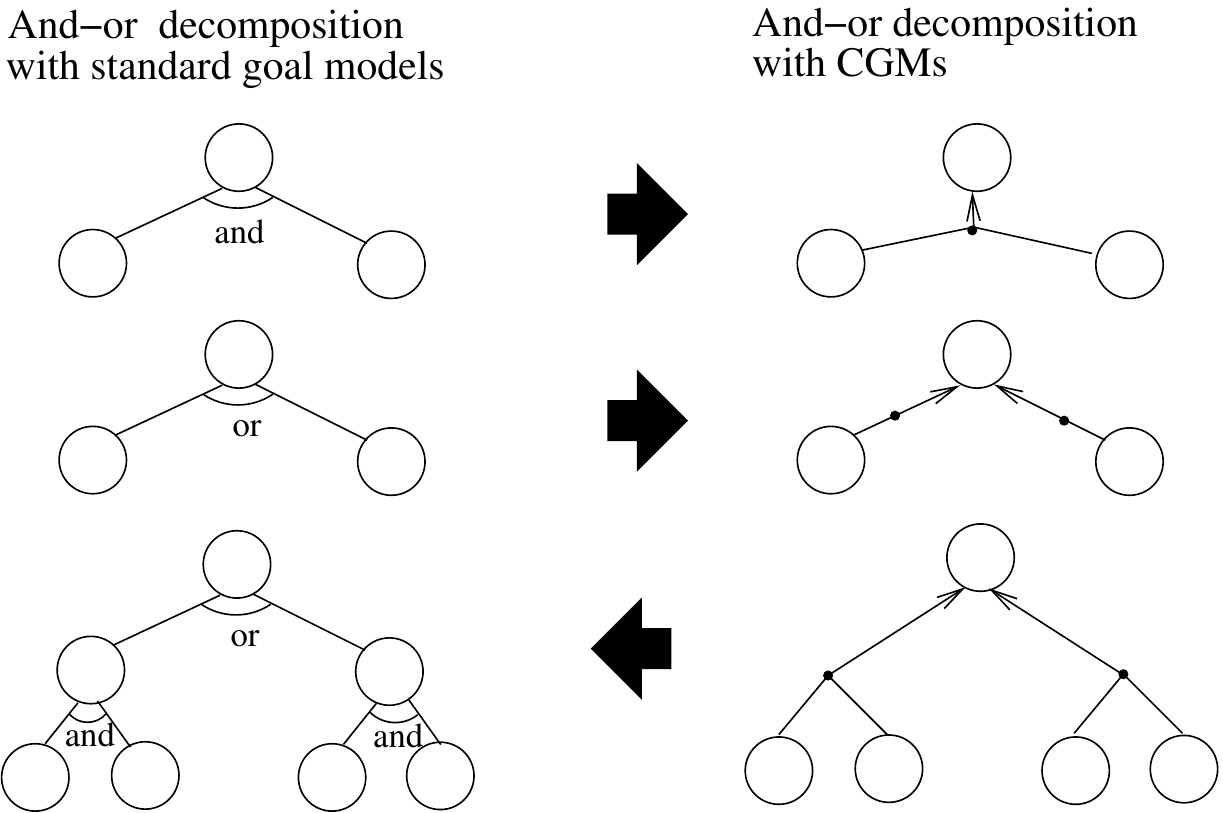  
}  \caption{{
Top: and-decomposition and its translation into CGM format as a single
multi-source refinement.
Middle: or-decomposition and its translation into CGM format as
multiple single-source refinements.
\mbox{Bottom: a simple piece of CGM (right) and its translation into standard}
      and-or goal model format (left): it is necessary to
      introduce two auxiliary goals
    $G'$ and $G''$ to encode the refinements $R_1$ and $R_2$.}
  \label{fig:cgmvsgm}}
\end{figure}

\begin{remark}
{
CGMs are more succinct in terms of number of goals 
than standard and-or goal models.
On the one hand, a standard $n$-ary and-decomposition 
of a
goal can be represented straightforwardly in a CGM by 
one refinement with $n$ sources {(Figure~\ref{fig:cgmvsgm},
  Top)},
and an or-decomposition by $n$ one-source refinements (Figure~\ref{fig:cgmvsgm},
  Middle), so that no extra goals are added.
On the other hand,
in order to represent a piece of CGM  with $n$ non-unary refinements by
standard goal models, we need  introducing $n$ 
new auxiliary intermediate goals to encode refinements, which CGMs
encode natively (Figure~\ref{fig:cgmvsgm},
  Bottom).
 We recall from
\sref{sec:goalmodels_example} that refinements do not 
need to be explicitly labeled unless 
{they need to be mentioned in other parts of the model}.}
\end{remark}  

Stakeholders might not be at ease in defining
a possibly-complex
{\em global} \smtlarat{} formula $\Psi$ to encode constraints among elements
and refinements, plus numerical variables. 
 To this extent, as mentioned in \sref{sec:goalmodels_example},
apart from the possibility of defining global formulas,
 CGMs provide  constructs allowing the
user to encode {\em graphically} and {\em locally} desired constraints
of frequent usage: \new{relation edges}, \new{prerequisite formulas}
$\Constraint{G}{\phi}$ and $\Constraint{R}{\phi}$ and \new{user assertions}.
Each is automatically converted into a simple \smtlarat formula 
as follows, and then conjoined to $\Psi$.

\begin{itemize}
\item[{\em Element-contribution edges},] 
  $\contributes{E_1}{E_2}$,  meaning that 
satisfying $E_1$ forces $E_2$ to be satisfied (but not vice versa). 
They 
are encoded into  the formula $(E_1\imp E_2)$.
(The edge
$\bicontributes{E_1}{E_2}$ can be used to denote the merging of the
two contribution edges
$\contributes{E_1}{E_2}$ and
$\contributes{E_2}{E_1}$ into one.)

\item[{\em Element-conflict edges},] 
  $\conflict{E_1}{E_2}$, meaning that 
$E_1$ and $E_2$ cannot be both
  satisfied. 
They 
are encoded into the formula
$\neg (E_1\wedge E_2)$.

\item[{\em Refinement-binding edges},] 
  $\bind{\Refinement{R_1}}{\Refinement{R_2}}$, 
{meaning that, if both 
the target goals of $R_1$ and $R_2$ (namely $E_1$ and  $E_2$
respectively) are satisfied, 
then $R_1$ refines $E_1$  if and only if 
$R_2$ refines $E_2$. They
}
are encoded into the formula 
$(E_1 \wedge E_2) 
\imp 
(\Refinement{R_1}\iff\Refinement{R_2})$.

\item[{\em User assertions},] $E_i:=\top$ and $E_j:=\bot$, are
  encoded into the formulas $(E_i)$, $(\neg E_j)$ respectively.

\item[{\em Prerequisite formulas},] $\Constraint{G}{\phi}$ 
[resp. $\Constraint{R}{\phi}$]
are
  encoded into the formulas 
$(G\imp \posConstraint{G}{\phi})$ and
$(\neg G\imp \negConstraint{G}{\phi})$ 
[resp. 
$(R\imp \posConstraint{R}{\phi})$ and
$(\neg R\imp \negConstraint{R}{\phi})$].

\end{itemize}

The following are instead encoded into \smtlarat ``soft''~%
\footnote{{In constraint programming and other related disciplines
  (e.g. MaxSAT, MaxSMT, OMT) constraints which must be satisfied are
  called ``hard'', whereas constraints which are preferably satisfied
  but which can be safely violated, although paying some penalty, are
  called ``soft''.}}
 constraints:
\begin{itemize}
\item[{\em Preference edges},] 
$\prefers{E_1}{E_2}$ [resp. $\prefers{R_1}{R_2}$], and their
equivalent {\em binary preference relations}
$\preferred{E_1}{E_2}$ [resp. $\preferred{R_1}{R_2}$], are implemented into the soft constraint
$\phi_{\preferred{E_1}{E_2}}\defas(E_1 \vee (\neg E_2))$ 
[resp. $\phi_{\preferred{R_1}{R_2}}\defas(R_1 \vee (\neg R_2))$].
{(See also Remark~\ref{remark:binaryprefs} in \sref{sec:preferences}.)}
Notice that $E_1$ and $E_2$ [resp. $R_1$ and $R_2$] must be of the same kind, i.e. they must 
be  both tasks, or both requirements, or both refinements, 
or both intermediate goals, or both domain
assumptions. 
\end{itemize}

Unlike with other constraints, these soft constraints 
are {\em not} added directly to $\Psi$. Rather, 
the following \smtlarat constraint, which defines 
a numeric Pseudo-Boolean {\em cost function}, is added to $\Psi$:
\begin{eqnarray}
\label{eq:costpref}
 (\numunsatprefs{} &=&
\sum_{\tuple{{E_i}{E_j}}\in \calp} \hspace{-.3cm}
{\sf ite}(\phi_{\preferred{E_i}{E_j}},0,1) +
\sum_{\tuple{{R_i}{R_j}}\in \calp} \hspace{-.3cm}{\sf ite}(\phi_{\preferred{R_i}{R_j}},0,1)), 
\end{eqnarray}
\noindent
where $\calp$ is the list of binary preference relations, and
``${\sf ite}(\phi_{*},0,1)$" denotes an
if-then-else arithmetical term, which is evaluated to 0 if $\phi_{*}$
is evaluated 
to true, to 1 otherwise. Hence, \numunsatprefs{} counts the number of
unsatisfied preferences, {that is, the number of binary
  preferences \preferred{P_i}{P_j} s.t. $P_i$ is false and $P_j$ is true.}
\footnote{In practice, the OMT solver OptiMathSAT \cite{st_cav15} provides more
efficient ad-hoc encodings for soft constraints like those in
\eqref{eq:costpref}, which we have 
exploited in the implementation of CGM-Tool; we refer the reader to
\cite{st_cav15} for details.}

Notice that, unlike refinements, relation edges and preference edges 
are allowed to create
loops, possibly involving refinements. 
{In fact, refinements are acyclic because they represent the
  and-or decomposition DAG or the CGM requirements. Other arcs (and
  formulas) represent relations and constraints among elements, and as
  such they are free to form loops, even with refinements.}

\smallskip
Finally we provide the user {of a list of
  syntactic-sugaring constructs, which allow for} defining, both
globally and locally, the most
standard and intuitive 
constraints among assumption, goal and refinement labels,
 with no need of defining
the corresponding complicate or less-intuitive propositional formulas. 
(In what follows, $P_1,...,P_n$ denote atomic propositions in \calb.)

\begin{itemize}
\item[$\predicate{Alt}{P_1,P_2}$] denotes the fact
$P_1$ and $P_2$ are alternative, e.g., that one and only one of them 
is satisfied. 
This is encoded by the formula $(P_1\iff \neg P_2)$. 
\item[$\predicate{Causes}{P_1,P_2}$] denotes the fact
  that satisfying $\Goal{P_1}$  causes $\Goal{P_2}$ to be satisfied.
This is
encoded by the formula $(P_1\imp P_2)$. 
\item[$\predicate{Requires}{P_1,P_2}$] denotes the fact
  that satisfying $\Goal{P_1}$ requires $\Goal{P_2}$ to be satisfied.
This is encoded by the formula $(P_1\imp P_2)$.~%
\footnote{
Notice that the relation edge $\contributes{P_1}{P_2}$, and the
Boolean constraints $\predicate{Causes}{P_1,P_2}$,
$\predicate{Requires}{P_1,P_2}$, and $(P_1\imp P_2)$ are equivalent 
from the perspective of Boolean semantics. Nevertheless, stakeholders may 
use them in different contexts: e.g., ``$\predicate{Causes}{P_1,P_2}$''
is used when event $P_1$ occurs before $P_2$ and the former causes the latter, 
whereas ``$\predicate{Requires}{P_1,P_2}$'' 
is used when $P_1$ occurs after $P_2$ and the former requires the
latter as a prerequisite.

}

\item[$\predicate{AtMostOneOf}{\{P_1,...,P_n\}}$] denotes the fact that at most
  one of $\{P_1,...,P_n\}$ must be satisfied. This is  encoded 
by the formula  $\bigl(\bigwedge_{1 \leq i < j \leq n} (\neg P_i \vee
\neg P_j)\bigr)$.  

\item[$\predicate{AtLeastOneOf}{\{P_1,...,P_n\}}$] denotes the fact
  that at least one of $\{P_1,...,P_n\}$ must be satisfied. This is
 encoded by the formula $\bigl(\bigvee_{1 \leq i \leq n}
  P_i\bigr)$.
\item[$\predicate{OneOf}{\{P_1,...,P_n\}}$] denotes the fact
  that exactly one of $\{P_1,...,P_n\}$ must be satisfied. This is
   encoded by the conjunction of the previous two formulas.

\end{itemize}

\subsection{{Semantics}}

The semantics of CGMs is formally defined{ in terms of the
semantics of simple Boolean expressions}, as follows.

\begin{definition}[Realization of a CGM]
\label{def:realization}
{Let $\calm\defas\tuple{\calb,\caln,\cald,\Psi}$ be a CGM.
A \new{realization} of \calm 
is a \larat-interpretation $\mu$ over $\calb\cup\caln$
such that:
}
\begin{aenumerate}
\item 
$\mu \models ((\bigwedge_{i=1}^{n} E_i) \equivalent R) \wedge  (R
\Implies E)$ for each refinement  $\refines{\bigl(\enum{E}{n}\bigr)}{E}$;
\item 
$\mu\models \bigl( E \Implies (\bigvee_{R_i \in \rm{Ref}(E)}
R_i)\bigr)  $, for each non-leaf element $E$;
\item $\mu\models\Psi$.
\end{aenumerate}
\noindent
We say that \calm
is \new{realizable} if it has at least one realization, 
 \new{unrealizable}
otherwise.
\end{definition}

\noindent

Alternatively and equivalently, {({\em a})} and {({\em b})} can be substituted by the 
conditions:
\begin{aenumerateprime}
\item 
$\mu \models ((\bigwedge_{i=1}^{n} E_i) \equivalent R)$ for each refinement  $\refines{\bigl(\enum{E}{n}\bigr)}{E}$;
\item 
$\mu\models \bigl( E \iff (\bigvee_{R_i \in \rm{Ref}(E)}
R_i)\bigr)  $, for each non-leaf element $E$,
\end{aenumerateprime} 
\noindent
which reveal the and-or structure of $\cald$.  
~%
({Recall Remark~\ref{remark:andor} and
    Footnote~\ref{footnote:andor}.})

\medskip
In a realization $\mu$ for a CGM
$\calm\defas\tuple{\calb,\caln,\cald,\Psi}$,
 each element $E$ or refinement $R$ can be
either \emph{satisfied} or \emph{denied} (i.e., their label can be
assigned true or false respectively by $\mu$), and each numerical
value is assigned a rational value. 
$\mu$ 
is represented graphically as the  sub-graph of \cald which includes 
all the satisfied elements and refinements and does not include
the denied elements and refinements. 
As an example, consider the realization {highlighted in yellow}
in Figure~\ref{figGMEx}, where
$\REQ{cost}=0$ and $\REQ{cost_{E}}=0$ for every element $E$. 
From Definition~\ref{def:realization}, 
a realization $\mu$ 
represents  a sub-graph of the CGM, such that:
\begin{aenumerate}
\item 
A refinement $R$ is part of $\mu$ if and only if all
its  source elements $E_i$ are also included.
Moreover, if $R$ is part of $\mu$, then also its 
target element $E$ is part of it. 
(See, e.g., refinement $R_1$ for \REQ{ScheduleMeeting}, with all its
source goals.)

\item If a non-leaf goal is in a realization sub-graph, then at least
  one of its refinements is included in the realization.
(See, e.g., refinement $R_5$ for \REQ{FindASuitableRoom}.)

\item A realization complies with all 
 Boolean and \smtlarat constraints of the CGM, including relational
 edges, global and local formulas, user assertions, and the definitions
of the numerical attributes and objectives. In particular:

\begin{itemize}
\item[$\contributes{E_1}{E_2}$:] 
If $E_1$ is in $\mu$, then $E_2$ is in $\mu$.
(See, e.g., the contribution edge \contributes{\REQ{BySystem}}{\REQ{CollectionEffort}}.)

\item[$\conflict{E_1}{E_2}$:] 
 $E_1$ and $E_1$ cannot be both part of $\mu$.
(See, e.g., the conflict edge \conflict{\REQ{Byperson}}{\REQ{CollectionEffort}}.)

\item[ $\bind{\Refinement{R_1}}{\Refinement{R_2}}$:] 
if both the target goals of $R_1$ and $R_2$ 
are part of the realization $\mu$, then  
$R_1$ is in $\mu$  if and only if $R_2$ is there.
(See, e.g., the binding \bind{R_{16}}{R_{17}}.)
\item[{\em User assertions}:] If $E_i$ is marked satisfied
  [resp. denied], then $E_i$ is [resp. is not] part of a realization $\mu$.
(See, e.g., the requirement \REQ{ScheduleMeeting}, which is mandatory,
i.e., it is marked satisfied.)

\item[$\posConstraint{G}{\phi}$:] if $G$ is part of a realization $\mu$, 
then $\posConstraint{G}{\phi}$ must be satisfied in $\mu$.
(E.g., \REQ{LowCost} is part of $\mu$, so that
$\posConstraint{G}{\phi}\defas ... \wedge(\REQ{cost}<100)$ is
satisfied, in compliance with the fact that $\mu$ sets $\REQ{cost}=0$.)

\item[$\negConstraint{G}{\phi}$:] if $G$ is {\em not} part of a
  realization $\mu$,  
then $\negConstraint{G}{\phi}$ must be satisfied in $\mu$.
(E.g., \REQ{UsePartnerInstitutions} is not part of $\mu$,
so that 
$\negConstraint{\REQ{UsePartnerInstitutions}}{\phi}$ --which includes
$(\REQ{cost_{UsePartnerInstitutions}}=0)$ by
\eqref{eq:globalnegdefault}-- is satisfied, in compliance with 
the fact that $\mu$ sets $\REQ{cost_E}=0$ for every $E$.)

\item[{\em Global formulas and attribute definitions}:] The realization complies with all
  global formulas and attribute definitions. (E.g., the global formula $(\REQ{cost}=\sum_E
  \REQ{cost_E})$, which defines the attribute \REQ{cost}, is satisfied
  by $\mu$ because $\REQ{cost}=0$ and $\REQ{cost_{E}}=0$ for every
  element  $E$. ) 

\end{itemize}

\end{aenumerate}


\noindent

\ignore{
In a realization,
 each element $E$ or refinement $R$ can be
either \emph{satisfied} or \emph{denied} (i.e., their label can be
assigned to $\top$ or $\bot$ respectively by $\mu$). 
If an element $E$ is not a leaf, then it can be satisfied only by satisfying
the set of source elements $\enum{E}{n}$ 
of one of its refinements
$\refines{\bigl(\enum{E}{n}\bigr)}{E}$.
If $\mu$ satisfies a refinement $\Refinement{R}$ of an element $E$, i.e.,
it satisfies all the source elements $\enum{E}{n}$, then it 
satisfies the element $E$, but not vice versa (condition $(a)$). 
For a non-leaf element to be satisfied, at least one
of its refinements must be satisfied (condition $(b)$).
We call this fact  \new{Closed
World Assumption (CWA)}.
 The satisfiability or deniability of each element or refinement can be
 further constrained by all the constraints defined inside the formula
 $\Psi$:  every realization $\mu$ 
 must satisfy such constraints (condition $(c)$). 
Notice that, by fulfilling condition $(c)$,  
 a realization must implicitly comply also with all the
relation edges, with the user assertions and
 with the
local pre-requisite constraints $\Constraint{E}{\phi}$ and
$\Constraint{R}{\phi}$, 
because the corresponding formulas are conjuncts of $\Psi$. 
Thus $\Psi$ contains also the  global and local
\smtlarat constraints over global and local numerical attributes
(e.g. 
$\REQ{Low Cost}\imp (\REQ{cost} \leq 100)$,
$\REQ{Use Partner Institutions}\imp (\REQ{cost_{Use Partner
    Institutions}} = 80)$, and the definitions of 
objectives
 (e.g., $(\REQ{cost} = \sum_{E \in \cale}\REQ{cost_{E}}$).
}

   \begin{remark}
\label{remark:onlynonzeros}
   Importantly, in the definition of objectives only
   non-zero terms of the sums need to be considered. (E.g., the sum
in $(\REQ{cost} = \sum_{E   \in \cale}\REQ{cost_{E}})$ can be safely
restricted to the elements 
\REQ{Use Partner Institutions} and  \REQ{Use Hotels And Convention
  Centers}.)
This allows for reducing drastically the number of rational variables 
involved in the encoding. {In the implementation of CGM-Tool we have
exploited this fact.}
   \end{remark}

\ignore{
  \begin{futureversion}
\subsection{Evolving Constrained Goal Models}
\label{sec:evolving_goalmodels_structure}
Constrained goal models may evolve in time: goals may be added,
refinements can be added, removed or simply modified, etc. 
If well-established realizations of previous versions 
are available, one may want to find realizations of the modified 
CGM  with make it minimal the penalty of {\em novel} subgoals to be
achieved. 

Let $\calm_1\defas \tuple{\calb_1,\caln_1,\cald_1,\Psi_1}$ be the original model, 
and $\mu_1$ be the realization of $\calm_1$. Let it be the case that the system 
of the original model $\calm_1$ is already built based on the realization $\mu_1$.
Over the time, the model $\calm_1$ evolves and becomes a new
model $\calm_2\defas \tuple{\calb_2,\caln_2,\cald_2,\Psi_1}$. In the elvolution
requirement engineering problem, we want to find a realization $\mu_2$ for $\calm_2$
such that we can make the most use out of $\mu_1$. The definition for ``most use"
can be varied, it can be the familiarity between the two realization $\mu_1$ and $\mu_2$,
or the effort needed to implement the system from $\mu_2$ provided that we already have 
$\mu_1$ implemented. Let $\cale_1$ and $\cale_2$ be the subset of \new{elements of interest}\footnote{The elements of interest can be either requirements, tasks, domain assumptions, and/or refinements. As the default it is set to be the set of elements of the CGM, users are free to modified it.} in $\calm_1$ and $\calm_2$ respectedly, we set the default choices of optimizing the realization of the evolved 
CGM as:

\begin{enumerate}
\item
``Familiarity": 

\begin{itemize}
\item
Minimize the hamming distance between $\mu_1$ and $\mu_2$ over the common elements $E \in \cale_1 \cap \cale_2$, or
\item 
Minimize the total number of satisfied new elements in $\mu_2$, i.e., the elements $E \in \cale_2 \setminus \cale_1$, or
\item
Minimize both the hamming distance between $\mu_1$ and $\mu_2$ over the common elements $E \in \cale_1 \cap \cale_2$ and the total number of satisfied new elements in $\mu_2$, i.e., the elements $E \in \cale_2 \setminus \cale_1$.
\end{itemize}

\item
``Effort": Minimize the total number of newly satisfied tasks $T$, i.e., all the tasks $T \in \cale_2$ such that $\mu_2(T) = \top$ and either $(T \notin \cale_1)$ or $(T \in \cale_1$ and  $\mu_1(T) = \bot)$.
\end{enumerate}

Which can be encoded as bellow,

Consider the following subsets of elements of $\cale_2$:
\begin{eqnarray}
\label{eq:rangeone}
  \cale_{new} & \defas & \{E_i \in \cale_2 \setminus \cale_1\},\\
\label{eq:rangetwo}
  \cale_{common} & \defas & \{E_i \in \cale_2 \cap \cale_1\},\\
\label{eq:rangethree}
  \calt_{new} & \defas & \{T_i \in \cale_{new} \  \mid \  \mu_2(T_i)=\top\},\\
\label{eq:rangefour}
  \calt_{common} & \defas & \{T_i \in \cale_{common} \ \mid \ \mu_2(E_i) \neq \mu_1(E_i)\}.\\
\end{eqnarray}

As the default, we have three different penalties of $\mu_2$ wrt. $\mu_1$ (user can define his own penalty using SMT variables):
  \begin{eqnarray}
\label{eq:addedcost_1} 
 \addedcostof{\mu_2\ |\  \mu_1}& \defas &   \text{ the total number of elements in } \calt_{new}, \\
\label{eq:addedcost_2} 
 \modifiedcostof{\mu_2\ |\  \mu_1}& \defas &  \text{ the total number of elements in } \calt_{common} , \\
\label{eq:addedcost_3} 
 \costof{\mu_2\ |\  \mu_1}& \defas &  \text{ the total number of elements in } \calt_{new} \cup \calt_{common} .
  \end{eqnarray}

Intuitively, to maximize the familiarity, the realization $\mu_2$ will need to have the minimum value of (\ref{eq:addedcost_1}) (i.e., as little new elements in the evolved system as possible), or the minimum value of (\ref{eq:addedcost_2}) (i.e., as little changes in the evolved system as possible), or both, depending on the user/stakeholder definition of familiarity; and to minimize the effort the realization $\mu_2$ will need to have the minimum value of (\ref{eq:addedcost_3}), where the elements of interest are restricted to task elements only (i.e., implemented as little new tasks as possible).

\ignore{
Constrained goal models may evolve in time: goals may be added,
refinements can be added, removed or simply modified, etc. 
If well-established realizations of previous versions 
are available, one may want to find realizations of the modified 
CGE  with make it minimal the penalty of {\em novel} subgoals to be
achieved. 

\PGJMTODO{Add motivations here}

\RSTODO{Migliorare notazione?\\}
\RSTODO{E' necessario $\calm_2$ nella Definizione~\ref{def:addedcost}?}

\begin{definition}[Added penalty of a realization and minimum added penalty
  of a WCGM] 
\label{def:addedcost}
%
Let $\calm_1\defas \tuple{\calb_1,\cald_1,\Psi_1,\calw_1}$ and
 $\calm_2\defas \tuple{\calb_2,\cald_2,\Psi_2,\calw_2}$
two WCGM s.t. $\calw_1(E_i)=\calw_2(E_i)$ for every 
$E_i\in\cale_1\cap\cale_2$.
Let \assertions{}  be a set of user assertions on elements in 
$E_i\in\cale_1\cap\cale_2$, and let
$\mu_1$ and $\mu_2$ be realizations of 
\tuple{\calm_1,\assertions} and \tuple{\calm_2,\assertions}
respectively.   
Consider the following subsets of elements of $\cale_2$:
\begin{eqnarray}
\label{eq:rangeone}
  \elementsone &\defas& \rangeone,\\
\label{eq:rangetwo}
  \elementstwo &\defas& \rangetwo.
\end{eqnarray}

We call the \new{added penalty} of $\mu_2$ wrt. $\mu_1$,
denoted as \addedcostof{\mu_2\ |\  \mu_1,\assertions}, the value
  \begin{eqnarray}
\label{eq:addedcost_1} 
&&    \sum_{E_i\ \in\ \elementsone\cup\elementstwo} \calw_2(E_i).
  \end{eqnarray}

We call the \new{minimum added penalty} of \tuple{\calm_2,\assertions} wrt. 
$\mu_1$, which we denote as 
\minaddedcostof{\calm_2\ |\ \mu_1}{\assertions}, the value 
  \begin{equation}
\label{eq:minaddedcost}
min_{\{\mu_2\ realization\ of\ \tuple{\calm_2,\assertions}\}}\ \{
\addedcostof{\mu_2\ |\ \mu_1,\assertions} \}.
  \end{equation}
\end{definition}

\noindent
Notice that, by construction, the elements of the designer's assertions in \assertions
do not occur in $\elementsone\cup\elementstwo$ because they occur
in both $\cale_1$ and $\cale_2$ and they are assigned the same truth
value by $\mu_1$ and $\mu_2$. 
Notice also that the definition \eqref{eq:minaddedcost} 
does depend on $\mu_1$: in fact,
you can have two realitations $\mu_1$ and $\mu_1'$ with the same penalty 
s.t. 
$$\minaddedcostof{\calm_2\ |\  \mu_1}{\assertions}
\neq
\minaddedcostof{\calm_2\ |\  \mu_1'}{\assertions}.
$$

Intuitively, once the designer's assertions \assertions are set,
\addedcostof{\mu_2\ |\  \mu_1,\assertions} is the sum of the 
penalties  of the leaf elements in $\calm_2$ which are satisfied by $\mu_2$  
and which either
 do not occur in $\calm_1$ \eqref{eq:rangeone}
or occur in $\calm_1$ but are denied in $\mu_1$ \eqref{eq:rangetwo}.
%
This represents the total penalty of the primitive goals which it is
necessary to achieve from scratch, either because they are new
\eqref{eq:rangeone}, or because they were previously denied
\eqref{eq:rangetwo}.  Notice that here we consider neither the common
leaf elements which previously were satisfied and now are denied,
because they have been previously achieved anyway, nor the root goals,
because they are not part of the designer's assunptions ans as such
they are not considered relevant.

Given a previous ``best'' realization $\mu_1$ for $\calm_1$, 
and given a novel version of the model $\calm_2$, we want to find a
novel realization $\mu_2$ which minimizes the penalty of unachieved leaf
elements to be achieved.

\begin{definition}[Encoding of a minimum-added-penalty problem]
\label{def:encoding_minimumaddedcost} 
Let $\calm_1$, $\calm_2$, $\assertions$, $\mu_1$, \elementsone{}, 
\elementstwo{} and \minaddedcostof{\calm_2\ |\ \calm_1}{\assertions} be 
as in Definition~\ref{def:addedcost}.
%
The \new{OMT encoding} of the 
\minaddedcostof{\calm_2\ |\ \calm_1}{\assertions} problem 
 is given by the pair 
\tuple{\encaddcostass,\cost}, where:
\begin{eqnarray}
\label{eq:omtenc_1}
  \encaddcostass &\defas& \enc{}  
\wedge 
(\cost =  
\sum_{E_i\ \in\ \elementsone\cup\elementstwo} t_i ) \\
\label{eq:omtenc_2}
&\wedge& \bigwedge_{E_i\ \in\ \elementsone\cup\elementstwo} \left( 
  \begin{array}{l}
(\pos E_i\imp  (t_i=\calw(E_i))) \wedge \\
(\neg E_i\imp (t_i=0))    \wedge \\
(t_i\ge 0) \wedge (t_i\le \calw(E_i))
  \end{array}
\right)
\end{eqnarray}
where
$\enc{}$ is a propositional formula as in
Definition~\ref{def:encoding},
\cost and the $t_i$'s are fresh variables, one for eachleaf
element in $\elementsone\cup\elementstwo$.

\end{definition}

The following fact is a direct consequence of 
Propositions~\ref{prop:main} and \ref{prop:main_omt} and of Definitions
~\ref{def:addedcost} and \ref{def:encoding_minimumaddedcost}.

\begin{proposition}
\label{prop:main_evolving}
Let $\calm_1$, $\calm_2$, $\assertions$, $\mu_1$, \elementsone{}, 
\elementstwo{}, \minaddedcostof{\calm_2\ |\ \calm_1}{\assertions} and 
\encaddcostass{} be 
as in Definitions~\ref{def:addedcost} and \ref{def:encoding_minimumaddedcost}.
Let $\mu_2$ be a truth assignment to the variables in $\calb_2$. 
Then
$\mu_2$ is a minimum-added-cost realization of \tuple{\calm_2,\assertions}  
wrt. $\mu_1$ 
if and only if
$\mu_2$ is a solution to the OMT problem 
\tuple{\encaddcostass,\cost}.
\end{proposition}

}    
  \end{futureversion}
}

\section{Automated Reasoning with Constrained Goal Models}
\label{sec:functionalities}
{In this section we describe how to perform automated
  reasoning functionalities on CGMs by encoding them into SMT and OMT.}
\ignore{
{We first show how to encode a CGM \calm into a \smtlarat formula
  $\Psi_\calm$, so that the search for an optimum realization of \calm 
reduces to an \omtlarat problem over the formula $\Psi_\calm$, which is then
  fed to an OMT solver. Then we present the reasoning functionalities
  over CGMs we have implemented on top of our OMT solver. 
}
}

\ignoreinshort{\subsection{Encoding of Constrained Goal Models}}
\label{sec:goalmodels_encodings}
\begin{definition}[\smtlarat Encoding of a CGM]
\label{def:encoding}
Let $\calm\defas\tuple{\calb,\caln,\cald,\Psi}$ be a CGM.
The \new{\smtlarat encoding} of \calm is the  formula 
$\enc{}\defas\Psi\wedge\Psi_\calr\wedge\Psi_\cale$, where:
\begin{eqnarray}
\label{eq:encoding1}
 \Psi_\calr&\defas& 
\bigwedge_{\refines{\bigl(\enum{E}{n}\bigr)}{E},\ R
  \in \calr}\  
\bigl(
 (\bigwedge_{i=1}^{n} E_i \equivalent R) \wedge  (R
\Implies E)
\bigr),
\\
\label{eq:encoding2}
\Psi_\cale&\defas& 
\bigwedge_{E\in\roots{\cald}\cup\internals{\cald}}\ 
\bigl( E \Implies (\bigvee_{R_i \in \rm{{\sf Refinements}}(E)} R_i) \bigr).
\end{eqnarray}
\noi \roots{\cald} and \internals{\cald} being the root and internal
elements of \cald respectively.
{We call $\Psi_\calm$ the {\em \smtlarat Encoding} of the
  CGM \calm.}
\end{definition}

Notice that the formulas $\Psi_\calr$ and $\Psi_\cale$ in 
\eqref{eq:encoding1}  and \eqref{eq:encoding2} encode
directly points ({\em a}) and  ({\em b}) in
Definition~\ref{def:realization}, for every element and refinement in
the CGM. In short, the $\Psi_\calr\wedge\Psi_\cale$ component of
$\Psi_\calm$ encodes the relation induced by the and-or goal
graph \cald in \calm. 
The component $\Psi$ is the formula described in point  ({\em c}) in
Definition~\ref{def:realization},
which encodes all 
 Boolean and \smtlarat constraints of the CGM, including relational
 edges, global and local formulas, user assertions, and the definitions
of the numerical attributes and objectives.

Therefore, the following facts are straightforward consequences of
Definitions~\ref{def:realization} and \ref{def:encoding} and of the
definition and \omtlarat.

\begin{proposition}
\label{prop:main} 
Let $\calm\defas\tuple{\calb,\caln,\cald,\Psi}$ be a CGM; let $\enc{}$
  its \smtlarat encoding {as in Definition~\ref{def:encoding}}; 
let $\mu$ a \larat-interpretation over $\calb\cup\caln$. 
Then $\mu$ is a realization of
\calm 
if and only if 
$
  \mu\models
\enc{}.
$
\end{proposition}
  In short, Proposition \ref{prop:main} says that $\mu$ is a
  realization for the CGM \calm if and only if $\mu$ is a model in
  \smtlarat for the formula $\Psi_\calm$. Therefore, a realization
  $\mu$ for \calm can be found by invoking a \smtlarat solver on 
the CGM encoding $\Psi_\calm$.

\begin{proposition}
\label{prop:main2} 
Let $\calm$ and $\enc{}$ be as in Proposition~\ref{prop:main},
and let $\mu$ be a realization of \calm. 
Let $\set{obj_1,...,obj_k}$ be numerical objectives occurring in $\Psi_\calm$.
Then we have that:
\begin{itemize}
\item[(i)] for every i in $1,...,k$, $\mu$ minimizes [resp. maximizes] $obj_i$
if and only if  $\mu$ is a solution of the \omtlarat minimization
[resp. maximization] problem 
\tuple{\enc{},\tuple{obj_i}};

\item[(ii)] $\mu$ lexicographically minimizes [resp. maximizes]
  $\tuple{obj_1,...,obj_k}$ 
if and only if  $\mu$ is a solution of the \omtlarat lexicographic minimization
[resp. maximization] problem 
\tuple{\enc{},\tuple{obj_1,...,obj_k}}.

\end{itemize}
\end{proposition}  

  In short, Proposition \ref{prop:main2} says that $\mu$ is a
  realization for the CGM \calm which optimizes lexicographically 
$\tuple{obj_1,...,obj_k}$ if and only 
 if  $\mu$ is a model in
  \smtlarat for the formula $\Psi_\calm$ which optimizes lexicographically 
$\tuple{obj_1,...,obj_k}$. Therefore, one such realization
  can be found by invoking a \omtlarat solver on 
$\Psi_\calm$ and $\tuple{obj_1,...,obj_k}$.
Notice that we are always looking for {\em one} realization at a time. 
Multiple realizations require multiple calls to the OMT solver.

\ignoreinshort{\subsection{Automated Reasoning on Constrained Goal Models}}
\label{sec:goalmodels_functionalities}
\noindent
Propositions~\ref{prop:main}  and \ref{prop:main2} suggest
 that realizations of a CGM \calm
can be produced by applying \smtlarat solving to the encoding
$\enc{}$, and that {\em optimal} realizations can 
be produced by applying \omtlarat to $\enc{}$ and  a list of
defined objectives $obj_1,...,obj_k$.
{(Notice that such list may include also the pre-defined 
objectives
\weight{},
\numdeniedrequirements{}, \numsatisfiedtasks{} and 
    \numunsatprefs{} of \sref{sec:goalmodels_example} and
    \eqref{eq:costpref}  to be minimized.)} 
This allowed us to  implement straightforwardly the following reasoning
functionalities on CGMs by interfacing with a SMT/OMT tool.

\begin{description} 
\item[{\em Search/enumerate realizations.}] 
  Stakeholders can automatically check the realizability of a CGM \calm{} --or to
  enumerate one or more of its possible realizations-- under a group
  of user assertions and of user-defined Boolean and \smtlarat
  constraints; 
 the tool performs this task by invoking the
 SMT solver on the formula $\Psi_\calm$ of
  Definition~\ref{def:encoding}. 

\item[{\em Search/enumerate minimum-penalty/maximum reward realizations.}]
  Stakeholders 
  can assert the desired requirements and set penalties
  of tasks; then the tool 
  finds automatically realizations 
  achieving the former while minimizing 
  the latter,  by invoking the OMT solver on $\Psi_\calm$ with the pre-defined
  \REQ{Weight} objective. The vice versa is obtained by 
  negating undesired tasks and setting the rewards of
  nice-to-have requirements. 
  Every intermediate situations can be also be obtained.

\item[{\em Search/enumerate optimal realizations wrt. pre-defined/user-defined
  objectives.}] Stakeholders can define their own objective functions
  $obj_1,...,obj_k$ over goals, refinements and their numerical
  attributes; then the tool finds automatically realizations optimizing
  them, either independently or lexicographically, by invoking the OMT
  solver on $\Psi_\calm$ and $obj_1,...,obj_k$.
  {User-defined objectives can also be combined with the pre-defined
    ones,
 like  \REQ{Weight}, \numdeniedrequirements,  \numsatisfiedtasks{} and
 \numunsatprefs. 
}

\end{description}
{\noi In particular, notice that \numunsatprefs allows for addressing the
fulfillment of the maximum number of binary preferences as the
optimization of a pre-defined objective.}
%
\begin{example}
\label{ex:lexicographically}
As a potentially frequent scenario, stakeholders may want to find a realization
which minimizes, in order 
of preference, the number of unsatisfied non-mandatory requirements, 
the number of unsatisfied binary preferences, and the number of
satisfied tasks.  This can be achieved by setting the following
ordered list of pre-defined objectives to minimize lexicographically:
$$
\tuple{\numdeniedrequirements,\numunsatprefs,\numsatisfiedtasks}.
$$
\end{example}  

Notice that all the above actions can be performed {\em interactively} by
marking an unmarking (nice-to-have) requirements, tasks and domain
assumptions, each time searching for a suitable or optimal
realization.

{
Importantly, when a CGM is found un-realizable under a group of user
assertions and of user-defined Boolean and \smtlarat
  constraints, it highlights the
subparts of the CGM and the subset of assertions causing the problem. 
This is implemented by asking the SMT/OMT solver to identify the 
{\em unsatisfiable core} of the input formula  ---i.e. the subset of
sub-formulas which caused the inconsistency, see e.g. \cite{cimattigs11_unsatcore}--- and mapping them back
into the corresponding information.
}

\section{Implementation}
\label{sec:implementation}
\begin{figure}[t]
\centering 
\includegraphics[width=\textwidth]{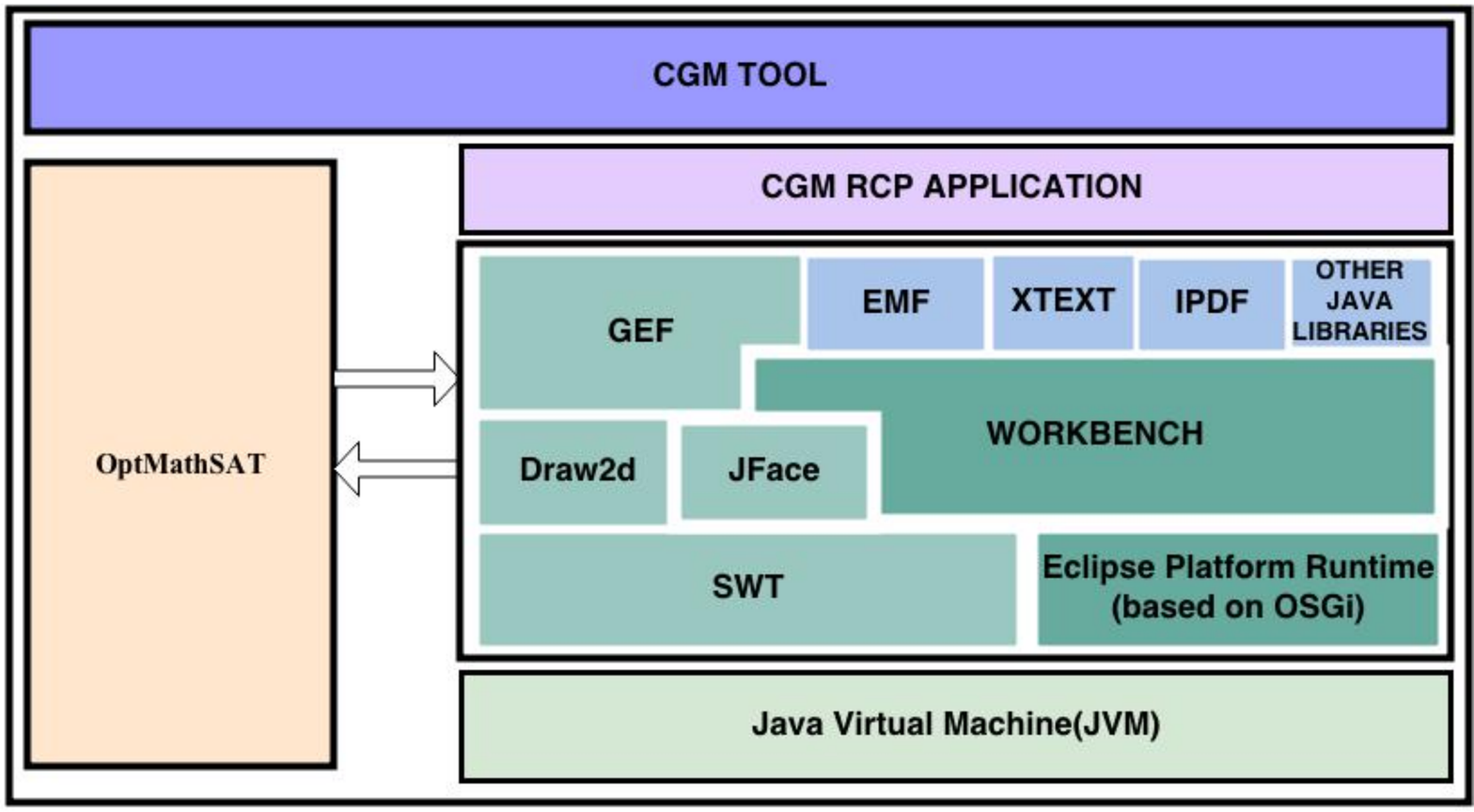}
\caption{\label{figCGM--Tool} CGM-Tool: Component view}
\end{figure}

\begin{figure*}
\centering 
\includegraphics[width=1\textwidth]{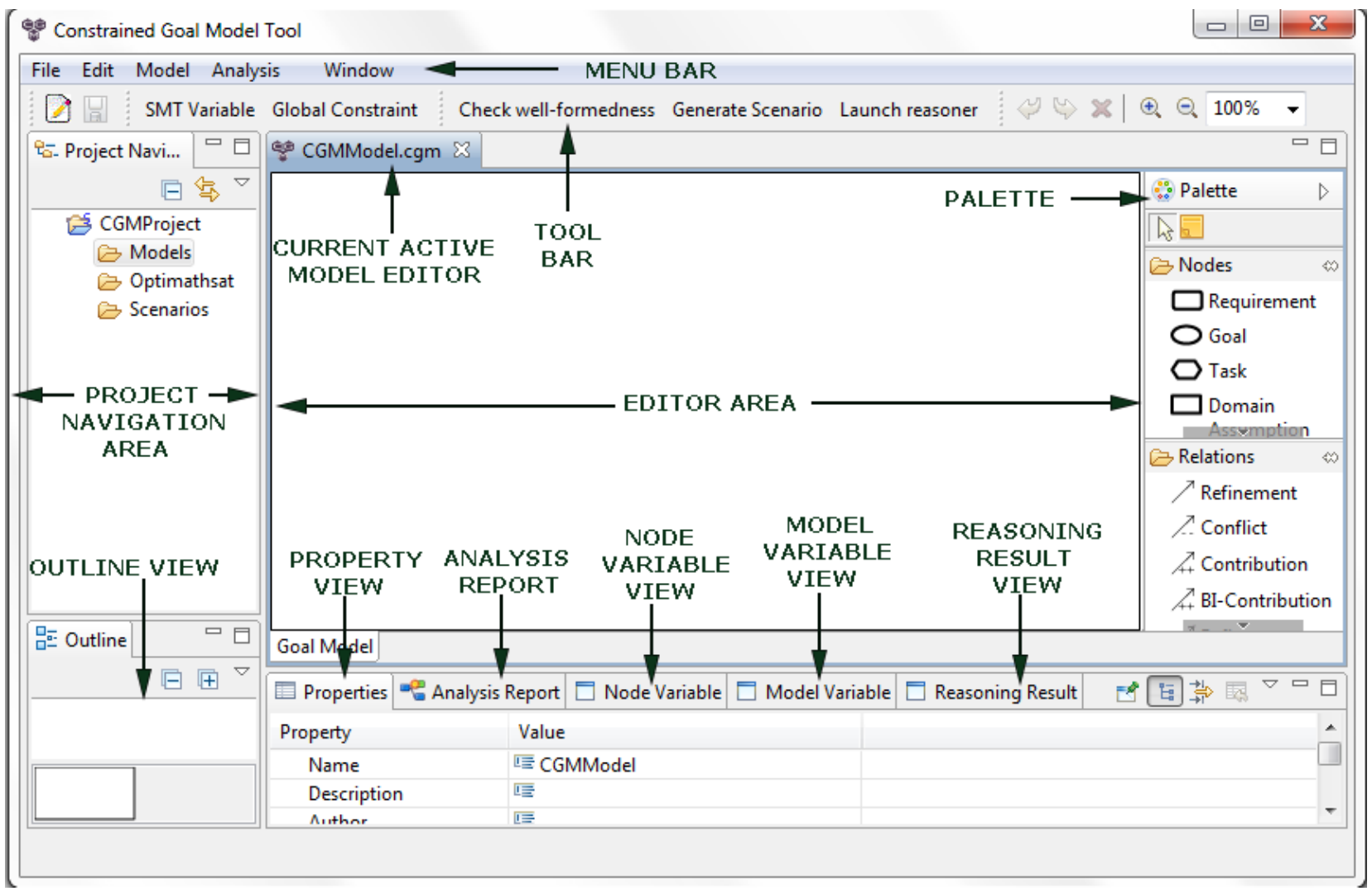}
\caption{\MCCHANGE{CGM-Tool: Graphical User Interface as in the tool manual \cite{CGMManual} (green notes are for description).}
\label{figGUI}}
\end{figure*}

\begin{figure*}
\centering 
\includegraphics[width=0.9\textwidth]{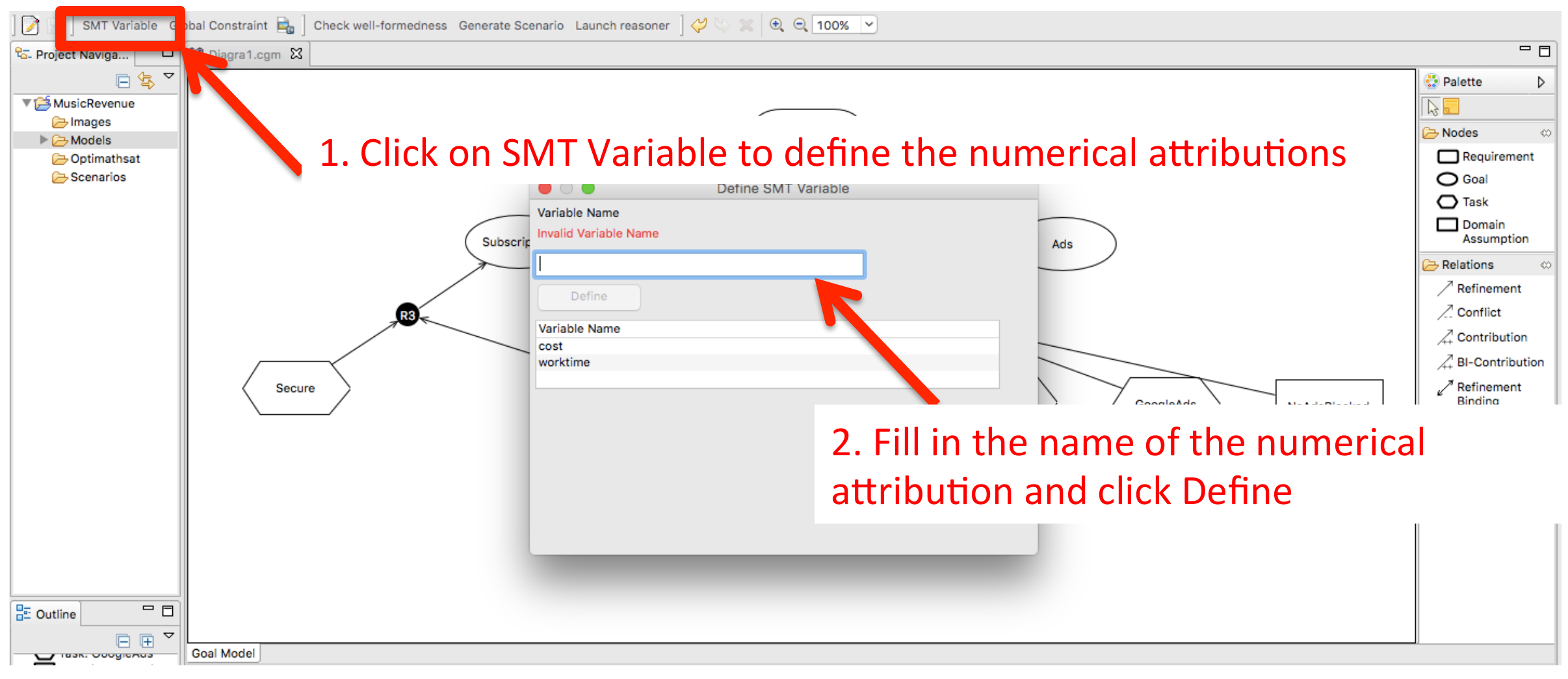}
\caption{\MCCHANGE{CGM-Tool: How to Define Numerical Attributes (instructions in red).}
\label{figGUI_SMT1}}
\end{figure*}

\begin{figure*}
\centering 
\includegraphics[width=0.9\textwidth]{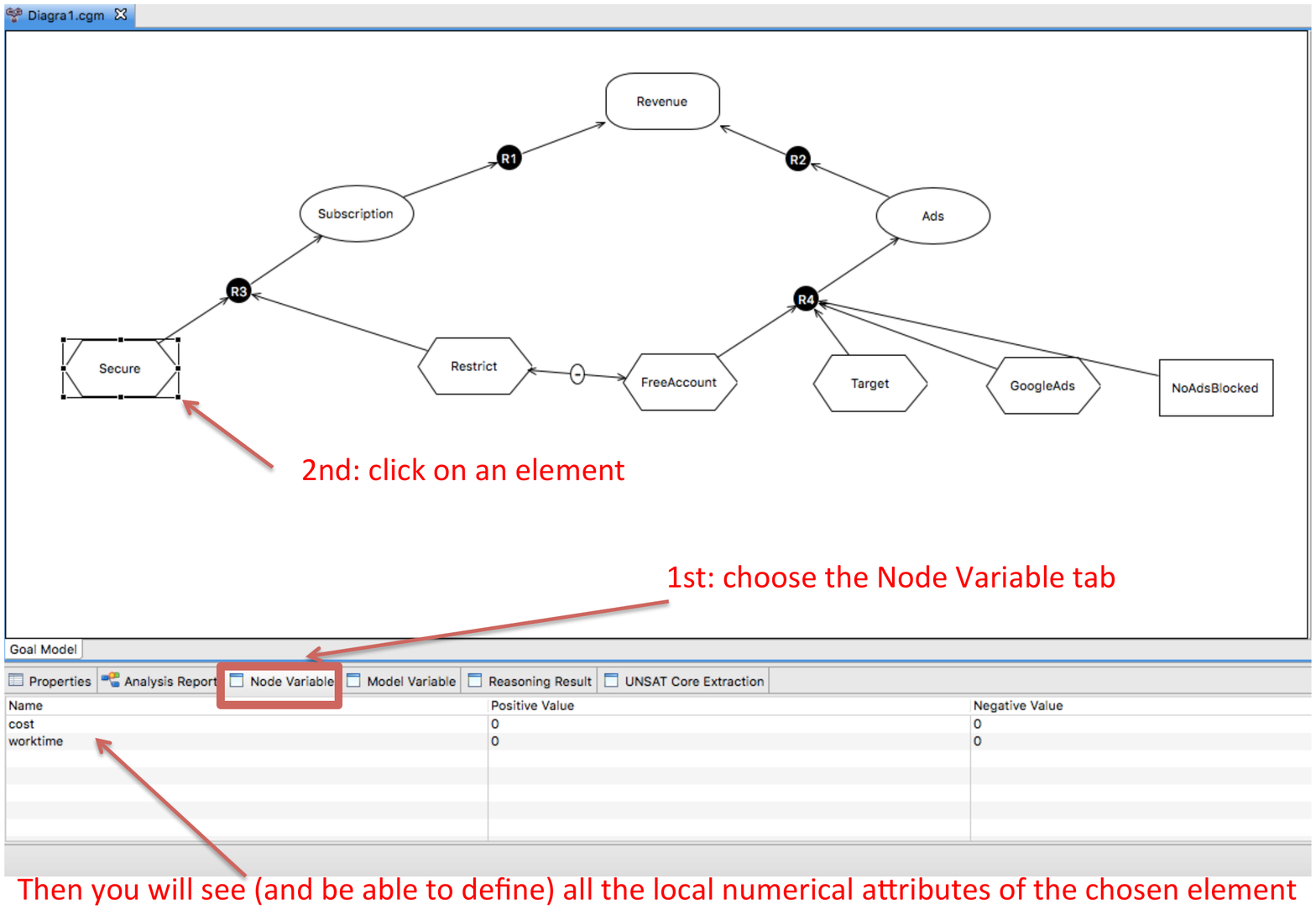}
\caption{\MCCHANGE{CGM-Tool: How to Define the Value of the Numerical Attributes Associated with Elements (instructions in red).}
\label{figGUI_SMT2}}
\end{figure*}

\begin{figure*}
\centering 
\includegraphics[width=0.9\textwidth]{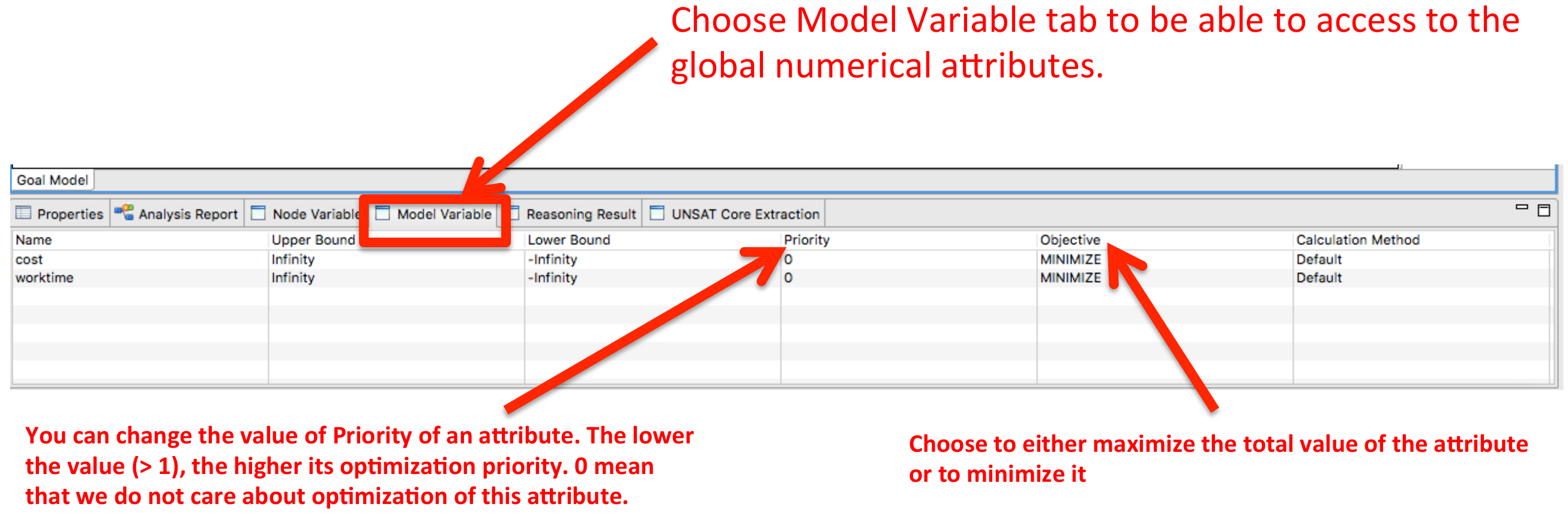}
\caption{\MCCHANGE{CGM-Tool: How to define objectives from Numerical Attributes (instructions in red).}
\label{figGUI_Optimization}}
\end{figure*}

\begin{figure*}
\centering 
\includegraphics[width=0.9\textwidth]{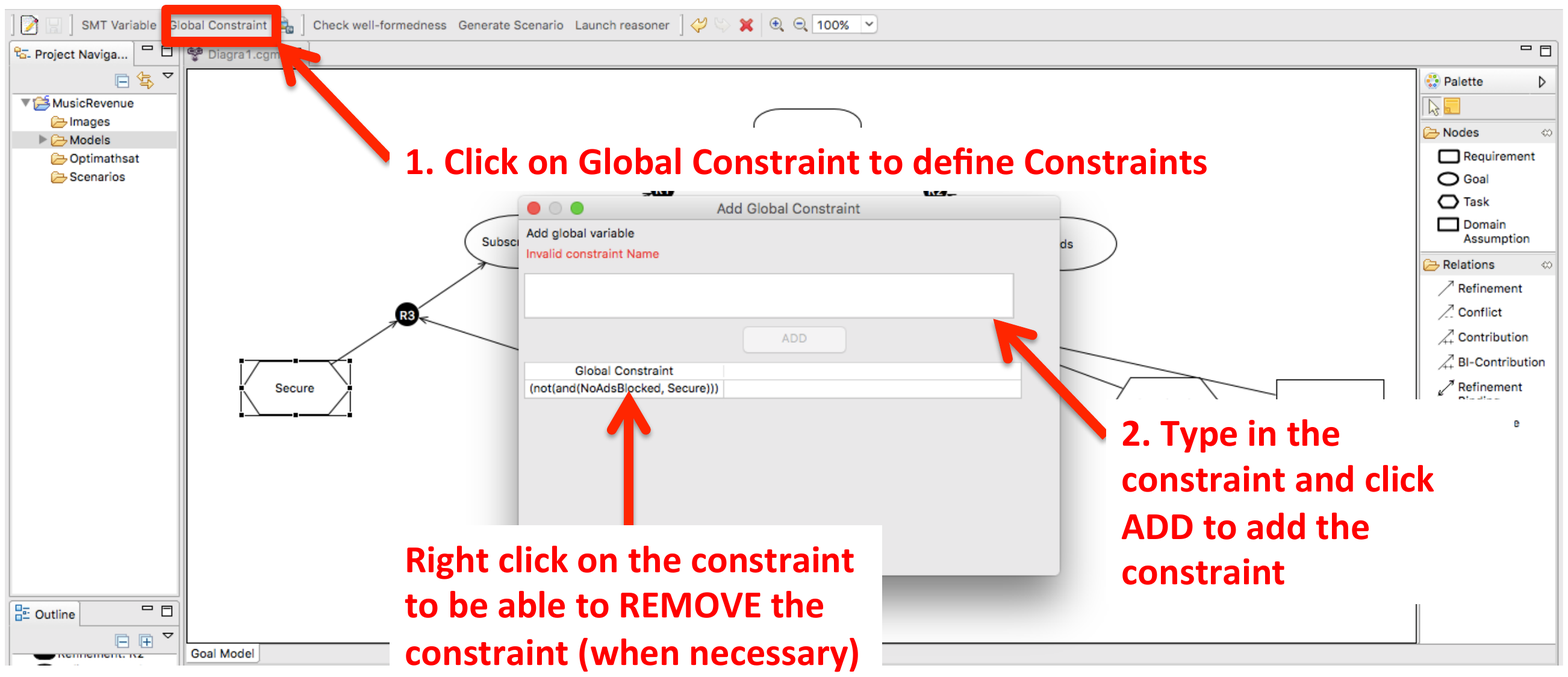}
\caption{\MCCHANGE{CGM-Tool: How to Define Global Constraints (instructions in red).}
\label{figGUI_Constraint}}
\end{figure*}
\begin{figure*}
\centering 
\includegraphics[width=0.8\textwidth]{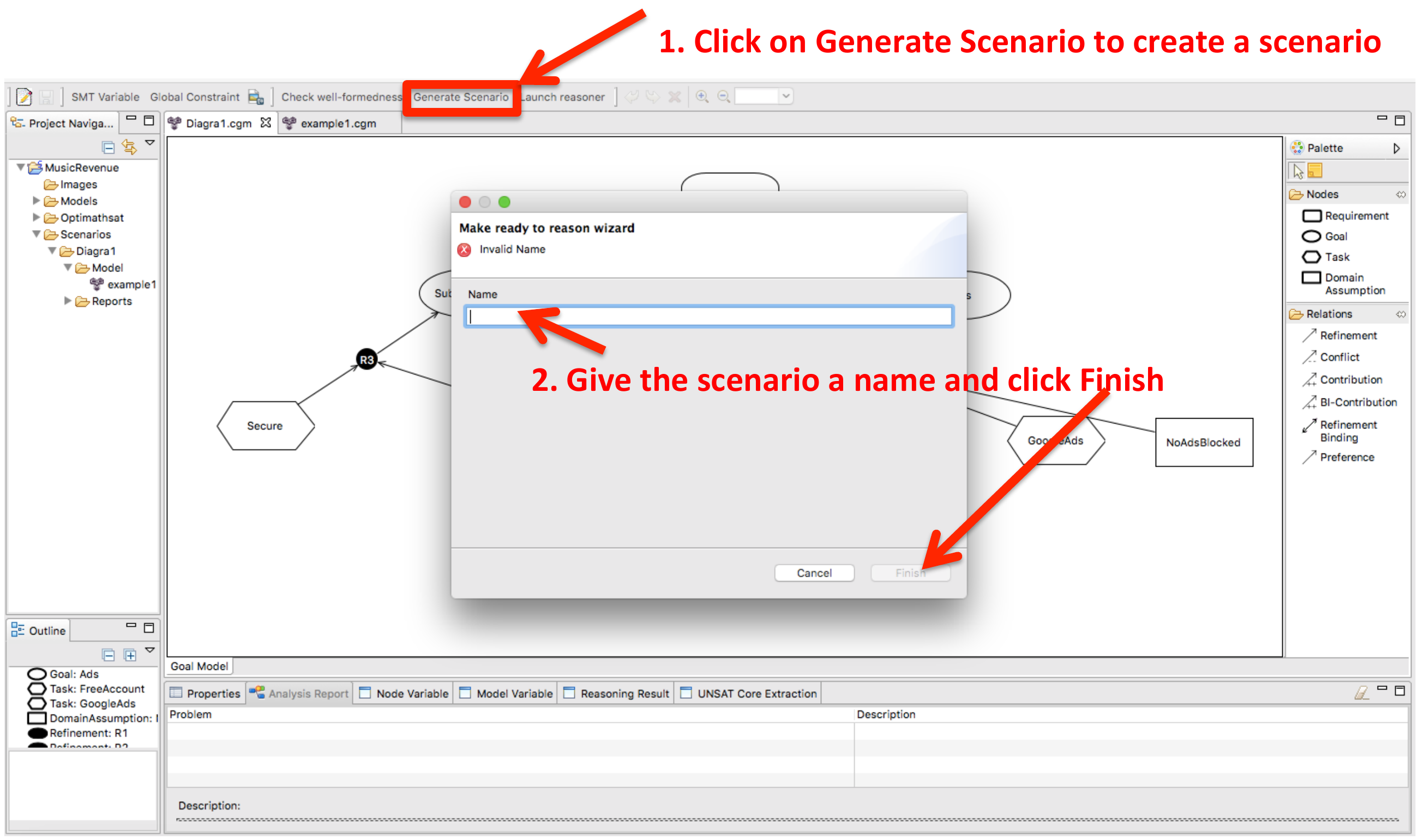}
\caption{\MCCHANGE{CGM-Tool: How to Create a Scenario (instructions in red).}
\label{figGUI_Scenario}}
\end{figure*}
\begin{figure*}
\centering 
\includegraphics[width=0.8\textwidth]{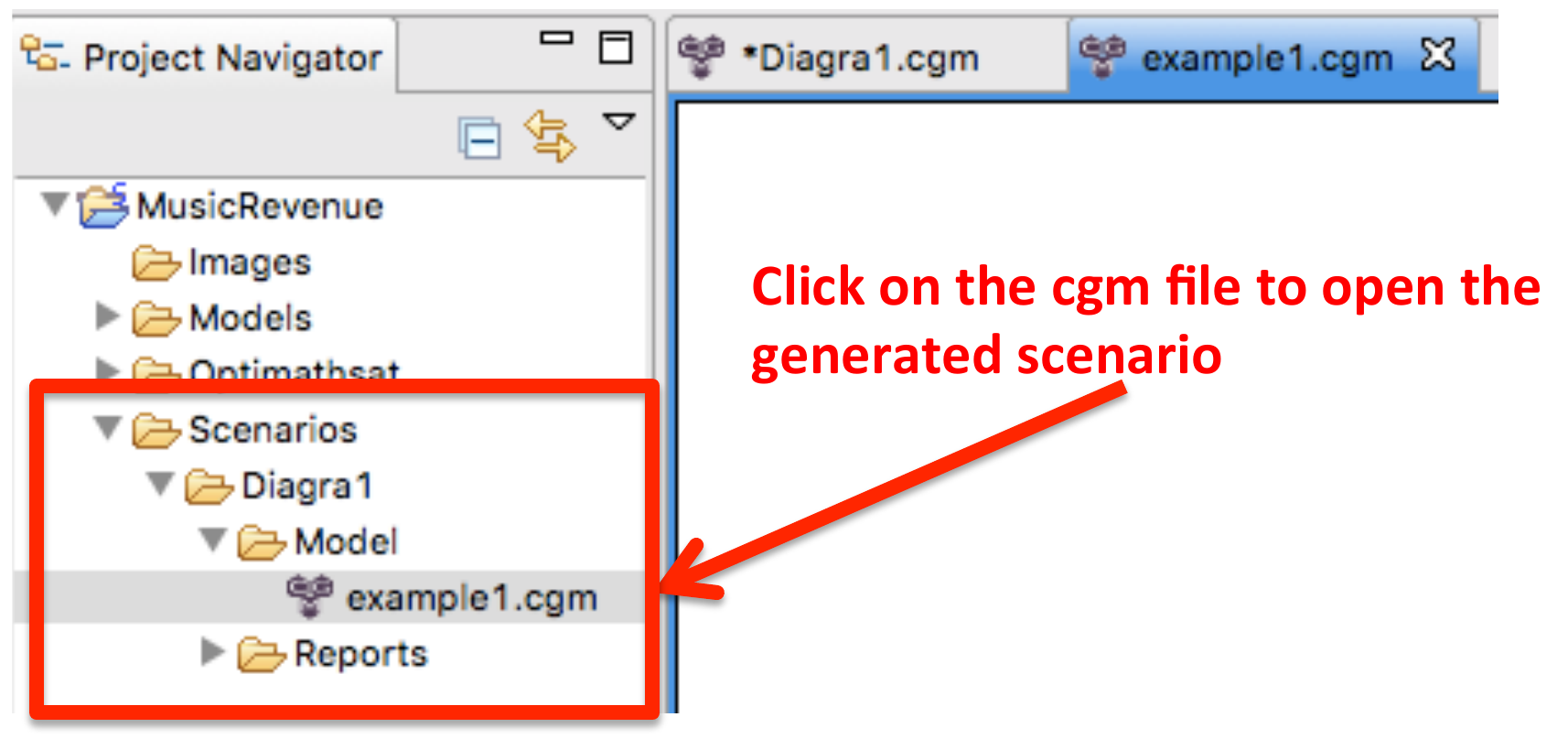}
\caption{\MCCHANGE{CGM-Tool: How to Open the created Scenario (instructions in red).}
\label{figGUI_Scenario2}}
\end{figure*}

\begin{figure*}
\centering 
\includegraphics[width=0.9\textwidth]{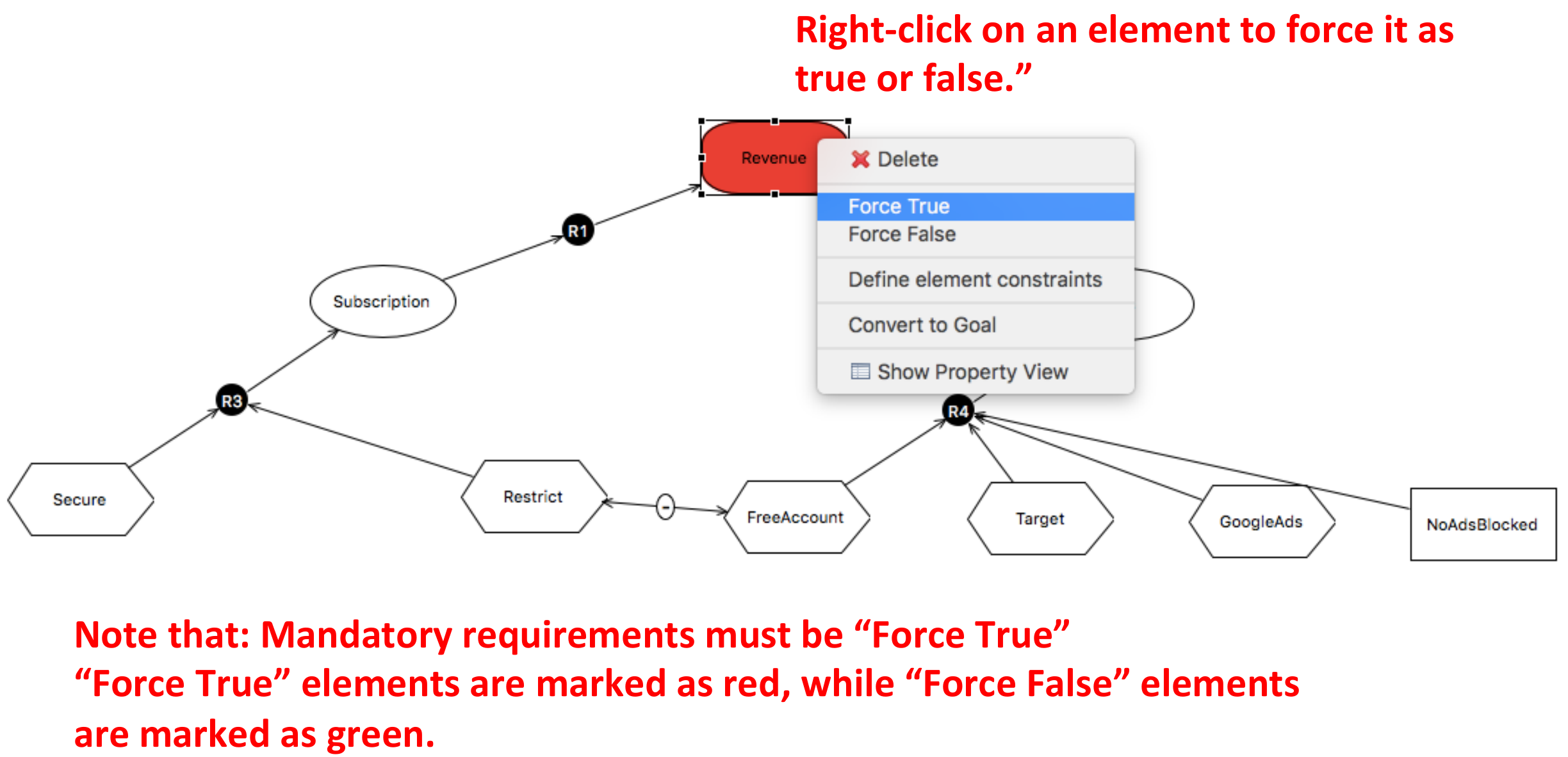}
\caption{\MCCHANGE{CGM-Tool How to Add User's Assertions (instructions in red).
\label{figGUI_UserAssertion}}
}
\end{figure*}

\begin{figure*}
\centering 
\includegraphics[width=0.9\textwidth]{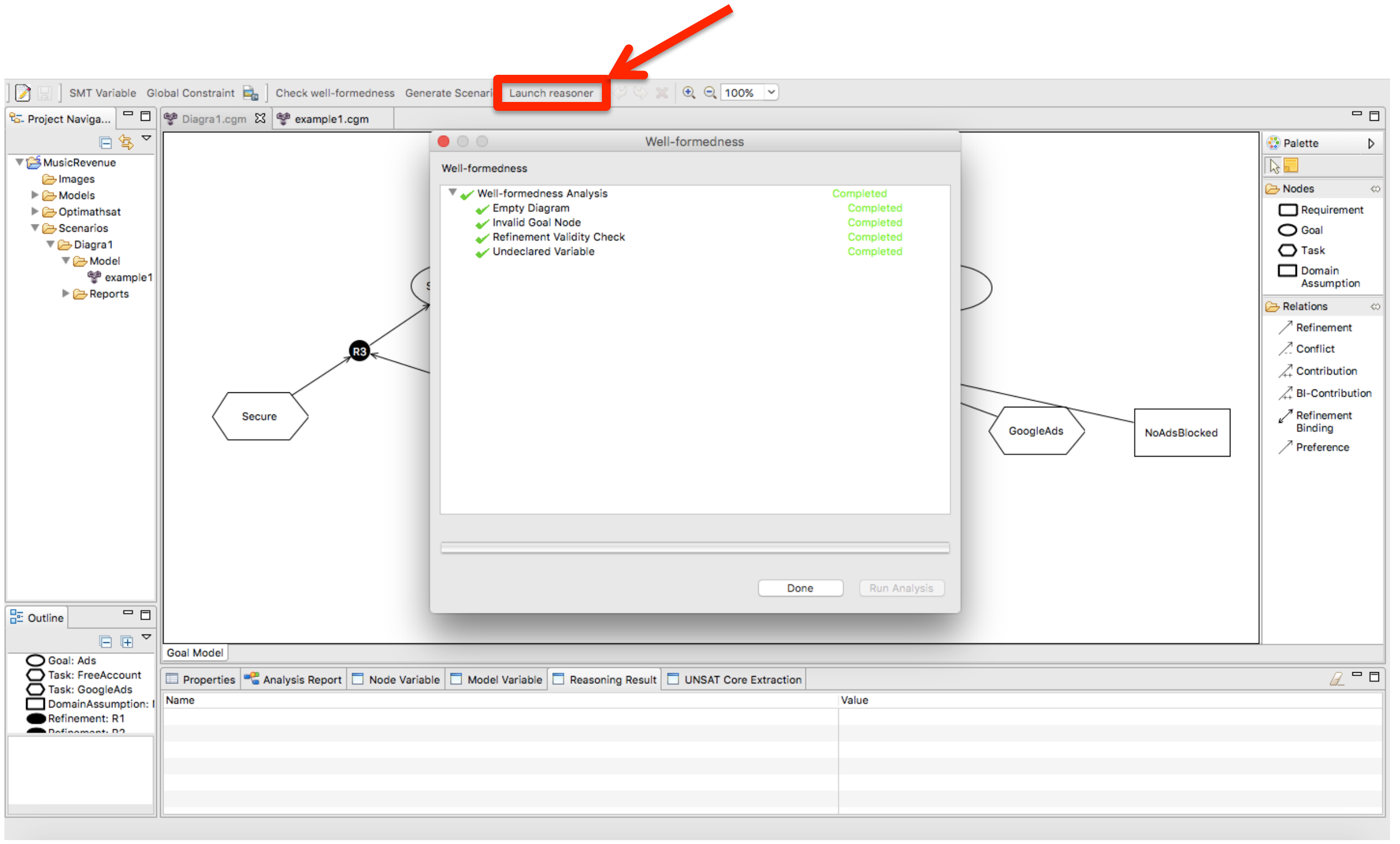}
\caption{\MCCHANGE{CGM-Tool How to Automatically Generate a Realization: click
    on \emph{Launch Reasoner} in the menu
 (instructions in red).
\label{figGUI_Launch}}
}
\end{figure*}

CGM-Tool provides support for modelling and reasoning on
CGMs.  Technically, CGM-Tool is a standalone
application written in Java and its core is based on Eclipse RCP
engine. Under the hood, it encodes CGMs and invokes
the OptiMathSAT%
~\footnote{\url{http://optimathsat.disi.unitn.it}} SMT/OMT solver
 \cite{st_cav15} to support reasoning on goal models. It is freely distributed as a compressed archive file for multiple platforms \footnote{\url{http://www.cgm-tool.eu/}}. CGM-Tool supports: 

\begin{description}
	\item[{\em Specification of projects:}] CGMs
are created within the scope of project containers. A project contains
a set of CGMs that can be used to generate  reasoning sessions with
OptiMathSAT (i.e., scenarios);	

\item[{\em Diagrammatic modelling:}] the tool enables the creation (drawing) of CGMs in terms of diagrams; furthermore it enhances the modelling process by providing real-time check for refinement cycles and by reporting invalid refinement, contribution and binding links;
	
\item[{\em Consistency/well-formedness check:}] CGM-Tool allows for the
  creation of diagrams conform with the semantics of the modelling
  language by providing the ability to run consistency analysis on the
  model;	

\item[{\em Automated Reasoning:}] CGM-Tool provides the automated reasoning
  functionalities of \sref{sec:goalmodels_functionalities} by encoding
  the model into an SMT formula. The results of OptiMathSAT are shown
  directly on the model as well as in a tabular form.

\end{description}

{One essential feature of the  tool is that expressive constructs (which may be more complex and difficult to use) are only available on demand: there are easy-to-use default settings for everything, so that the user can decide the level of expressiveness he/she  feels at ease with.}

CGM-Tool extends the STS-Tool~\cite{paja:2012:ststool:re} as an RCP application by using the major frameworks shown in Figure~\ref{figCGM--Tool}: {\em Rich Client Platform (RCP)}, a platform for building rich client applications, made up of a collection of low level frameworks such as OSGi, SWT, JFace and Equnix, which provide us a workbench where to get things  like menus, editors and views; 
{\em Graphical Editing Framework (GEF)}, a framework used to create
graphical editors for graphical modelling tools (e.g., tool palette and
figures which can be used to graphically represent the underlying data
model concepts); {\em Eclipse Modelling Framework (EMF)}, a modelling
framework and a code generation facility for building tools and
applications based on a structured data model. 

{With CGM-Tool, a CGM is built progressively as a sequence of 
  \new{scenarios}, which are versions of the CGM to which 
the automated reasoning
functionalities of the CGM-Tool can be applied. 
} 
Figure \ref{figGUI} shows the graphical user interface (GUI) of the
tool. Figures \ref{figGUI_SMT1} and \ref{figGUI_SMT2} show
respectively how 
to define a numerical attribute of an element and how to set its value. Figure
\ref{figGUI_Optimization} shows how to set objective functions from
the numerical attributes (e.g., set the priorities, 
choose the form of optimization (maximize/minimize), \ldots). 
Figure \ref{figGUI_Constraint}
shows how to define the global constraints in the model.
{Figure \ref{figGUI_Scenario} and Figure
  \ref{figGUI_Scenario2} show how to create and open a
  scenario. Figure \ref{figGUI_UserAssertion} shows how the user
  assertions can be added by using the option ``Force True" (element
  that must be included in the realization) and ``Force False" (element
  that must not be included in the realization)}. 
{Figure \ref{figGUI_Launch} shows how to automatically
  generate a realization  for the current scenario by invoking the
  automated-reasoning functionalities.}

\section{{
Scalability of
    the Reasoning Tool}}
\label{sec:expeval}
We address the issue of the scalability of the
automated-reasoning functionalities of
\sref{sec:goalmodels_functionalities} wrt. the  size of CGMs,
%
by providing an empirical evaluation of the
performance of CGM-Tool on increasingly-large CGMs.
(For the sake of readability, here we provide only a qualitative
description, whereas the data and plots are reported in an Appendix.)
As in \sref{sec:goalmodels_example}, all experiments have been run on
{a MacBook Air laptop, Intel Core i5 1.8 GHz, 2 cores, 256 KB L2 Cache per Core, 3 MB L3 Cache, 4GB RAM.}

For the {readers}' convenience, a compressed directory containing 
all the material to reproduce these experiments (models, 
tools, scripts, etc.) is available at \url{http://www.cgm-tool.eu/experiment-version/}.

We consider first the schedule-meeting CGM of
\sref{sec:goalmodels_example} as a {seed model}.
The model consists in $32$ goals --among which there are $1$ mandatory
 requirement, $4$ nice-to-have requirements, and $18$ tasks-- plus
 $20$ refinements 
 and $2$ domain assumptions, totaling $54$ nodes.
The CGM contains also $3$
numerical objectives: \REQ{cost}, \REQ{workTime},
and \REQ{Weight}.
The user-defined objectives \REQ{cost} and \REQ{workTime}  involve
respectively $2$ and $5$ tasks and no requirement, whilst the pre-defined 
 attributes \REQ{Weight} involves $16$
tasks plus all $4$ non-mandatory requirements. 
This involves $3+2+5+0+0+16+4=30$  rational variables (recall
Remark~\ref{remark:onlynonzeros}). 
There are also three binary preference relations \eqref{eq:preferences}.

In the example reported in \sref{sec:goalmodels_example}
with different configurations, the tool returned the optimal solutions
in negligible time 
(all took less than $0.02$ seconds).
This is not surprising: as mentioned in \sref{sec:background_goalmodels}, 
in previous empirical evaluation of OMT-encoded problems from formal
verification,  OptiMathSAT 
 successfully handled optimization problems with up
to thousands Boolean/rational variables
\cite{sebastiani15_optimathsat}, so that
 hand-made CGMs resulting into SMT formulas with few tens of Boolean and 
rational variables, like that in \sref{sec:goalmodels_example}, are 
not a computational challenge. 

In perspective, since CGM-Tool is supposed to be used 
to design CGMs representing possibly-large projects, 
we wonder how its
automated-reasoning functionalities will scale on large
models. 
To do this, we choose to build benchmark CGMs of increasing
size, by combining different instances of the schedule-meeting CGM of
\sref{sec:goalmodels_example} in various ways, and testing them with
different combination of objectives. 

\subsection{Experiment Setup.}

In all our experiments CGMs were produced as follows, according to
three positive integer parameters $N$, $k$, and $p$, and some choices
of objectives.

Given $N$  and $k$, we pick $N$ distinct instances of the
schedule-meeting CGM of 
\sref{sec:goalmodels_example}, each with a fresh set of Boolean labels
and
rational variables, we create {an artificial} root goal $G$ with
only one refinement $R$ whose source goals are the $N$ mandatory
requirements ``$\REQ{Schedule
  Meeting}_i$'' of each CGM instance. 
Hence,  the resulting CGM has
$54\cdot N+2$ nodes and $30\cdot N$ rational variables (see
Figure~%
\ref{tabExpData_2}%
).
In another group of experiments 
(see Figure~%
\ref{tabExpData}%
) 
we dropped the non-mandatory requirements and their 4 direct
sub-tasks, so that each instance contains $24$ goals, $2$ domain
assumptions and $18$ refinements, and the resulting CGM has
$44\cdot N+2$ nodes and $26\cdot N$ rational variables.

Then we randomly add $(k-1)\cdot N$ contribution relations 
``\contributes{}{}''
and $N$ conflict relations ``\conflict{}{}'' between tasks 
belonging to different instances.  
When binary preference relations are involved (see below), we also
randomly add $p\cdot N$ binary preference relations, each involving two 
 refinements of one same goal.

In each group of experiments we fix the definition of the objectives
and we set the value of $k$ (and $p$ when it applies), and increase the
values of $N$. For every choice of $N$, we {automatically}~%
\footnote{{To perform this test automatically, 
  we developed an automated problem generator/manipulator
  which interfaces directly with the
  internal data structure representing the CGMs inside CGM-Tool.}}
generate 100 instances
of random problems as in the above schema, which we feed to our tool,
and collect the median CPU times over the solved instances 
--including both encoding and solving
times-- as well as the number of unrealizable instances as
well as the number of instances which OptiMathSAT could not solve
within a timeout of $1000$ seconds. 

Notice that, following some ideas from a different context
\cite{horrocks-igpl,rs-jair03}, 
 the parameters $N$, $k$ and $p$ have been chosen so that to
allow us to {\em increase monotonically} and
{\em tune} some
essential features of the CGMs under test, which may significantly
influence the performances. E.g., 
\begin{itemize}
\item $N$ 
increases linearly the number  of Boolean and rational
variables, 
\item $k$ (and, to some extent, $p$) increases the 
connectivity of the graph and the ratio between 
unrealizable and realizable CGMs.
\item Importantly, $k$ and $p$  also play an essential role in
  drastically reducing 
 the {\em symmetry} of the resulting CGMs, and insert some degree of randomness.
\end{itemize}
Another important parameter, which we borrowed from the schedule-meeting
CGM, is the number of Boolean atoms per objective.

\begin{remark}
  We are aware that the CGMs produced with this 
  approach may not represent {\em realistic} problems.
  However, we stress the fact that here we focus only on providing a
 test on the {\em
    scalability} of our automated-reasoning functionalities. 
\end{remark}

\subsection{Results.}

\begin{rschange}
We run two groups of experiments in which we focus on optimizing, respectively:
\begin{itemize}
\item 
{\em numerical attributes}, like cost,
work-time, penalty/rewards;
\item 
{\em discrete features}, like the number of
binary preferences, of want-to-have
  requirements and of tasks to accomplish.
\end{itemize}
%
\end{rschange}

\begin{figure*}
\centering 
\includegraphics[height=0.95\textheight]{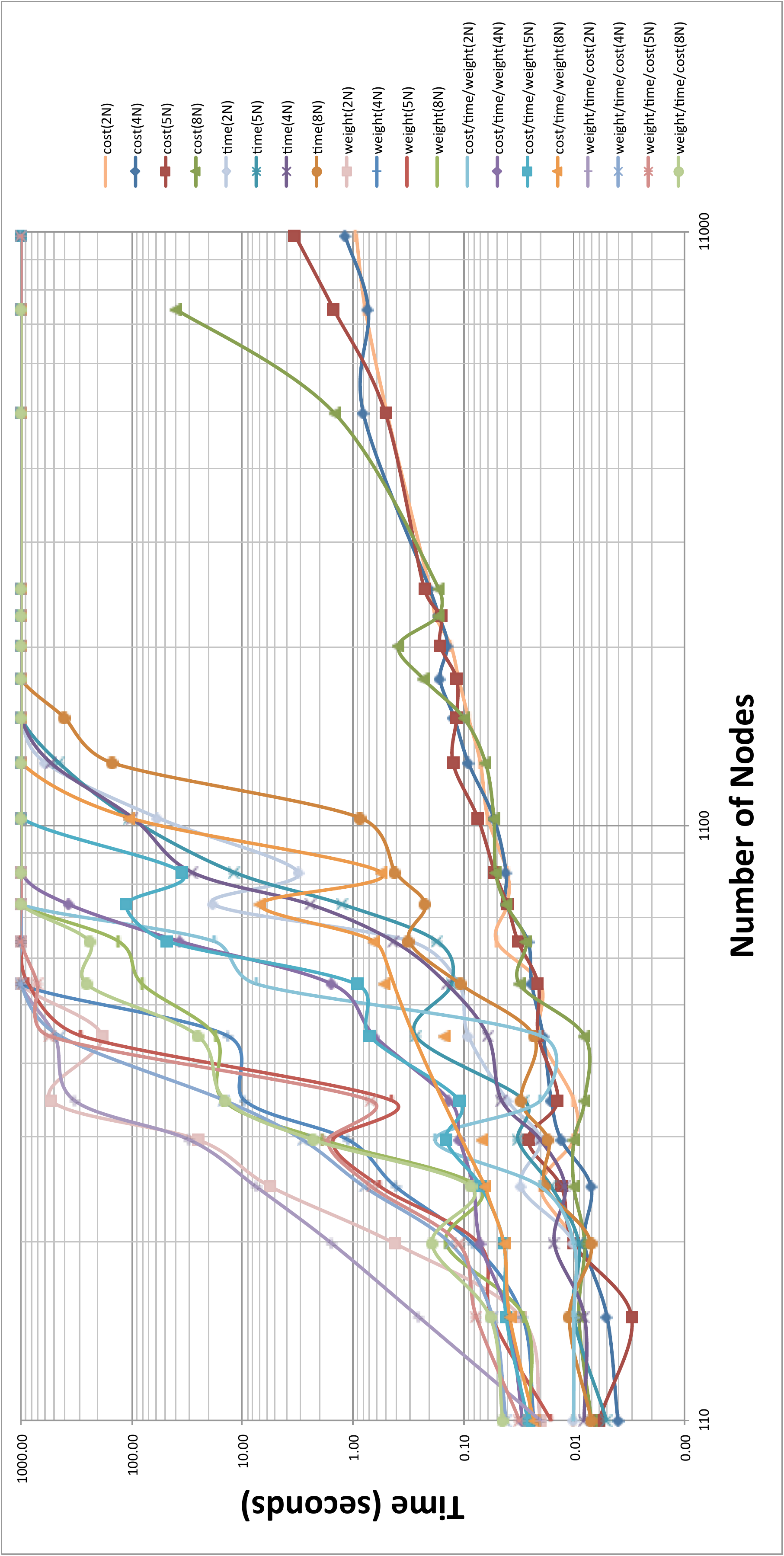}
\caption{{First group of experiments: overall median CPU times over solved instances.}
\RSCHANGE{The name of each plot denotes the cost function
    used and the value of $N$: e.g., {\tt cost/time/weight(2N)}
    detotes the lexicographic optimization of \set{{\sf cost},{\sf
        WorkTime},\weight} on problems built on $N=2$ replicas.
\label{figRuntimeComp}}
}
\end{figure*}

%
In the first group of experiments 
we consider the reduced version of the CGMs (i.e. 
without nice-to-have requirements) 
without random binary preference relations. We fix $k = 2, 4, 5, 8$.
In each setting, we run experiments on three functionalities:
\begin{itemize}
\item[a.] plain realizability check (without objectives),
\item[b.] single-objective optimization on \REQ{cost}, \REQ{workTime},
  and \REQ{Weight} respectively, 
 
\item[c.] lexicographic optimization respectively on
  \tuple{\REQ{cost}, \REQ{workTime}, \REQ{Weight}} and on 
  \tuple{\REQ{Weight}, \REQ{workTime}, \REQ{cost}}. 
\end{itemize}
Figure \ref{figRuntimeComp} shows the overall median CPU time over the solved instances 
of the first group of experiments{, which are plotted against the
total number of nodes of the CGM under test.~%
\footnote{{The choice of using the total number of nodes for
    the X axis in all our plots aims at providing an eye-catching
    indication of the     actual size of the CGMs under test.
}
} (For more details about the experiment data and the median
CPU time over the solved instances for each special case please see Figures~%
\ref{tabExpData1}-%
\ref{figRuntime8N} in the Appendix.)}

First, we notice that checking the realizability of the CGM, that is, 
finding one realization or verifying there is none, requires
negligible time, even with huge CGMs ($>\!8,000$ nodes, $>\!5,000$
rational variables) and even when 
the CGM is not realizable. 
Second, the  time taken to find optimal solutions on single objectives
seem to depend more on the number of variables in the objective than on the
actual size of the CGM: for \REQ{cost} ($2\cdot N$ variables) the solver can
find optimum solutions very quickly even with huge CGMs ($>\!8.000$
nodes, $>\!5,000$
rational variables) whilst with \REQ{Weight} ($16\cdot N$ variables) it can handle problems
of up to
 $~\approx{}400$ nodes and $~\approx{}200$ rational variables. 
Third, lexicographic optimization takes more time than single-objective
optimization, but the time mostly depends on the first objective in
the list. 

\begin{figure*}
\centering 
\includegraphics[height=0.95\textheight]{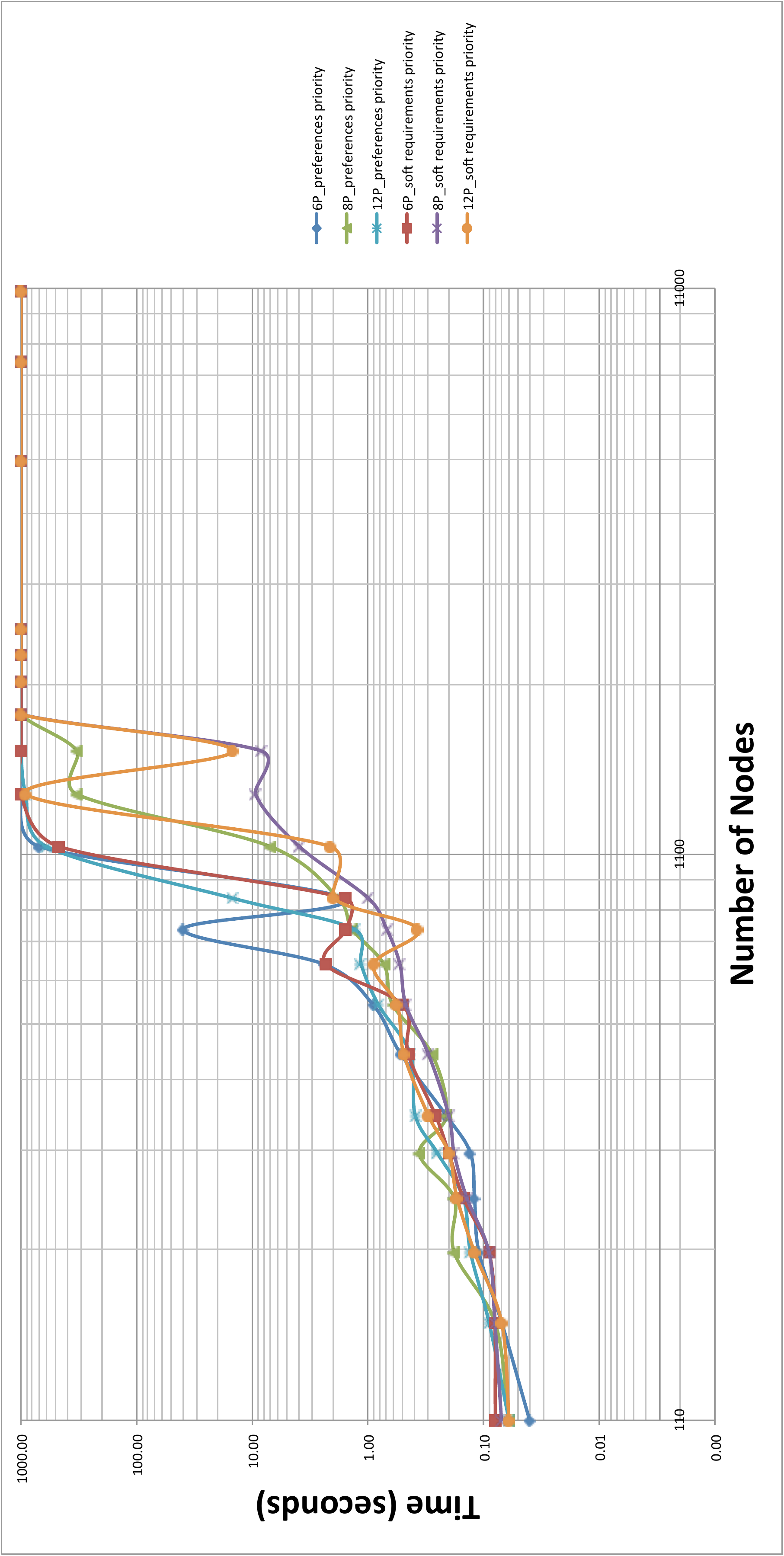}
\caption{{Second group of experiments: overall median CPU times over solved instances.}
\label{figRuntimeComp_Pref}}
\end{figure*}

In the second group of experiments
%
%
we consider the full version of the CGMs (with nice-to-have requirements) 
and introduce the random binary preference relations. 
We fix $k=2$ and we run different experiments for $p=6$, $p=8$ and
$p=12$. In each setting, we run experiments on three functionalities:
\begin{itemize}
\item[a.] plain realizability check (without objectives),
\item[b.] lexicographic optimization on
  \tuple{\numunsatprefs,\numdeniedrequirements,\numsatisfiedtasks} (PRT), 
\item[c.] lexicographic optimization on
  \tuple{\numdeniedrequirements,\numunsatprefs,\numsatisfiedtasks} (RPT).
\end{itemize}
\noindent
Figure \ref{figRuntimeComp_Pref} shows the overall median CPU time over the solved instances 
of the second group of experiments. {(For more details about the experiment data and the median
CPU time over the solved instances for each special case please see Figures~%
\ref{tabExpData11}-%
\ref{figRuntime2N12P} in the Appendix.)}

First, checking realizability is accomplished in negligible time even with
huge CGMs ($>\!10,000$ nodes, $>\!6,000$ rational variables), as before.  
Second, we notice that  optimal solutions, even with a three-level
lexicographic combination of objectives, can be found with large
CGMs  ($>\!1,000$ nodes, $>\!600$ rational variables). 

{ On the negative side, for some problems, in particular
  large ones with objectives involving large amounts of elements, we
  notice that the search for the optimal realization could not be
  accomplished within the timeout.  }

To this extent, a few remarks are in order. 

First, when interrupted by a timeout, OptiMathSAT can be instructed to
return the current best solution. Since OptiMathSAT typically takes most of its
time in fine-tuning the optimum and in checking  there is no better one
(see \cite{sebastiani15_optimathsat}), we envisage that good
sub-optimal solutions can be found even when optimal ones are out of
reach. 

Second, our CGMs are very large in breadth and 
small in depth, with a dominating percentage of tasks over the total
number of goals. 
We envisage that this may have made the number of variables in the
sums defining \REQ{Weight} and \numsatisfiedtasks{} unrealistically
large wrt. the total size of the CGMs.
This underscores the need for
further experimentation to confirm the scalability of our proposal.

Third, in our experiments we did not consider user assertions which,
if considered,
would force deterministic assignments and hence reduce drastically the size
of the OMT search space. 

{%
Fourth, OMT is a recent technology \cite{st-ijcar12} which is progressing at a
very high pace, so that it is reasonable to expect further 
performance improvements for the future versions of OMT tools. 
In particular, a recent enhancement for handling Pseudo-Boolean
cost functions as in \eqref{eq:costpref}
 has provided interesting preliminary results \cite{st_smt16}.
}

\smallskip
Overall, our evaluation showed that CGM-Tool always checks
the realizability of huge CGMs in negligible time and finds optimal
realizations on problems whose size ranges from few hundreds 
to thousands of nodes, mostly depending on the number of variables involved in
the objective functions. 

\begin{figure*}[b] 
\centering
\rotatebox{0}{
\begin{tabular}{rrrrrrrr}
\hline\noalign{\smallskip}
\rot{Experiment} & \rot{Number of Instances} & \rot{Number of Replicas
  (N)} & \rot{Number of Goals} & \rot{Number of Refinements} &
\rot{Number of Domain Assumptions} & \rot{Total Number of Nodes} &
\rot{Number of Rational Variables} \\
\noalign{\smallskip}\hline\noalign{\smallskip}
1 & 100 & 2 & 49 & 37 & 4 & 90 & 52\\
2 & 100 & 3 & 73 & 55 & 6 & 134 & 78\\
3 & 100 & 4 & 97 & 73 & 8 & 178 & 104 \\
4 & 100 & 5 & 121 & 91 & 10 & 222 & 130\\
5 & 100 & 6 & 145 & 109 & 12 & 266 & 156\\
6 & 100 & 7 & 169 & 127 & 14 & 310 & 182\\
7 & 100 & 9 & 217 & 163 & 18 & 398 & 234\\
8 & 100 & 11 & 265 & 199 & 22 & 486 & 286\\
9 & 100 & 13 & 313 & 235 & 26 & 574 & 338\\
10 & 100 & 15 & 361 & 271 & 30 & 662 & 390\\
11 & 100 & 17 & 409 & 307 & 34 & 750 & 442\\
12 & 100 & 21 & 505 & 379 & 42  & 926 & 546\\
13 & 100 & 26 & 625 & 469 & 52 & 1146 & 676\\
14 & 100 & 31 & 745 & 559 & 62 & 1366 & 806\\
15 & 100 & 36 & 865 & 649 & 72 & 1586 & 936\\
16 & 100 & 41 & 985 & 739 & 82 & 1806 & 1066\\
17 & 100 & 46 & 1105 & 829 & 92 & 2026 & 1196\\
18 & 100 & 51 & 1225 & 919 & 102 & 2246 & 1326\\
19 & 100 & 101 & 2425 & 1819 & 202 & 4446 & 2626\\
20 & 100 & 151 & 3625 & 2719 & 302 & 6646 & 3926\\
21 & 100 & 201 & 4825 & 3619 & 402 & 8846 & 5226\\
\noalign{\smallskip}\hline
\end{tabular}
}
\caption{First group of experiments, summary of experimental
  data.\label{tabExpData}
}

\begin{tabular}{rrrrrrrr}
\hline\noalign{\smallskip}
\rot{Experiment} & \rot{Number of Instances} & \rot{Number of Replicas
  (N)} & \rot{Number of Goals} & \rot{Number of Refinements} & \rot{Number of Domain Assumptions} & \rot{Total Number of Nodes} & \rot{Number of Rational Variables} \\
\noalign{\smallskip}\hline\noalign{\smallskip}

1 & 100 & 2 & 65 & 41 & 4 & 110 & 60\\

2 & 100 & 3 & 97 & 61 & 6 & 164 & 90\\

3 & 100 & 4 & 129 & 81 & 8 & 218 & 120 \\

4 & 100 & 5 & 161 & 101 & 10 & 272 & 150\\

5 & 100 & 6 & 193 & 121 & 12 & 326 & 180\\

6 & 100 & 7 & 225 & 141 & 14 & 380 & 210\\

7 & 100 & 9 & 289 & 181 & 18 & 488 & 270\\

8 & 100 & 11 & 353 & 221 & 22 & 596 & 330\\

9 & 100 & 13 & 417 & 261 & 26 & 704 & 390\\

10 & 100 & 15 & 481 & 301 & 30 & 812 & 450\\

11 & 100 & 17 & 545 & 341 & 34 & 920 & 510\\

12 & 100 & 21 & 673 & 421 & 42  & 1136 & 630\\

13 & 100 & 26 & 833 & 521 & 52 & 1406 & 780\\

14 & 100 & 31 & 993 & 621 & 62 & 1676 & 930\\

15 & 100 & 36 & 1151 & 721 & 72 & 1946 & 1080\\

16 & 100 & 41 & 1313 & 821 & 82 & 2216 & 1230\\

17 & 100 & 46 & 1473 & 921 & 92 & 2486 & 1380\\

18 & 100 & 51 & 1633 & 1021 & 102 & 2756 & 1530\\

19 & 100 & 101 & 3233 & 2021 & 202 & 5456 & 3030\\

20 & 100 & 151 & 4833 & 3021 & 302 & 8156 & 4530\\

21 & 100 & 201 & 6433 & 4021 & 402 & 10856 & 6030 \\

\noalign{\smallskip}\hline
\end{tabular}
\caption{Second group of experiments, summary of experimental data.\label{tabExpData_2}
} 
\end{figure*}

\section{Related work}
\label{sec:related}
We next offer a quick overview of, and comparison with some the state
of the art goal-oriented modelling languages. 
{~\cite{Lapouchnian05}, ~\cite{Jureta08}, and ~\cite{Borgida13}} 
provide better
and deeper comparisons on requirements modelling languages and the
goal-oriented approach, including their advantages and limitations. 

\fakesubsubsection{KAOS.}
KAOS~\cite{dardenne93} supports a rich ontology for requirements that goes well beyond goals, as well as a Linear Temporal Logic (LTL)-grounded formal language for constraints. This language is coupled with a concrete methodology for {capturing and analyzing} requirements problems. KAOS supports a number of analysis techniques, including obstacle, inconsistency and probabilistic goal analysis. However, unlike our proposal, KAOS does not support nice-to-have requirements and preferences, nor does it exploit SAT/SMT solver technologies for scalability.

  \fakesubsubsection{Sebastiani et al.}. Sebastiani et al.
  \cite{Giorgini04formalreasoning,sebastiani_caise04} propose a formal
  goal modelling language that supports scalable reasoning using 
  SAT-solving techniques. Our proposal subsumes that work in many
  ways, including a more expressive language and much more advanced
  SMT/OMT-solving technology.

 There is one construct of
 \cite{Giorgini04formalreasoning,sebastiani_caise04} that was left out
 of the CGM language: $+$ and $-$ contributions from goals to
 goals. There are several reasons for this decision.  In
 un-constrained) goal models, formalizing ($+$, $-$) contributions
 require a 4-value logic (fully/partially satisfied/denied). In
 principle our CGM framework could be extended to such a logic, with
 the following drawbacks:
 \begin{aenumerate}
 \item 
The size of the Boolean search space would extend from $2^N$ to
   $4^N$. Given that reasoning functionality in this paper are much
   more sophisticated and computationally more demanding than those in
   our earlier papers{, this might drastically reduce the efficiency 
   of the approach.} 
 \item 
   Unlike standard 2-value logic, which allows us to give a clear
   semantics of ``realization'', without any vagueness, it is not
   obvious to us what a ``realization'' could be in such logic. (E.g.,
   should realizations admit partially satisfied/denied
   tasks/requirements/assumptions?  If yes, how should an user
   interpret a partially-satisfied/denied requirement/task/assumption
   in a realization returned by the system? In which sense a
   realization involving partial values can be considered ``optimal''
   or ``optimum''? 
 \end{aenumerate}

There are other differences between the two
   proposals. In CGMs, we have made and/or-decompositions explicit by
   making refinement a first class citizen that can be named and
   talked about (see Figure~\ref{fig:cgmvsgm} {and Remark~\ref{remark:andor}}). Moreover, unlike with
   \cite{Giorgini04formalreasoning,sebastiani_caise04}, we have a 
   backbone and/or DAG, where arbitrary constraints can be added. This
   DAG is such that a non-leaf goal is equivalent to the disjunction
   (``or'') of its refinements, and each refinement is equivalent to
   the conjunction (``and'') of its source goals. Relation edges,
   constraints and assertions further constrain this structure.

\fakesubsubsection{$I^*$ and Tropos.} 
$i^*$~\cite{Yu97} focuses on modelling actors for a requirements
engineering problem (stakeholders, users, analysts, etc.), their goals
and inter-dependencies. $i^*$ provides two complementary views of
requirements: the Actor Strategic Dependency Model (SD model) and the
Actor Strategic Rationale Model (SR model). Typically, SD models are
used to analyze alternative networks of delegations among actors for
fulfilling stakeholder goals, whilst SR models are used to explore
alternative ways of fulfilling a single actor's goals. $i^*$ is
expressively lightweight, intended for early stages of requirements
analysis, and did not support formal reasoning until recent thesis
work by Horkoff~\cite{Horkoff12}. Tropos~\cite{Castro02} is a
requirements-driven agent-oriented software development methodology
founded on $i^*$. Goal models can be formalized in Tropos by using
Formal Tropos~\cite{Fuxman04}, an extension of $i^*$ that supports
LTL for formalizing constraints. The main deficiencies of this work
relative to our proposal is that Formal Tropos is expressive but not
scalable.  

\fakesubsubsection{Techne and Liaskos.} 
Techne~\cite{JuretaBEM10} is a recent proposal for a
{family} of goal-modelling languages that supports
nice-to-have goals and preferences, but it is strictly propositional
{and uses hand-crafted algorithms}, {and therefore}
{does not support optimization goals.} 
\cite{Ernst2010} constitutes a first attempt to reason with
nice-to-have requirements (aka preferences).  The scalability
experiments conducted used the SAT solver of Sebastiani et
al. \cite{sebastiani_caise04} and added 
local search algorithms to deal with preferences. All experiments
where conducted on a model with about 500 elements and the search
algorithms returned maximal consistent solution but also
near-solutions. \cite{ErnstBJ11} focuses on finding new solutions for
a goal model that has changed  (new goals were added/removed), such
that the change minimizes development  effort (EvoR1) or maximizes
familiarity (EvoR2). Note that EvoR1, EvoR2 are  evolution
requirements. The paper uses a Truth-Maintenance System (TMS) and
builds algorithms on top for finding solutions to EvoR1, EvoR2 that
``repair'' the previous solution and construct a new one. The search
algorithms would need to be redone if we used different evolution
algorithms, unlike  the CGM tool where you can formally express EvoR1,
EvoR2 or variants, and  search is handle by the backend OMT/SMT
solver. \cite{ErnstBMJ12,Ernst2014} continue the study of reasoning
with Techne models and use SAT solvers and hand-crafted search
algorithms to establish scalability for models size O(1K). Nevertheless the
resulting tools from this work still can't handle quantitative
optimization problems and other features of CGMs. 


Liaskos~\cite{Liaskos10,Liaskos12} has proposed extensions to qualitative goal models to support nice-to-have goals and preferences, as well as decision-theoretic concepts such as utility. This proposal is comparable to our proposal in this paper, but uses AI reasoners for reasoning (AI planners and GOLOG) and, consequently, does not scale very well relative to our proposal.

\fakesubsubsection{Feature Models.}
Feature models~\cite{ClassenBH11} share many similarities with goal models: they are hierarchically structured, with AND/OR refinements, constraints and attributes. However, each feature represents a bundle of functionality or quality and as such, feature models are models of software {configurations}, not requirements. Moreover, reasoning techniques for feature models are limited relative to their goal model cousins.

\fakesubsubsection{Search-Based Software Engineering.} 
Scalable reasoning for optimization problems has been studied by
Harman et al in the context of formalizing and solving the next
release problem~\cite{Yuanyuan}: given a set of preferences with
associated cost and customer value attributes, select a subset {of
preferences} to be included in the next release that optimizes given
attributes. That work uses genetic algorithms and other search
techniques that may return close-to-optimal solutions and use
heuristics (meaning that reasoning is not complete).

\section{Conclusions and Future Work}
\label{sec:concl} 
We have proposed, {an expressive} goal-based modelling language for requirements that
supports the representation of nice-to-have requirements, preferences, optimization requirements,
constraints and more. Moreover, we have exploited automated reasoning
solvers in order to develop a tool that {supports sound and complete reasoning with respect to such goal models, and} scales well {to goal models with thousands of elements. Our proposal advances the state-of-the-art on goal modelling and reasoning with respect to both expressiveness and scalability of reasoning.}

\ignore{
Notice that, the goal model concept has been accepted and widely used 
in software engineering as well as requirements modelling for decades. 
Our claim in this paper is not to introduce a completely new modelling concept, 
rather our main contribution lies in the expressiveness of the new formalization 
of the constrained goal model and the automated reasoning functionalities of our CGM-Tool.
}
The contributions of this work are being exploited in several directions. \cite{Aydemir} has proposed an expressive modelling framework for the next release problem that is founded on the same OMT/SMT solver technology as this work. \cite{Angelopoulos16} has offered a formalization of the next adaptation problem that chooses a next adaptation for an adaptive software system that minimizes the degree of failure over existing requirements. And \cite{Nguyen16} has exploited CGMs to capture evolution requirements, such as ``System evolution shall minimize implementation costs" and showed how to conduct scalable reasoning over models that include such requirements.

As future work, we have planned to do an empirical validation of the CGM-Tool 
with modelers and domain experts. We are currently working in this direction 
within our research group with PhD students and post-docs who are expert 
in the modelling field. Next, we will extend the validation to industrial experts 
of different domains. We have also planned to do different case studies with 
real-life-complex-large-scale goal models of a specific domain, such as 
Air-Traffic Control Management, healthcare, and smart cities and smart environments.

\ignore{
In the future, we plan to formalize \new{evolutionary} versions of
CGMs, which can handle evolution requirements problems
~\cite{ErnstBJ11}. For such problems, the goal models can be changed
even after a solution has been implemented. Thus, given a modified CGM
and its previous solution, we have to find a new solution that
minimize the effort of applying changes. {Again, we plan to address
this problem via OMT encodings.}
}

  Our proposal does not address another notorious scalability problem
  of goal models, namely scalability-of-use. Goal models have been
  shown empirically to become more difficult to conceptualize and
  comprehend as they grow in size \cite{Estrada2006}, and therefore
  become unwieldy for use. As with other kinds of artifacts (e.g.,
  programs, ontologies) where scalability-of-use is an issue, the
  solution lies in introducing modularization facilities that limit
  interactions between model elements and make the resulting models
  easier to understand and evolve. This is an important problem on our
  agenda for future research on goal models.

\ignore{
\begin{tabular}{lll}
\hline\noalign{\smallskip}
first & second & third  \\
\noalign{\smallskip}\hline\noalign{\smallskip}
number & number & number \\
number & number & number \\
\noalign{\smallskip}\hline
\end{tabular}
\end{table}
}

\ignoreinshort{
\begin{acknowledgements}
We would {like to} thank Dagmawi
Neway  for his technical support in developing CGM-Tool, and Patrick
Trentin for assistance with the usage of OptiMathSAT. 
\end{acknowledgements}
}

\bibliographystyle{spmpsci} 
\bibliography{pg,mc,rs_refs,rs_ownrefs,rs_specific,sathanbook}


\newpage
\appendix
\section{Appendix: Data Tables and Plots}
%
%
\FloatBarrier
\subsection{First Group of Experiments}
\begin{figure*}[b] 
\centering
\resizebox{\textwidth}{!}{
\begin{tabular}{rrrrrrrrrrrrrrrrrr}
\hline\noalign{\smallskip}
&&&&&&&& \multicolumn{2}{r}{Optimum} & \multicolumn{2}{r}{Optimum} & \multicolumn{2}{r}{Optimum} & \multicolumn{2}{r}{Lexic. Order} & \multicolumn{2}{r}{Lexic. Order} \\
&&&&&&&&\multicolumn{2}{r}{cost} & \multicolumn{2}{r}{time} & \multicolumn{2}{r}{weight} &\multicolumn{2}{r}{cost time weight}&\multicolumn{2}{r}{weight time cost}\\
&&&&&&&&\multicolumn{2}{r}{(2N terms)} & \multicolumn{2}{r}{(5N terms)} & \multicolumn{2}{r}{(16N terms)} &&&&\\
\rot{Experiment} & \rot{Number of Instances} & \rot{Number of Replicas  (N)} & \rot{Total Number of Nodes} & \rot{Number of Rational Variables} & \rot{\% Unrealizable} & \rot{Solving Time} & \rot{Time for Proving Unrealizable}&\rot{Optimization Time} & \rot{\% Timeout} & \rot{Optimization Time} & \rot{\% Timeout} & \rot{Optimization Time} & \rot{\% Timeout} & \rot{Optimization Time} & \rot{\% Timeout} & \rot{Optimization Time} & \rot{\% Timeout}\\
\noalign{\smallskip}\hline\noalign{\smallskip}
1   & 100 & 2     & 90      & 52 &   1  & 0.00 & 0.00 & 0.00 & 0 & 0.01     & 0   & 0.02     & 0   & 0.01     & 0   & 0.02 & 0\\

2   & 100 & 3     & 134    & 78 & 1  & 0.00 & 0.00 & 0.01 & 0 & 0.01     & 0   & 0.03     & 0   & 0.01     & 0   & 0.25 & 0\\

3   & 100 & 4     & 178    & 104 &  3  & 0.00 & 0.00 & 0.01 & 0 & 0.01     & 0   & 0.41     & 0   & 0.01     & 0   & 1.53 & 0\\

4   & 100 & 5     & 222    &  130 & 3  & 0.00 & 0.00 & 0.02 & 0 & 0.03     & 0   & 5.51     & 0   & 0.02     & 0   & 7.48 & 0\\

5   & 100 & 6     & 266    &  156 & 2  & 0.00 & 0.00 & 0.01 & 0 & 0.02     & 0   & 24.74   & 0   & 0.18     & 0   & 29.74 & 2\\

6   & 100 & 7     & 310    &  182 & 5  & 0.00 & 0.00 & 0.01 & 0 & 0.04     & 0   & 533.90 & 19 & 0.02     & 0   & 329.34 & 30\\

7   & 100 & 9     & 398    &  234 & 7  & 0.00 & 0.00 & 0.02 & 0 & 0.09     & 0   & 185.23 & 84 & 0.02     & 4   & 494.33 & 87\\

8   & 100 & 11   & 486    &  286 & 4  & 0.00 & 0.00 & 0.02 & 0 & 0.11     & 0   &     ---     & --- & 7.29     & 30 &     ---     & ---\\

9   & 100 & 13   & 574    & 338 &  7  & 0.00 & 0.00 & 0.05 & 0 & 0.30     & 0   &     ---     & --- & 17.98   & 83 &     ---     & ---\\

10 & 100 & 15   & 662    & 390 & 13  & 0.00 & 0.00 & 0.04 & 0 & 18.13   & 0   &     ---     & --- &     ---     & --- &     ---     & ---\\

11 & 100 & 17   & 750    & 442 & 15  & 0.00 & 0.00 & 0.04 & 0 & 3.11      & 0  &     ---     & --- &     ---     & --- &     ---     & ---\\

12 & 100 & 21   & 926    & 546 & 14  &  0.00 & 0.00 & 0.06 & 0 & 58.08   & 11 &     ---     & --- &     ---     & --- &     ---     & ---\\

13 & 100 & 26   & 1146  & 676 & 13  & 0.00 & 0.00 & 0.07 & 0 & 600.99 & 78 &     ---     & --- &     ---     & --- &     ---     & ---\\

14 & 100 & 31   & 1366  & 806 & 14  & 0.00 & 0.00 & 0.09 & 0 &    ---     & ---  &    ---     & --- &     ---     & --- &     ---     & ---\\

15 & 100 & 36   & 1586  & 936 & 19  & 0.00 & 0.00 & 0.11 & 0 &    ---     & ---  &    ---     & --- &     ---     & --- &     ---     & ---\\

16 & 100 & 41   & 1806  & 1066 & 26  & 0.00 & 0.00 & 0.13 & 0 &    ---     & ---  &    ---     & --- &     ---     & --- &     ---     & ---\\

17 & 100 & 46   & 2026  & 1196 & 24  & 0.00 & 0.00 & 0.18 & 0 &    ---     & ---  &    ---     & --- &     ---     & --- &     ---     & ---\\

18 & 100 & 51   & 2246  & 1326 & 32  & 0.00 & 0.00 & 0.20 & 0 &    ---     & ---  &    ---     & --- &     ---     & --- &     ---     & ---\\

19 & 100 & 101 & 4446  & 2626 & 49  & 0.00 & 0.00 & 0.49 & 0 &    ---     & ---  &    ---     & --- &     ---     & --- &     ---     & ---\\

20 & 100 & 151 & 6646  & 3926 & 68  & 0.00 & 0.00 & 0.77 & 0 &    ---     & ---  &    ---     & --- &     ---     & --- &     ---     & ---\\

21 & 100 & 201 & 8846  & 5226 & 71  & 0.00 & 0.00 & 0.93 & 0 &    ---     & ---  &    ---     & --- &     ---     & --- &     ---     & ---\\
\noalign{\smallskip}\hline
\end{tabular}
}
\caption{First group of experiments, $k=2$: median time over solved
  instances.\label{tabExpData1}
} 
\resizebox{\textwidth}{!}{
\begin{tabular}{rrrrrrrrrrrrrrrrrr}
\hline\noalign{\smallskip}
&&&&&&&& \multicolumn{2}{r}{Optimum} & \multicolumn{2}{r}{Optimum} & \multicolumn{2}{r}{Optimum} & \multicolumn{2}{r}{Lexic. Order} & \multicolumn{2}{r}{Lexic. Order} \\
&&&&&&&&\multicolumn{2}{r}{cost} & \multicolumn{2}{r}{time} & \multicolumn{2}{r}{weight} &\multicolumn{2}{r}{cost time weight}&\multicolumn{2}{r}{weight time cost}\\
&&&&&&&&\multicolumn{2}{r}{(2N terms)} & \multicolumn{2}{r}{(5N terms)} & \multicolumn{2}{r}{(16N terms)} &&&&\\
\rot{Experiment} & \rot{Number of Instances} & \rot{Number of Replicas  (N)} & \rot{Total Number of Nodes} & \rot{Number of Rational Variables} & \rot{\% Unrealizable} & \rot{Solving Time} & \rot{Time for Proving Unrealizable}&\rot{Optimization Time} & \rot{\% Timeout} & \rot{Optimization Time} & \rot{\% Timeout} & \rot{Optimization Time} & \rot{\% Timeout} & \rot{Optimization Time} & \rot{\% Timeout} & \rot{Optimization Time} & \rot{\% Timeout}\\
\noalign{\smallskip}\hline\noalign{\smallskip}
1 & 100 & 2 & 90 & 52 & 2 & 0.00 & 0.00 & 0.00 & 0 & 0.01 & 0 & 0.02 & 0 & 0.03 & 0 & 0.04 & 0 \\

2 & 100 & 3 & 134 & 78 & 1 & 0.00 & 0.00 & 0.01 & 0 & 0.01 & 0 & 0.03 & 0 & 0.04 & 0 & 0.05 & 0\\

3 & 100 & 4 & 178 & 104 & 2 & 0.00 & 0.00 & 0.01 & 0 & 0.02 & 0 & 0.08 & 0 & 0.06 & 0 & 0.13 & 0\\

4 & 100 & 5 & 222 & 130 & 4 & 0.00 & 0.00 & 0.01 & 0 & 0.01 & 0 & 0.41 & 0 & 0.08 & 0 & 0.79 & 0 \\

5 & 100 & 6 & 266 & 156 & 7 & 0.00 & 0.00 & 0.01 & 0 & 0.02 & 0 & 1.09 & 0 & 0.11 & 0 & 2.82 & 0\\

6 & 100 & 7 & 310 & 182 & 7 & 0.00 & 0.00  & 0.02 & 0 & 0.05 & 0 & 9.53 & 4 & 0.13 & 0 & 14.47 & 4\\
  
7 & 100 & 9 & 398 & 234 & 7 & 0.00 & 0.00  & 0.02 & 0 & 0.06 & 0 & 13.54 & 56 & 0.64 & 0 & 447.13 & 67\\
  
8 & 100 & 12 & 486 & 286 & 10 & 0.00 & 0.00  & 0.02 & 0 & 0.14 & 0 & --- & --- & 1.53 & 5 & --- & ---\\
  
9 & 100 & 13 & 574 & 338 & 9 & 0.00  & 0.00 & 0.03 & 0 & 0.43 & 0 & --- & --- & 36.78 & 23 & --- & ---\\
  
10 & 100 & 15 & 662 & 390 & 11 & 0.00 & 0.00  & 0.04 & 0 & 2.42 & 0 & --- & --- & 368.94 & 55 & --- & ---\\
  
11 & 100 & 17 & 750 & 442 & 9 & 0.00  & 0.00 & 0.04 & 0 & 28.45 & 0 & --- & --- & --- & --- & --- & ---\\
  
12 & 100 & 21 & 926 & 546 & 15 & 0.00  & 0.00 & 0.05 & 0 & 97.86 & 2 & --- & --- & --- & --- & --- & ---\\
  
13 & 100 & 26 & 1146 & 676 & 22 & 0.00  & 0.00 & 0.09 & 0 & 537.73 & 62 & --- & --- & --- & --- & --- & ---\\
  
14 & 100 & 31 & 1366 & 806 & 25 & 0.00  & 0.00 & 0.12 & 0 & --- & --- & --- & --- & --- & --- & --- & ---\\
  
15 & 100 & 36 & 1586 & 936 & 27 & 0.00  & 0.00 & 0.16 & 0 & --- & --- & --- & --- & --- & --- & --- & ---\\
  
16 & 100 & 41 & 1806 & 1066 & 32 & 0.00  & 0.00 & 0.14 & 0 & --- & --- & --- & --- & --- & --- & --- & ---\\
  
17 & 100 & 46 & 2026 & 1196 & 36 & 0.00  & 0.00 & 0.17 & 0 & --- & --- & --- & --- & --- & --- & --- & ---\\
  
18 & 100 & 51 & 2246 & 1326 & 40 & 0.00  & 0.00 & 0.20 & 0 & --- & --- & --- & --- & --- & --- & --- & ---\\
  
19 & 100 & 101 & 4446 & 2626 & 55 & 0.00  & 0.00 & 0.80 & 0 & --- & --- & --- & --- & --- & --- & --- & --- \\
  
20 & 100 & 151 & 6646 & 3926 & 77 & 0.00  & 0.00 & 0.72 & 0 & --- & --- & --- & --- & --- & --- & --- & ---\\
  
21 & 100 & 201 & 8846 & 5226 & 85 & 0.00  & 0.00 & 1.18 & 0 & --- & --- & --- & --- & --- & --- & --- & ---\\
  
\noalign{\smallskip}\hline
\end{tabular}
}
\caption{First group of experiments, $k=4$: median time over solved instances.\label{tabExpData2}
}
\end{figure*}

\begin{figure*}[t] 
\centering
\resizebox{\textwidth}{!}{
\begin{tabular}{rrrrrrrrrrrrrrrrrr}
\hline\noalign{\smallskip}
&&&&&&&& \multicolumn{2}{r}{Optimum} & \multicolumn{2}{r}{Optimum} & \multicolumn{2}{r}{Optimum} & \multicolumn{2}{r}{Lexic. Order} & \multicolumn{2}{r}{Lexic. Order} \\
&&&&&&&&\multicolumn{2}{r}{cost} & \multicolumn{2}{r}{time} & \multicolumn{2}{r}{weight} &\multicolumn{2}{r}{cost time weight}&\multicolumn{2}{r}{weight time cost}\\
&&&&&&&&\multicolumn{2}{r}{(2N terms)} & \multicolumn{2}{r}{(5N terms)} & \multicolumn{2}{r}{(16N terms)} &&&&\\
\rot{Experiment} & \rot{Number of Instances} & \rot{Number of Replicas  (N)} & \rot{Total Number of Nodes} & \rot{Number of Rational Variables} & \rot{\% Unrealizable} & \rot{Solving Time} & \rot{Time for Proving Unrealizable}&\rot{Optimization Time} & \rot{\% Timeout} & \rot{Optimization Time} & \rot{\% Timeout} & \rot{Optimization Time} & \rot{\% Timeout} & \rot{Optimization Time} & \rot{\% Timeout} & \rot{Optimization Time} & \rot{\% Timeout}\\
\noalign{\smallskip}\hline\noalign{\smallskip}
  
1 & 100 & 2 & 90 & 52 & 4 & 0.00 & 0.00  & 0.01 & 0 & 0.01 & 0 & 0.02 & 0 & 0.03 & 0 & 0.03 & 0 \\
  
2 & 100 & 3 & 134 & 78 & 3 & 0.00 & 0.00  & 0.00 & 0 & 0.01 & 0 & 0.06 & 0 & 0.04 & 0 & 0.08 & 0\\
  
3 & 100 & 4 & 178 & 104 & 6 & 0.00 & 0.00  & 0.01 & 0 & 0.01 & 0 & 0.07 & 0 & 0.04 & 0 & 0.11 & 0 \\
  
4 & 100 & 5 & 222 & 130 & 6 & 0.00 & 0.00  & 0.01 & 0 & 0.02 & 0 & 0.56 & 0 & 0.07 & 0 & 0.67 & 0\\
  
5 & 100 & 6 & 266 & 156 & 7 & 0.00 & 0.00  & 0.03 & 0 & 0.03 & 0 & 1.51 & 0 & 0.14 & 0 & 1.69 & 0\\
  
6 & 100 & 7 & 310 & 182 & 5 & 0.00 & 0.00  & 0.01 & 0 & 0.03 & 0 & 0.45 & 0 & 0.11 & 0 & 0.69 & 0\\
  
7 & 100 & 9 & 398 & 234 & 7 & 0.00 & 0.00  & 0.02 & 0 & 0.27 & 0 & 284.79 & 31 & 0.71 & 0 & 557.00 & 36\\
  
8 & 100 & 11 & 486 & 286 & 9 & 0.00 & 0.00  & 0.02 & 0 & 0.13 & 0 & 852.66 & 80 & 0.92 & 0 & 705.92 & 85\\
  
9 & 100 & 13 & 574 & 338 & 17 & 0.00 & 0.00 & 0.03 & 0 & 0.17 & 0 & --- & --- & 47.55 & 9 & --- & --- \\
  
10 & 100 & 15 & 662 & 390 & 14 & 0.00 & 0.00  & 0.04 & 0 & 1.23 & 0 & --- & --- & 111.58 & 28 & --- & ---\\
  
11 & 100 & 17 & 750 & 442 & 13 & 0.00 & 0.00  & 0.05 & 0 & 11.87 & 0 & --- & --- & 35.31 & 56 & --- & ---\\
  
12 & 100 & 21 & 926 & 546 & 24 & 0.00 & 0.00  & 0.07 & 0 & 104.67 & 0 & --- & --- & --- & --- & --- & ---\\
  
13 & 100 & 26 & 1146 & 676 & 27 & 0.00 & 0.00  & 0.12 & 0 & 455.20 & 51 & --- & --- & --- & --- & --- & ---\\
  
14 & 100 & 31 & 1366 & 806 & 32 & 0.00 & 0.00  & 0.12 & 0 & --- & --- & --- & --- & --- & --- & --- & ---\\
  
15 & 100 & 36 & 1586 & 936 & 33 & 0.00 & 0.00  & 0.12 & 0 & --- & --- & --- & --- & --- & --- & --- & ---\\
  
16 & 100 & 41 & 1806  & 1066 & 33 & 0.00 & 0.00  & 0.16 & 0 & --- & --- & --- & --- & --- & --- & --- & ---\\
  
17 & 100 & 46 & 2026 & 1196 & 53 & 0.00 & 0.00  & 0.16 & 0 & --- & --- & --- & --- & --- & --- & --- & ---\\
  
18 & 100 & 51 & 2246 & 1326 & 48 & 0.00 & 0.00  & 0.23 & 0 & --- & --- & --- & --- & --- & --- & --- & ---\\
  
19 & 100 & 101 & 4446 & 2626 & 73 & 0.00 & 0.00  & 0.51 & 0 & --- & --- & --- & --- & --- & --- & --- & ---\\
  
20 & 100 & 151 & 6646 & 3926 & 76 & 0.00 & 0.00  & 3.33 & 0 & --- & --- & --- & --- & --- & --- & --- & ---\\
  
21 & 100 & 201 & 8846 & 5226 & 93 & 0.00 & 0.00  & 1.49 & 0 & --- & --- & --- & --- & --- & --- & --- & ---\\
\noalign{\smallskip}\hline  
\end{tabular}
}
\caption{First group of experiments, $k=5$: median time over solved instances.\label{tabExpData3}
}
\end{figure*}

\begin{figure*}[t] 
\centering
\resizebox{\textwidth}{!}{
\begin{tabular}{rrrrrrrrrrrrrrrrrr}
\hline\noalign{\smallskip}
&&&&&&&& \multicolumn{2}{r}{Optimum} & \multicolumn{2}{r}{Optimum} & \multicolumn{2}{r}{Optimum} & \multicolumn{2}{r}{Lexic. Order} & \multicolumn{2}{r}{Lexic. Order} \\
&&&&&&&&\multicolumn{2}{r}{cost} & \multicolumn{2}{r}{time} & \multicolumn{2}{r}{weight} &\multicolumn{2}{r}{cost time weight}&\multicolumn{2}{r}{weight time cost}\\
&&&&&&&&\multicolumn{2}{r}{(2N terms)} & \multicolumn{2}{r}{(5N terms)} & \multicolumn{2}{r}{(16N terms)} &&&&\\
\rot{Experiment} & \rot{Number of Instances} & \rot{Number of Replicas  (N)} & \rot{Total Number of Nodes} & \rot{Number of Rational Variables} & \rot{\% Unrealizable} & \rot{Solving Time} & \rot{Time for Proving Unrealizable}&\rot{Optimization Time} & \rot{\% Timeout} & \rot{Optimization Time} & \rot{\% Timeout} & \rot{Optimization Time} & \rot{\% Timeout} & \rot{Optimization Time} & \rot{\% Timeout} & \rot{Optimization Time} & \rot{\% Timeout}\\
\noalign{\smallskip}\hline\noalign{\smallskip}
  
1 & 100 & 2 & 90 & 52 & 10 & 0.00 & 0.00  & 0.01 & 0 & 0.01 & 0 & 0.03 & 0 & 0.02 & 0 & 0.04 & 0 \\
  
2 & 100 & 3 & 134 & 78 & 15 & 0.00 & 0.00 & 0.01 & 0 & 0.01 & 0 & 0.03 & 0 & 0.04 & 0 & 0.06 & 0 \\
  
3 & 100 & 4 & 178 & 104 & 9 & 0.00 & 0.00  & 0.01 & 0 & 0.01 & 0 & 0.14 & 0 & 0.04 & 0 & 0.19 & 0 \\
  
4 & 100 & 5 & 222 & 130 & 11 & 0.00 & 0.00   & 0.01 & 0 & 0.02 & 0 & 0.07 & 0 & 0.06 & 0 & 0.09 & 0 \\
  
5 & 100 & 6 & 266 & 156 & 21 & 0.00 & 0.00  & 0.01 & 0 & 0.03 & 0 & 1.85 & 0 & 0.07 & 0 & 2.25 & 0 \\
  
6 & 100 & 7 & 310 & 182 & 24 & 0.00 & 0.00  & 0.01 & 0 & 0.02 & 0 & 14.71 & 0 & 0.10 & 0 & 14.11 & 0 \\
  
7 & 100 & 9 & 398 & 234 & 33 & 0.00 & 0.00  & 0.01 & 0 & 0.11 & 0 & 17.37 & 1 & 0.15 & 0 & 25.14 & 1 \\
  
8 & 100 & 11 & 486 & 286 & 23 & 0.00 & 0.00 & 0.03 & 0 & 0.31 & 0 & 79.55 & 19 & 0.51 & 0 & 253.57 & 28 \\
  
9 & 100 & 13 & 574 & 338 & 28 & 0.00 & 0.00 & 0.03 & 0 & 0.22 & 0 & 131.37 & 55 & 0.64 & 0 & 240.96 & 59  \\
  
10 & 100 & 15 & 662 & 390 & 36 & 0.00 & 0.00  & 0.04 & 0 & 0.41 & 0 & --- & --- & 6.89 & 0 & --- & --- \\
  
11 & 100 & 17 & 750 & 442 & 20 & 0.00 & 0.00  & 0.05 & 0 & 0.86 & 0 & --- & --- & 0.56 & 1 & --- & --- \\
  
12 & 100 & 21 & 926 & 546 & 48 & 0.00 & 0.00   & 0.05 & 0 & 149.86 & 7 & --- & --- & 104.81 & 17 & --- & ---  \\
  
13 & 100 & 26 & 1146 & 676 & 43 & 0.00 & 0.00  & 0.06 & 0 & 406.31 & 23 & --- & --- & --- & --- & --- & --- \\
  
14 & 100 & 31 & 1366 & 806 & 61 & 0.00 & 0.00  & 0.10 & 0 & --- & --- & --- & --- & --- & --- & --- & --- \\
  
15 & 100 & 36 & 1586 & 936 & 67 & 0.00 & 0.00  & 0.23 & 0 & --- & --- & --- & --- & --- & --- & --- & --- \\
  
16 & 100 & 41 & 1806  & 1066 & 71 & 0.00 & 0.00  & 0.39 & 0 & --- & --- & --- & --- & --- & --- & --- & --- \\
  
17 & 100 & 46 & 2026 & 1196 & 77 & 0.00 & 0.00  & 0.17 & 0 & --- & --- & --- & --- & --- & --- & --- & --- \\
  
18 & 100 & 51 & 2246 & 1326 & 75 & 0.00 & 0.00  & 0.17 & 0 & --- & --- & --- & --- & --- & --- & --- & --- \\
  
19 & 100 & 101 & 4446 & 2626 & 98 & 0.00 & 0.00  & 1.47 & 0 & --- & --- & --- & --- & --- & --- & --- & --- \\
  
20 & 100 & 151 & 6646 & 3926 & 97 & 0.00 & 0.00   & 40.11 & 0 & --- & --- & --- & --- & --- & --- & --- & --- \\
  
21 & 100 & 201 & 8846 & 5226 & 100 & 0.00 & 0.00 & --- & --- & --- & --- & --- & --- & --- & --- & --- & --- \\
\noalign{\smallskip}\hline  
\end{tabular}
}
\caption{First group of experiments, $k=8$: median time over solved instances.\label{tabExpData4}
}
\end{figure*}

\begin{figure*}
\centering 
\includegraphics[height=0.95\textheight]{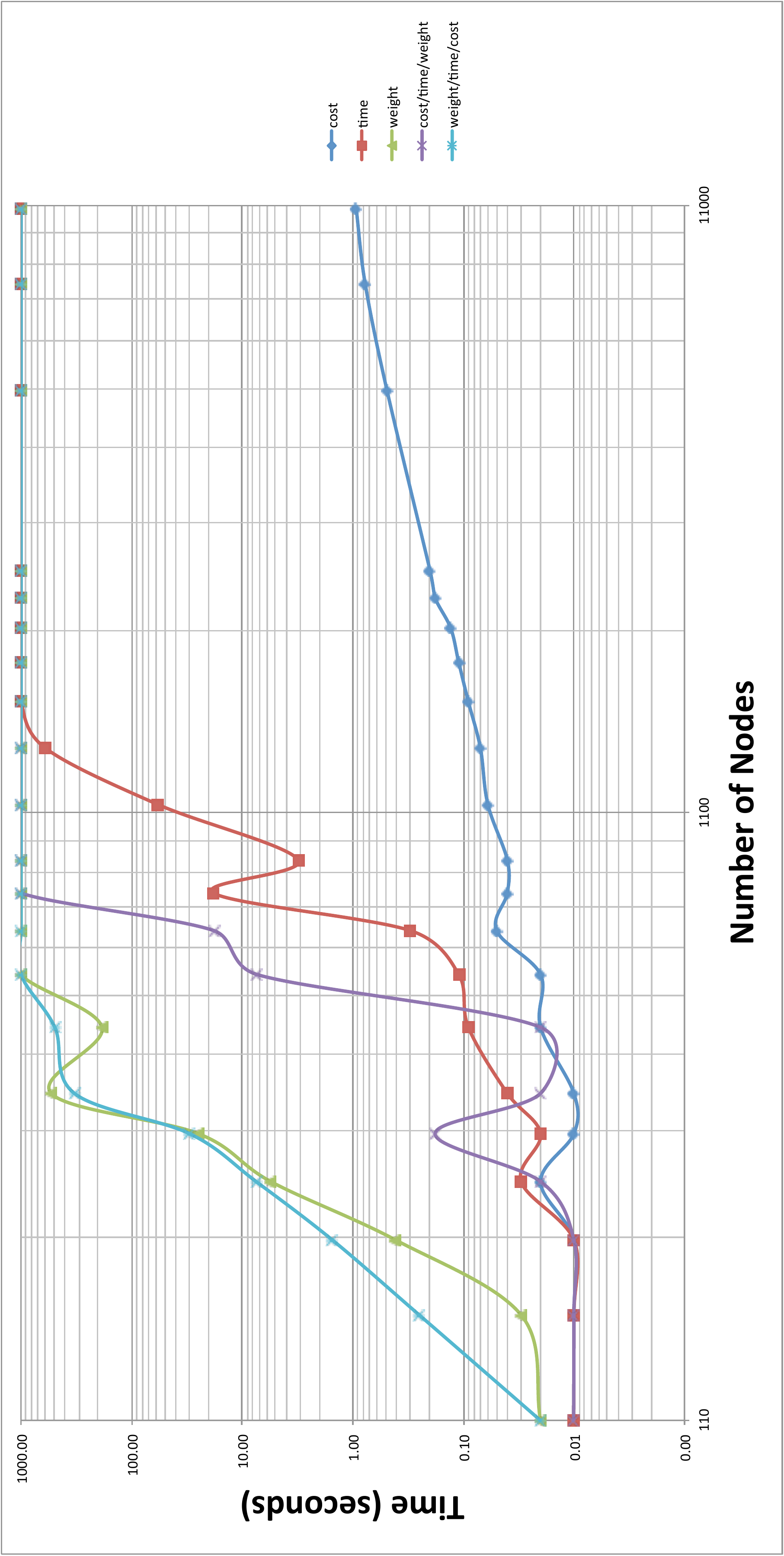}
\caption{{First group of experiments, $k=2$, median run times over solved instances.}
\label{figRuntime2N}}
\end{figure*}

\begin{figure*}
\centering 
\includegraphics[height=0.95\textheight]{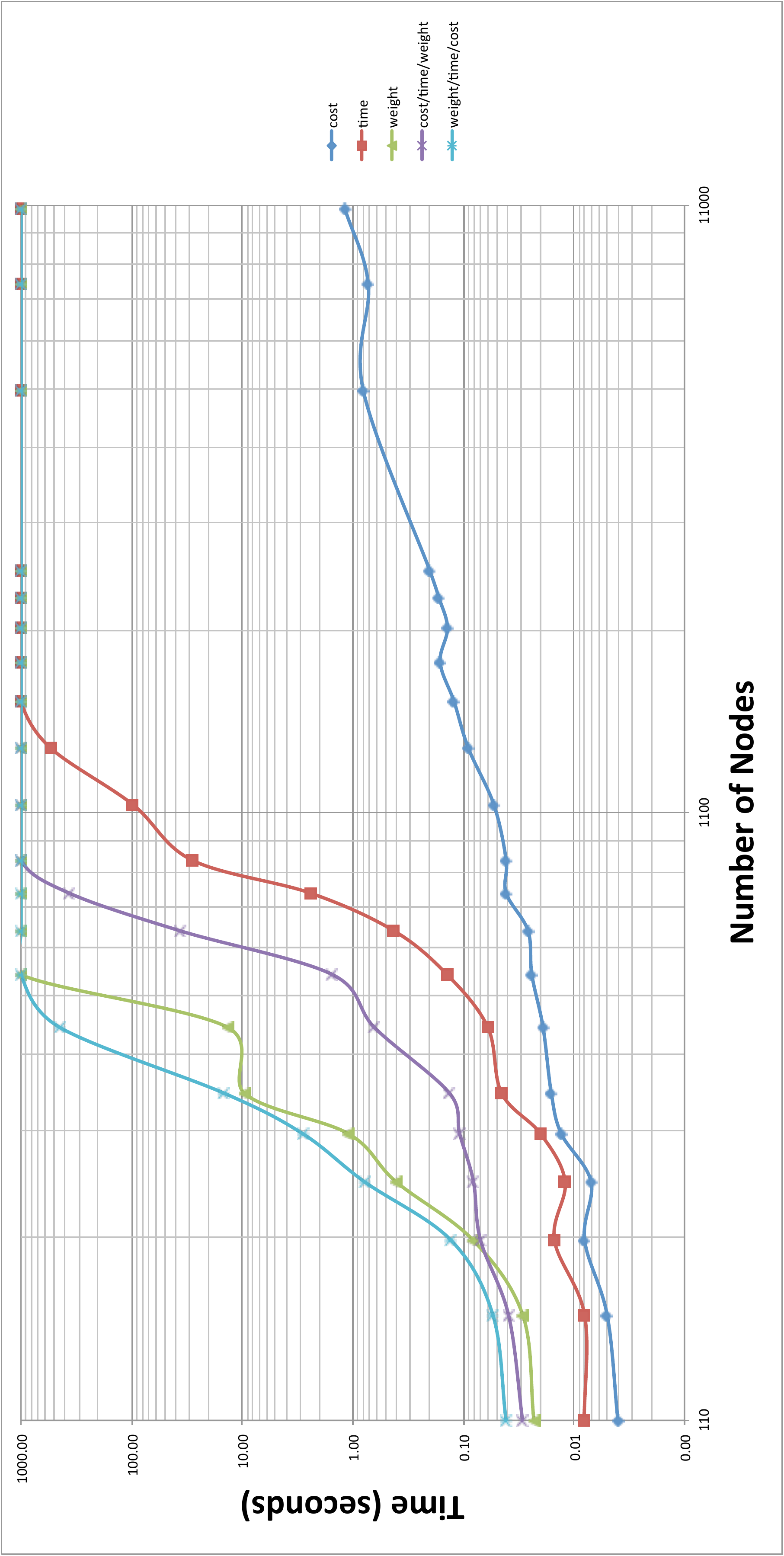}
\caption{{First group of experiments, $k=4$, median run times over solved instances.}
\label{figRuntime4N}}
\end{figure*}

\begin{figure*}
\centering 
\includegraphics[height=0.95\textheight]{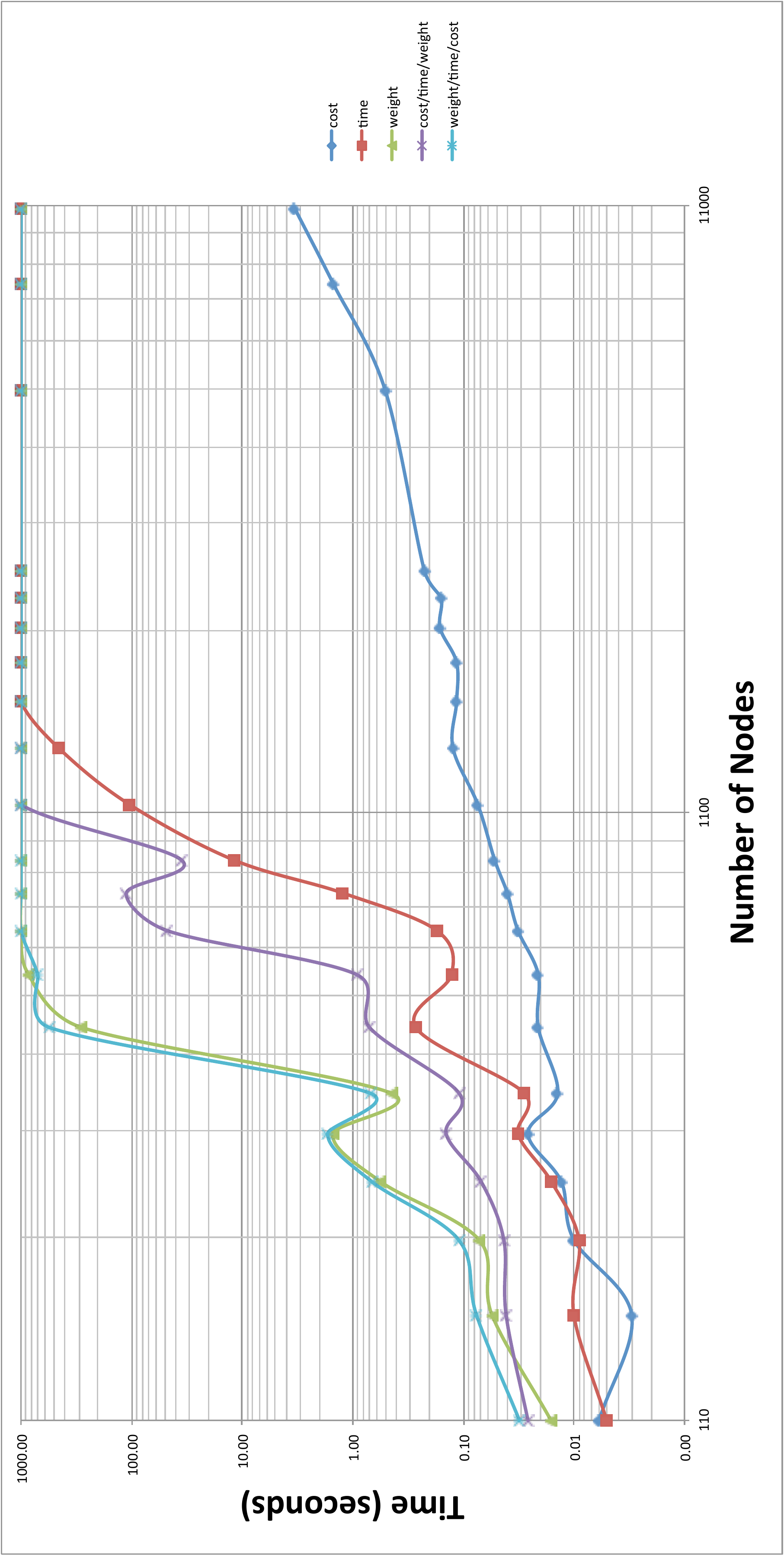}
\caption{{First group of experiments, $k=5$, median run times over solved instances.}
\label{figRuntime5N}}
\end{figure*}

\begin{figure*}
\centering 
\includegraphics[height=0.95\textheight]{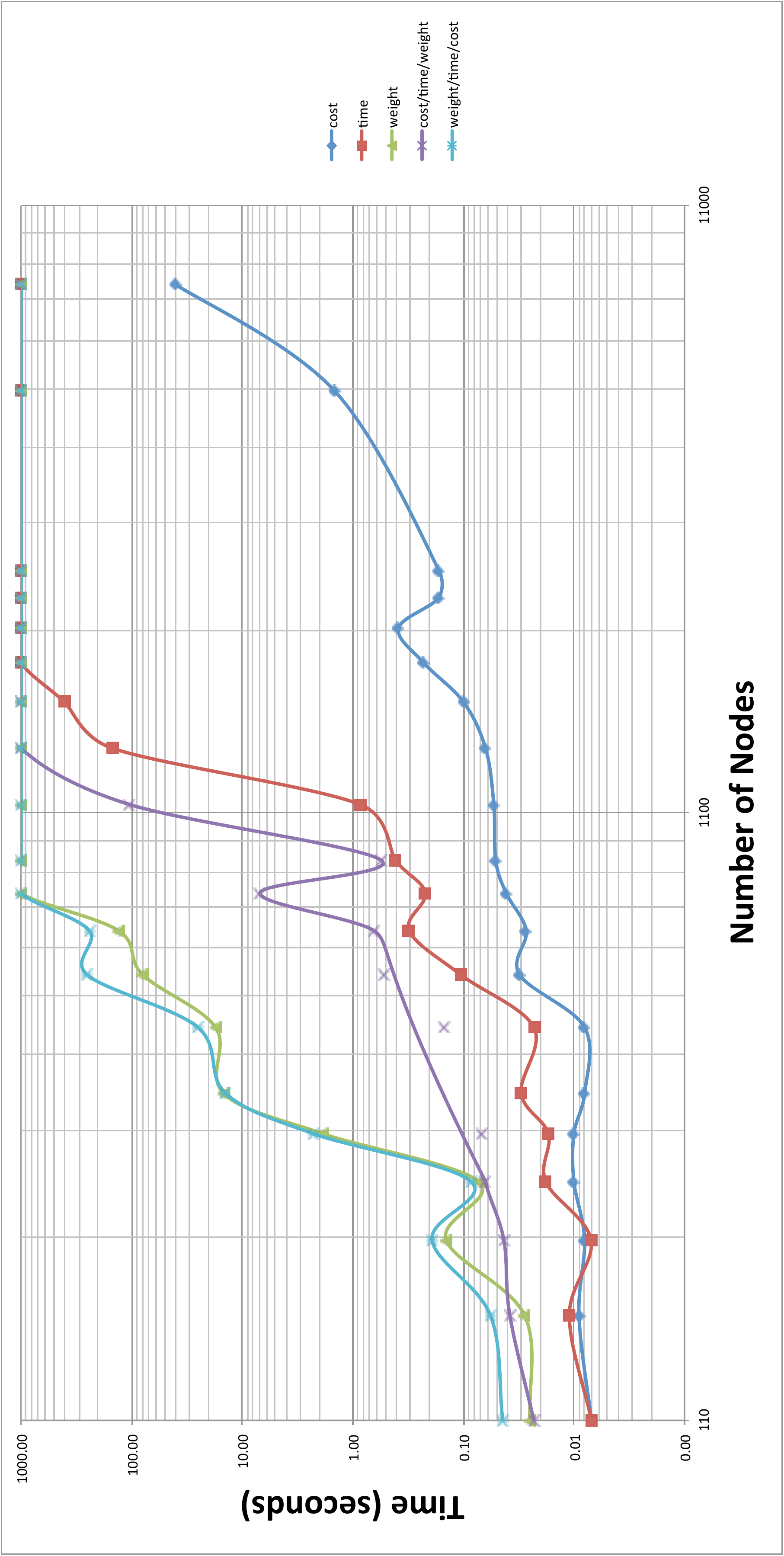}
\caption{{First group of experiments, $k=8$, median run times over solved instances.}
\label{figRuntime8N}}
\end{figure*}

\FloatBarrier
\subsection{Second Group of Experiments}
\begin{figure*}[b] 
\centering
    \begin{tabular}{rrrrrrrrrrrr}
    \hline\noalign{\smallskip}
    &&&&&&& \multicolumn{2}{r}{Lexic. Order PRT} & \multicolumn{2}{r}{Lexic. Order RPT} \\
    \rot{Experiment} & \rot{Number of Instances.} & \rot{Number of Replicas (N)} & \rot{Total Number of Nodes} &\rot{Number of Rational Variables} & \rot{\% Unrealizable} & \rot{Solving Time} & \rot{Time for Proving Unrealizable}&\rot{Optimization Time} & \rot{\% Timeout} & \rot{Optimization Time} & \rot{\% Timeout} \\
    \noalign{\smallskip}\hline\noalign{\smallskip}

        1   & 100 &   2 & 110 & 60 &  1 & 0.00 & 0.00 &   0.04 &   0 & 0.08 &   0 \\
        
        2   & 100 &   3 & 164 & 90 &  2 & 0.00 & 0.00 &   0.07 &   0 & 0.08 &   0 \\
        
        3   & 100 &   4 & 218 & 120 &  1 & 0.00 & 0.00 &   0.11 &   0 & 0.09 &   0 \\
        
        4   & 100 &   5 & 272 & 150 &  3 & 0.00 & 0.00 &   0.12 &   0 & 0.15 &   0 \\
        
        5   & 100 &   6 & 326 & 180 &  2 & 0.00 & 0.00 &   0.13 &   0 & 0.20 &   0 \\
        
        6   & 100 &   7 & 380 & 210 &  3 & 0.00 & 0.00 &   0.21 &   0 & 0.26 &   0 \\
        
        7   & 100 &   9 & 488 & 270 &  2 & 0.00 & 0.00 &   0.52 &   0 & 0.45 &   0 \\
        
        8   & 100 &  11 & 596 & 330 &  7 & 0.00 & 0.00 &   0.90 &   0 & 0.50 &   0 \\
        
        9   & 100 &  13 & 704 & 390 &  8 & 0.00 & 0.00 &   2.42 &   0 & 2.33 &   0 \\
        
        10  & 100 &  15 & 812 & 450 & 4 & 0.00 & 0.00 &   39.45 &   0 & 1.55 &   0 \\
        
        11  & 100 &  17 & 920 & 510 &  6 & 0.00 & 0.00 &   1.64 &   0 & 1.57 &   0 \\
        
        12  & 100 &  21 & 1136 & 630 & 7 & 0.00 & 0.00 &   694.50 &   52 & 468.88 &   20 \\
        
        13  & 100 &  26 & 1406 & 780 & 6 & 0.00 & 0.00 & --- &  --- & --- &   --- \\
        
        14  & 100 &  31 & 1676 & 930 & 14 & 0.00 & 0.00 & --- &  --- & --- &  --- \\
        
        15  & 100 &  36 & 1946 & 1080 & 15 & 0.00 & 0.00 &  ---   & --- & ---  & --- \\ 
        
        16  & 100 &  41 & 2216 & 1230 & 19 & 0.00 & 0.00 &  ---   & --- & ---  & --- \\
        
        17  & 100 &  46 & 2486 & 1380 & 16 & 0.00 & 0.00 &  ---   & --- & ---  & --- \\
        
        18  & 100 &  51 & 2756 & 1530 & 27 & 0.00 & 0.00 &  ---   & --- & ---  & --- \\
        
        19  & 100 & 101 & 5456 & 3030 & 33 & 0.00 & 0.00 &  ---   & --- & ---  & --- \\
        
        20  & 100 & 151 & 8156 & 4530 & 46 & 0.00 & 0.00 &  ---   & --- & ---  & --- \\
        
        21  & 100 & 201 & 10856 & 6030 & 56 & 0.00 & 0.00 &  ---   & --- & ---  & --- \\
    
\noalign{\smallskip}\hline
    \end{tabular}
\caption{Second group of experiments, $k=2$, $p=6$: median time over solved instances.\label{tabExpData11}
}

\end{figure*}

\begin{figure*}[t] 
\centering
    \begin{tabular}{rrrrrrrrrrrr}
    \hline\noalign{\smallskip}
    &&&&&&& \multicolumn{2}{r}{Lexic. Order PRT} & \multicolumn{2}{r}{Lexic. Order RPT} \\
    \rot{Experiment} & \rot{Number of Instances.} & \rot{Number of Replicas (N)} & \rot{Total Number of Nodes} &\rot{Number of Rational Variables} & \rot{\% Unrealizable} & \rot{Solving Time} & \rot{Time for Proving Unrealizable}&\rot{Optimization Time} & \rot{\% Timeout} & \rot{Optimization Time} & \rot{\% Timeout} \\
    \noalign{\smallskip}\hline\noalign{\smallskip}
    
        1   & 100 &   2 & 110 & 60 &  0 & 0.00 & 0.00 &   0.06 &   0 & 0.07 &   0 \\
        
        2   & 100 &   3 & 164 & 90 &  1 & 0.00 & 0.00 &   0.08 &   0 & 0.08 &   0 \\
        
        3   & 100 &   4 & 218 & 120 &  0 & 0.00 & 0.00 &   0.18 &   0 & 0.09 &   0 \\
        
        4   & 100 &   5 & 272 & 150 &  2 & 0.00 & 0.00 &   0.18 &   0 & 0.14 &   0 \\
        
        5   & 100 &   6 & 326 & 180 &  1 & 0.00 & 0.00 &   0.36 &   0 & 0.18 &   0 \\
        
        6   & 100 &   7 & 380 & 210 &  2 & 0.00 & 0.00 &   0.21 &   0 & 0.20 &   0 \\
        
        7   & 100 &   9 & 488 & 270 &  6 & 0.00 & 0.00 &   0.28 &   0 & 0.30 &   0 \\
        
        8   & 100 &  11 & 596 & 330 &  4 & 0.00 & 0.00 &   0.61 &   0 & 0.47 &   0 \\
        
        9   & 100 &  13 & 704 & 390 &  6 & 0.00 & 0.00 &   0.73 &   0 & 0.53 &   0 \\
        
        10  & 100 &  15 & 812 & 450 & 12 & 0.00 & 0.00 &   1.38 &   0 & 0.69 &   0 \\
        
        11  & 100 &  17 & 920 & 510 &  6 & 0.00 & 0.00 &   1.81 &   0 & 0.99 &   0 \\
        
        12  & 100 &  21 & 1136 & 630 & 10 & 0.00 & 0.00 &   7.00 &   0 & 3.92 &   0 \\
        
        13  & 100 &  26 & 1406 & 780 & 11 & 0.00 & 0.00 & 330.39 &  10 & 9.38 &   1 \\
        
        14  & 100 &  31 & 1676 & 930 & 11 & 0.00 & 0.00 & 327.86 &  72 & 8.40 &  10 \\
        
        15  & 100 &  36 & 1946 & 1080 & 14 & 0.00 & 0.00 &  ---   & --- & ---  & --- \\
        
        16  & 100 &  41 & 2216 & 1230 & 13 & 0.00 & 0.00 &  ---   & --- & ---  & --- \\
        
        17  & 100 &  46 & 2486 & 1380 & 14 & 0.00 & 0.00 &  ---   & --- & ---  & --- \\
        
        18  & 100 &  51 & 2756 & 1530 & 20 & 0.00 & 0.00 &  ---   & --- & ---  & --- \\
        
        19  & 100 & 101 & 5456 & 3030 & 33 & 0.00 & 0.00 &  ---   & --- & ---  & --- \\
        
        20  & 100 & 151 & 8156 & 4530 & 40 & 0.00 & 0.00 &  ---   & --- & ---  & --- \\
        
        21  & 100 & 201 & 10856 & 6030 & 59 & 0.00 & 0.00 &  ---   & --- & ---  & --- \\
    
    \end{tabular}
\caption{Second group of experiments, $k=2$, $p=8$: median time over solved instances.\label{tabExpData12}
}
    \begin{tabular}{rrrrrrrrrrrr}
    \hline\noalign{\smallskip}
    &&&&&&& \multicolumn{2}{r}{Lexic. Order PRT} & \multicolumn{2}{r}{Lexic. Order RPT} \\
    \rot{Experiment} & \rot{Number of Instances.} & \rot{Number of Replicas (N)} & \rot{Total Number of Nodes} &\rot{Number of Rational Variables} & \rot{\% Unrealizable} & \rot{Solving Time} & \rot{Time for Proving Unrealizable}&\rot{Optimization Time} & \rot{\% Timeout} & \rot{Optimization Time} & \rot{\% Timeout} \\
    \noalign{\smallskip}\hline\noalign{\smallskip}
            
            1   & 100 &   2 &   110 & 60 & 1 & 0.00 & 0.00 &   0.06 &   0 &  0.06 &   0 \\
            
            2   & 100 &  3 &   164 & 90 & 0 & 0.00 & 0.00 &   0.09 &   0 &  0.07 &   0 \\
            
            3   & 100 &   4 &   218 & 120 &  0 & 0.00 & 0.00 &   0.13 &   0 &  0.12 &   0 \\
            
            4   & 100 &   5 &   272 & 150 & 0 & 0.00 & 0.00 &   0.15 &   0 &  0.17 &   0 \\
            
            5   & 100 &   6 &   326 & 180 & 0 & 0.00 & 0.00 &   0.25 &   0 &  0.20 &   0 \\
            
            6   & 100 &   7 &   380 & 210 & 0 & 0.00 & 0.00 &   0.39 &   0 &  0.30 &   0 \\
            
            7   & 100 &   9 &   488 & 270 & 0 & 0.00 & 0.00 &   0.43 &   0 &  0.49 &   0 \\
            
            8   & 100 &  11 &   596 & 330 & 0 & 0.00 & 0.00 &   0.81 &   0 &  0.56 &   0 \\
            
            9   & 100 &  13 &   704 & 390 & 1 & 0.00 & 0.00 &   1.15 &   0 &  0.89 &   0 \\
            
            10  & 100 &  15 &   812 & 450 & 1 & 0.00 & 0.00 &   1.32 &   0 &  0.37 &   0 \\
            
            11  & 100 &  17 &   920 & 510 & 2 & 0.00 & 0.00 &  14.66 &   0 &  1.97 &   0 \\
            
            12  & 100 &  21 & 1136 & 630 &  0 & 0.00 & 0.00 & 602.22 &  23 &  2.13 &   0 \\
            
            13  & 100 &  26 & 1406 & 780 &  2 & 0.00 & 0.00 & 911.26 &  87 & 905.11  & 9 \\
            
            14  & 100 &  31 & 1676 & 930 &  4 & 0.00 & 0.00 &  ---   & --- & 14.79  & 24 \\
            
            15  & 100 &  36 & 1946 & 1080 &  0 & 0.00 & 0.00 &  ---   & --- & ---  & --- \\
            
            16  & 100 &  41 & 2216 & 1230 &  1 & 0.00 & 0.00 &  ---   & --- & ---  & --- \\
            
            17  & 100 &  46 & 2486 & 1380 &  2 & 0.00 & 0.00 &  ---   & --- & ---  & --- \\
            
            18  & 100 &  51 & 2756 & 1530 &  1 & 0.00 & 0.00 &  ---   & --- & ---  & --- \\
            
            19  & 100 & 101 & 5456 & 3030 &  5 & 0.00 & 0.00 &  ---   & --- & ---  & --- \\
            
            20  & 100 & 151 & 8156 & 4530 &  5 & 0.00 & 0.00 &  ---   & --- & ---  & --- \\
            
            21  & 100 & 201 & 10856 & 6030 & 10 & 0.00 & 0.00 &  ---   & --- & ---  & --- \\
            
\noalign{\smallskip}\hline
            \end{tabular}
    \caption{Second group of experiments, $k=2$, $p=12$: median time over solved instances.\label{tabExpData13}
}
        \end{figure*}

\begin{figure*}
\centering 
\includegraphics[height=0.95\textheight]{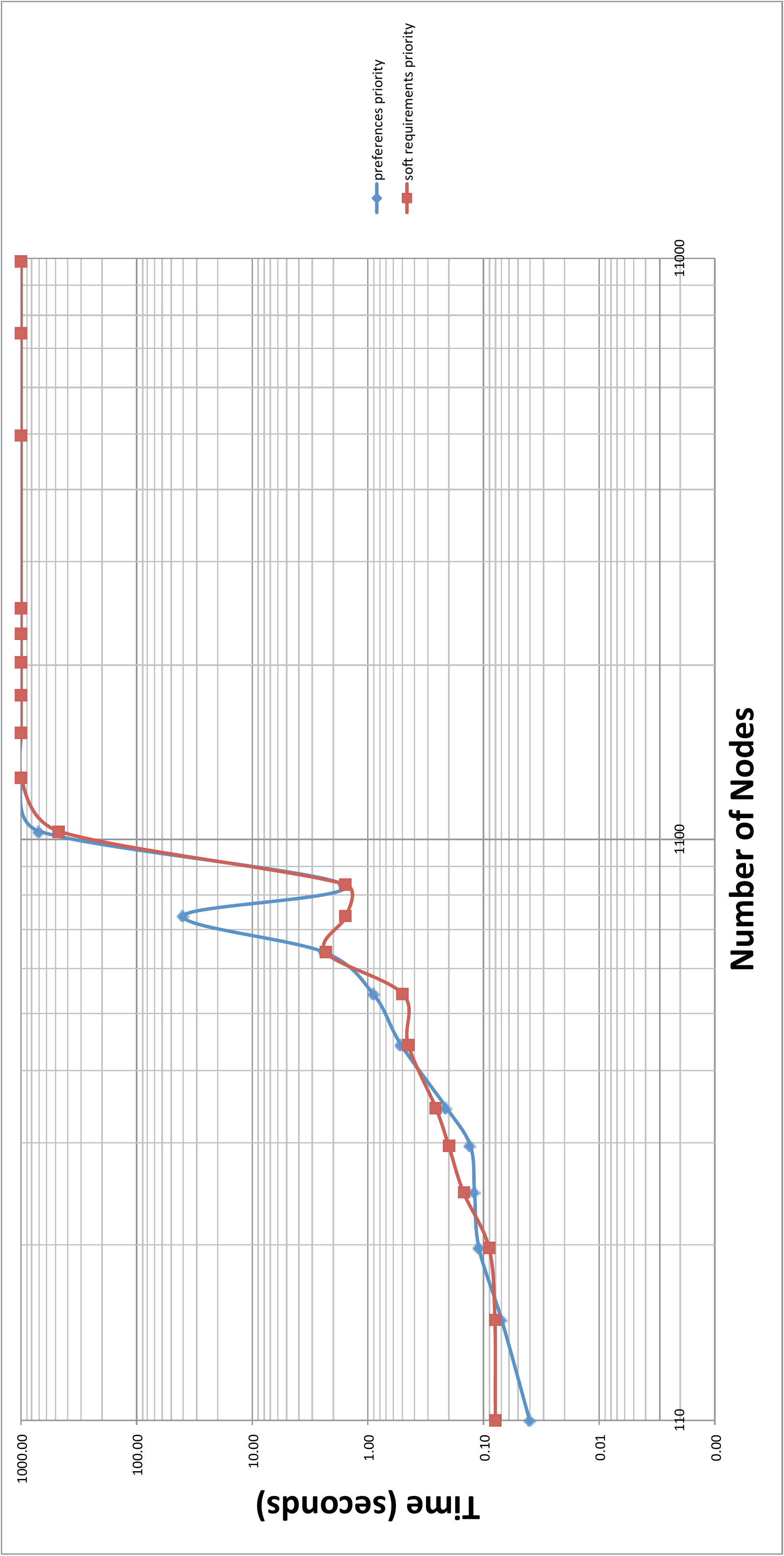}
\caption{{Second group of experiments, $k=2$, $k=6$, median run times over solved instances.}
\label{figRuntime2N6P}}
\end{figure*}

\begin{figure*}
\centering 
\includegraphics[height=0.95\textheight]{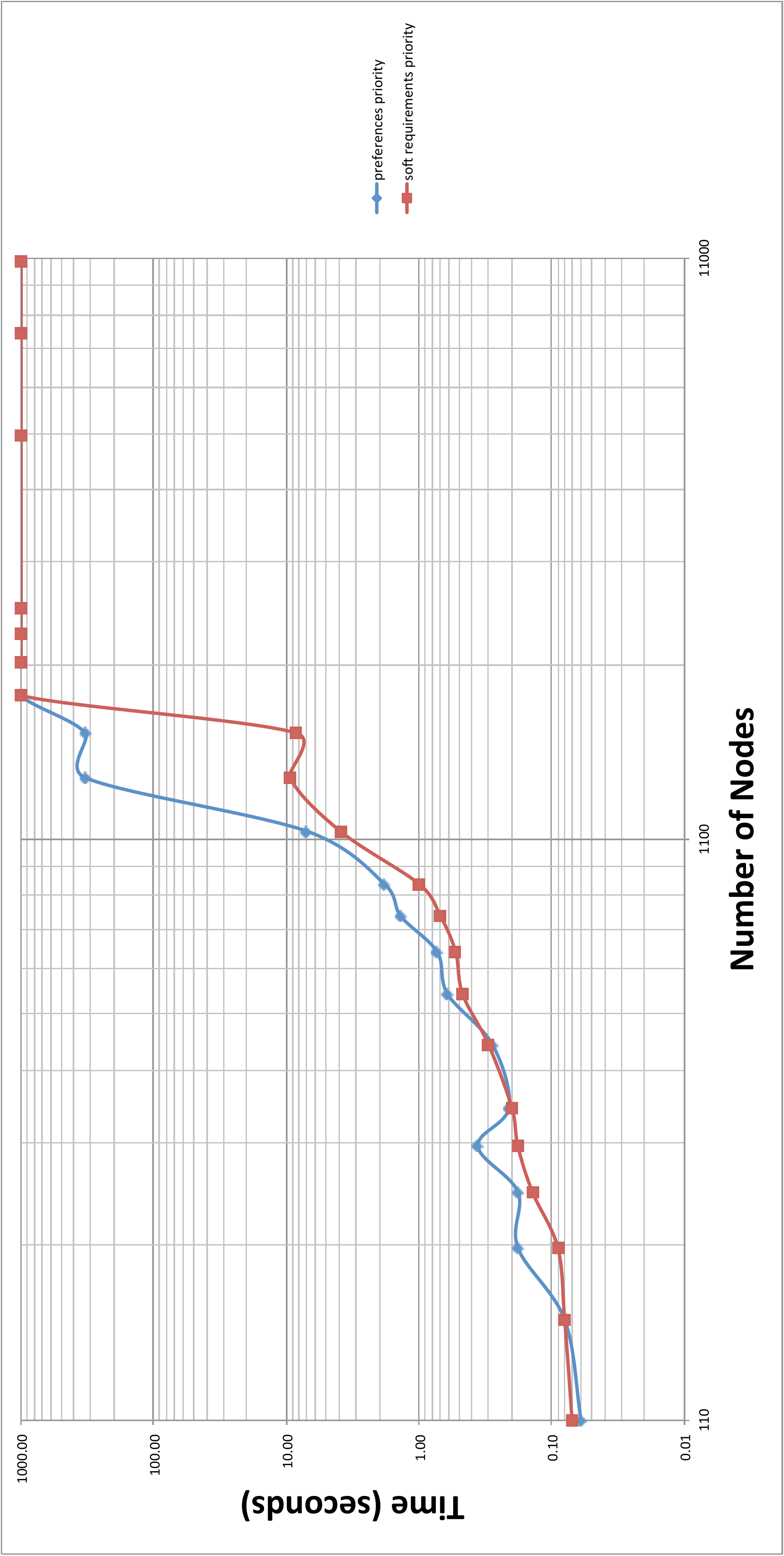}
\caption{{Second group of experiments, $k=2$, $k=8$, median run times over solved instances.}
\label{figRuntime2N8P}}
\end{figure*}

\begin{figure*}
\centering 
\includegraphics[height=0.95\textheight]{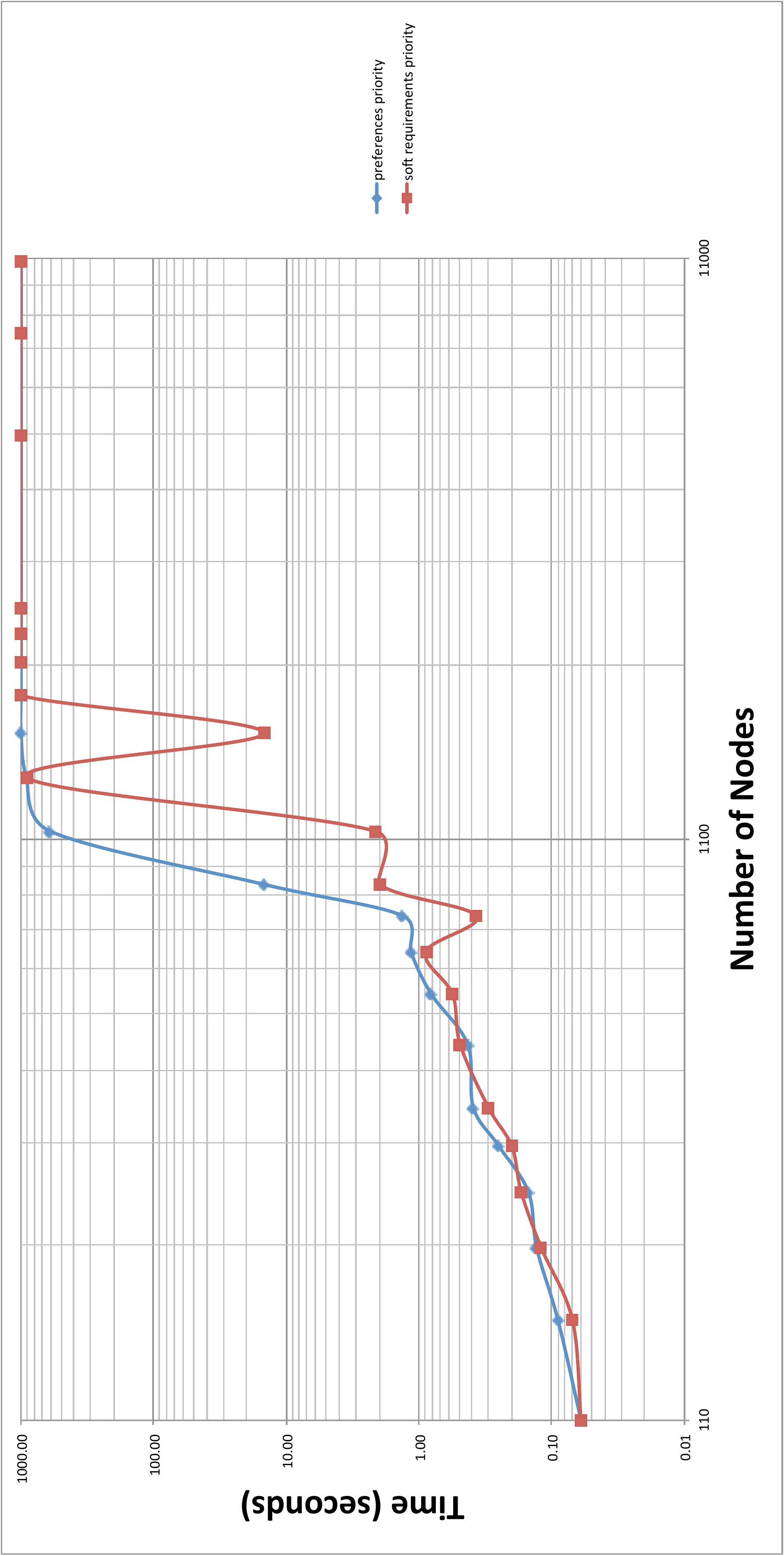}
\caption{{Second group of experiments, $k=2$, $k=12$, median run times over solved instances.}}
\label{figRuntime2N12P}
\end{figure*}

\ignore{
\begin{figure*}
\centering 
\includegraphics[height=0.95\textheight]{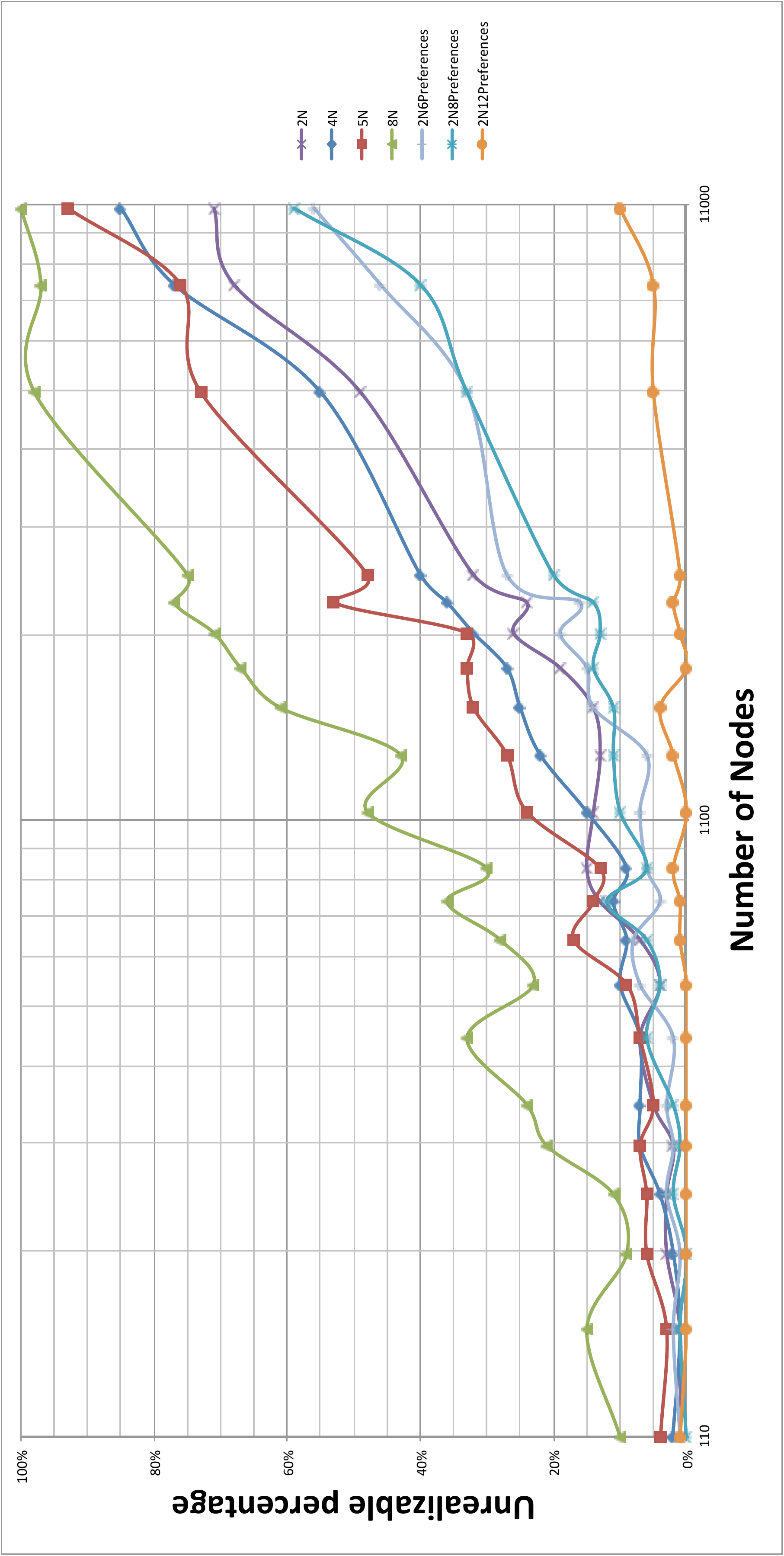}
\caption{{Percentage of un-realizable instances, both groups of experiments.}
\label{figUNSATPercentage}}
\end{figure*}
}


\end{document}